\documentclass[a4paper, 11pt]{article}
\usepackage[english]{babel}
\usepackage[utf8x]{inputenc}
\usepackage[a4paper,top=2.54cm,bottom=2.54cm,left=2.54cm,right=2.54cm]{geometry}
\usepackage{amsmath}
\usepackage{graphicx}
\usepackage[colorinlistoftodos]{todonotes}
\usepackage[colorlinks=true, allcolors=blue]{hyperref}
\usepackage{natbib}
\usepackage{setspace}
\usepackage{soul}
\usepackage{enumitem}
\usepackage{algorithm}
\usepackage{algpseudocode}
\usepackage{epstopdf}
\usepackage{float}
\title{The FEDHC Bayesian network learning algorithm}

\author{Michail Tsagris \\
Department of Economics, University of Crete, Greece \\
mtsagris@uoc.gr}

\begin{document}

\maketitle

\begin{abstract}
\noindent The paper proposes a new hybrid Bayesian network learning algorithm, termed Forward Early Dropping Hill Climbing (FEDHC), devised to work with either continuous or categorical variables. Further, the paper manifests that the only implementation of MMHC in the statistical software \textit{R}, is prohibitively expensive and a new implementation is offered. Further, specifically for the case of continuous data, a robust to outliers version of FEDHC, that can be adopted by other BN learning algorithms, is proposed. The FEDHC is tested via Monte Carlo simulations that distinctly show it is computationally efficient, and produces Bayesian networks of similar to, or of higher accuracy than MMHC and PCHC. Finally, an application of FEDHC, PCHC and MMHC algorithms to real data, from the field of economics, is demonstrated using the statistical software \textit{R}. \\
\\
Keywords: Causality, Bayesian networks, scalability
\end{abstract}

%\newpage

\section{Introduction} \label{intro}
Learning the causal relationships among variables using non-experimental data is of high importance in many scientific fields, such as economics and econometrics\footnote{For a general definition of causality specifically in economics and econometrics see \cite{hoover2017}.}. When the aim is particularly to recreate the causal mechanism that generated the data, graphical models, such as causal networks and Bayesian Networks (BNs)\footnote{Also known as Bayes networks, belief networks, decision networks, Bayes(ian) models or probabilistic directed acyclic graphical models.} are frequently employed. The advantages of BNs include simultaneous variable selection, among all variables and hence detection of conditional associations between variables. On a different route BNs form the scheme for synthetic population generation \citep{sun2015} and have been used synergetically with agent based models \citep{kocabas2009,kocabas2013}. 

BNs enjoy applications to numerous fields, but the focus of the current paper is on economics related fields applications, such as production economics \citep{hosseini2016}, macroeconomics \citep{spiegler2016} and environmental resource economics \citep{xue2017}. Applications of BNs can be found also in financial econometrics \citep{mele2017} banking and finance \citep{chong2018}, credit scoring \citep{leong2016}, insurance \citep{sheehan2017} and customer service \citep{cugnata2014} to name a few. Despite the plethora of applications of BNs, not many BN algorithms exist, and most importantly fewer are publicly available in free software environments, such as the statistical software \textit{R}. The Max-Min Hill Climbing (MMHC) \citep{tsamardinos2006} is an example of a widely used BN learning algorithm\footnote{The relevant paper is one of the classic papers in the Artificial Intelligence field and has received more than 1,870 citations according to \textit{google.scholar} by July 2022.} that is publicly available, in the \textit{R} package \textit{bnlearn} \citep{bnlearn2010}. PC Hill Climbing (PCHC) \citep{tsagris2021} is a recently suggested hybrid algorithm that is also publicly available, in the \textit{R} package \textit{pchc} \citep{pchc2021}.

\citep{tsagris2021} showed that when the sample size is at the order of hundreds of thousands MMHC's implementation in the \textit{R} package \textit{bnlearn} requires more than a day with continuous data, even with 50 variables. On the contrary, PCHC is a computationally efficient and scalable BN learning algorithm \citep{tsagris2021}. With modern technology and vast data generation, the computational cost is a considerable parameter. Every novel algorithm must be computationally efficient and scalable to large sample sizes. Seen from the green economy point of view, this cost also has an economic and environmental impact; a faster algorithm will produce results in a more timely manner, facilitating faster decision making, consuming less energy and hence reducing its carbon footprint. 

Moving along those lines this paper proposes a new computationally highly efficient algorithm termed Forward with Early Dropping Hill Climbing (FEDHC) that is publicly available in the \textit{R} package \textit{pchc}. FEDHC shares common ideas with PCHC and MMHC. It applies the Forward Backward with Early Dropping (FBED) variable selection algorithm \citep{borboudakis2019} to each variable as a means of skeleton identification, followed by a Hill Climbing (HC) scoring phase. FEDHC can handle millions of observations in just a few minutes and retains similar or better accuracy levels than PCHC and MMHC. With continuous data FEDHC performs fewer errors than PCHC but the converse is true with categorical data. FEDHC further enjoys the same scalability property as PCHC, its computational cost is proportional to the sample size of the data. Increasing the sample size by a factor increases the execution time by the same factor. Finally, a new, computationally efficient, implementation of MMHC is offered that is also publicly available in the \textit{R} package \textit{pchc}.

The choice of the BN learning algorithm is not only computational cost dependent, but also quality dependant. Regardless of the algorithm used, the quality of the learned BN can be significantly affected by outliers. Robustness to outliers is an important aspect that, surprisingly enough, has not attracted substantial research attention in the field of BN learning. \cite{kalisch2008} were the first to propose a robustified version of the PC algorithm by replacing the empirical standard deviations with robust scale estimates. \cite{cheng2018} on the other hand, removed the outliers but their algorithm is only applicable to BNs with a known topology. Robustification of the proposed FEDHC algorithm takes place by adopting techniques from the robust statistics literature. The key concept is to identify and remove the outliers prior to applying FEDHC.

The rest of the paper is structured as follows. Preliminaries regarding BNs that will assist in making the paper comprehensible and a brief presentation of PCHC and MMHC algorithms are unveiled in Section \ref{prelim}. The FEDHC is introduced in Section \ref{fedhc} along with its robustified version\footnote{ The robustified versions is applicable to PCHC and MMHC as well.} that will be shown to be remarkable and utterly insensitive to outliers. Theoretical properties and computational details of FEDHC and the conditional independence tests utilised for continuous and categorical data are delineated in the same section. Section \ref{mc} contains Monte Carlo simulation studies comparing FEDHC to PCHC and MMHC in terms of accuracy, computational efficiency and number of tests performed. Section \ref{real} illustrates the FEDHC, PCHC and MMHC on two real cross-sectional datasets using the \textit{R} package \textit{pchc}. The first dataset contains continuous data on the credit history for a sample of applicants for a type of credit card \citep{greene2003}. The second dataset contains categorical data on the household income plus some more demographic variables. Ultimately, Section \ref{concl} contains the conclusions drawn from this paper.

\section{Preliminaries} \label{prelim}
Graphical models or probabilistic graphical models express visually the conditional (in)dependencies between random variables ($V_i$, $i=1,\ldots,D$). Nodes (or vecrtices) are used to represent the variables $V_i$ and edges between the nodes, for example $V_i-V_j$, indicate relationship between the variables $V_i$ and $V_j$. Directed graphs are graphical models that contain arrows, instead of edges, indicating the direction of the relationship, for example $V_i \rightarrow V_j$. The parents of a node $V_i$ are the nodes whose direction (arrows) points towards $V_i$. Consequently, node $V_i$ is termed child of those nodes and the common parents of that nodes are called spouses. Directed acyclic graphs (DAG) are stricter in the sense that they impose no cycles on these directions. For any path between $V_i$ and $V_j$, $V_i \rightarrow V_k \rightarrow \ldots \rightarrow V_j $, no path from $V_j$ to $V_i$ ($V_j \rightarrow \ldots \rightarrow V_i$) exists. In other words, the natural sequence or relationship between parents and children or ancestors and descendants is mandated. 

\begin{figure}[!ht]
\centering
\includegraphics[scale = 0.5, trim = 0 20 0 0]{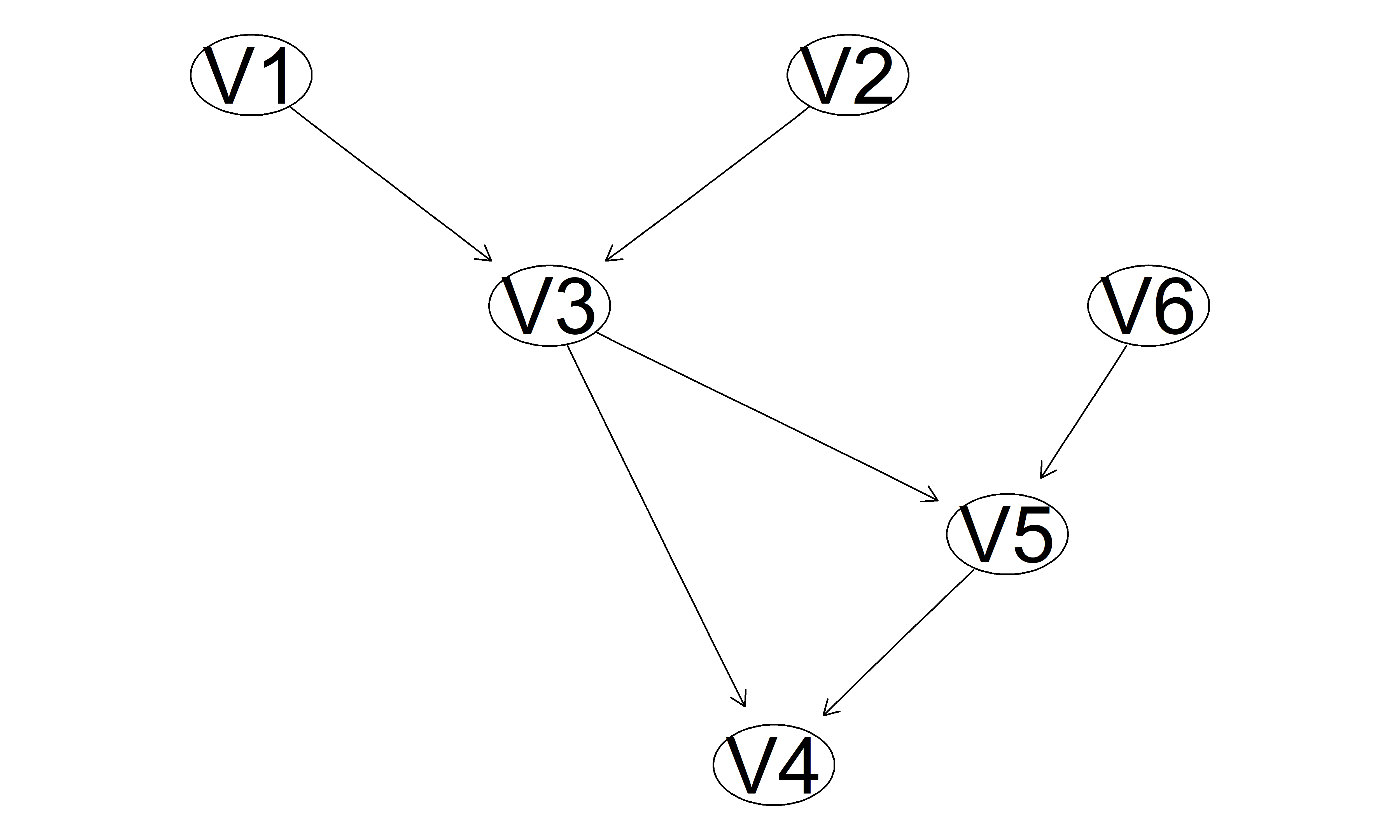} 
\caption{An example of a DAG. Nodes $V1$ and $V2$ are the parents of $V3$, whose children are nodes $V4$ and $V5$. $V2$ is the spouse of $V1$ (and vice versa, $V1$ is the spouse of $V2$).}
\label{dag1}
\end{figure}

\subsection{Bayesian Networks}
Assume there is a collection $\mathbf{V}$ of $D$ variables whose joint distribution $P$ is known. 
The BN\footnote{BN is a special case of a DAG.} \citep{pearl1988,spirtes2000} $B = \langle G, \mathbf{V} \rangle$ arises from linking $P$ to $G$ through the Markov condition (or Markov property), which states that each variable is conditionally independent of its non-descendants given its parents. By using this condition, the joint distribution $P$ can be factorised as 

\begin{eqnarray} \label{markov} 
P(V_1, \dots, V_D) = \prod_{i=1}^D P\left(V_i | Pa(V_i)\right),
\end{eqnarray}
where Pa($V_i$) denotes the parents set of $V_i$ in $G$.

If $G$ entails only conditional independencies in $P$ and all conditional independencies in $P$ are entailed by $G$, based on the Markov condition, then $G$, $P$ and $G$ are faithful to each other, and $G$ is a perfect map of $P$ \citep{spirtes2000}. 

The BN whose edges can be interpreted causally is called causal BN, an edge $V_i \rightarrow V_j$ exists if $V_i$ is a direct cause of $V_j$. A necessary assumption made by the algorithms under study is causal sufficiency; there are no latent (hidden, non observed) variables among the observed variables $\mathbf{V}$. 

The triplet $(V_i, V_k, V_j)$ where $V_i \rightarrow V_k \leftarrow V_j$ is called V-structure. If there is no edge between $V_i$ and $V_j$ the node $V_k$ is called unshielded collider. In Figure \ref{dag1} the triplet $(V_1, V_3, V_2)$ is a V-structure as there is no edge between $V_1$ and $V_2$ and hence node $V_3$ is an unshielded collider. The unshielded collider $V_k$ implies that $V_i$ and $V_j$ ae independependent conditioning on $V_k$, provided that the faithfulness property holds true \citep{spirtes2000}. Conversely, the triplet of nodes $(V_i, V_k, V_j)$ such that $V_k \rightarrow V_i$ and $V_k \rightarrow V_j$ is termed $\Lambda$-structure (nodes $V_3, V_4$ and $V_5$ in Figure \ref{dag1} is such an example). The $\Lambda$-structure implies that $V_i$ and $V_j$ are conditionally independent given $V_k$.

Two or more BNs are called Markov equivalent if and only if they have the same skeleton and the same V-structures \citep{verma1991}. The set of all Markov equivalent BNs forms the Markov equivalence class that can be represented by a complete partially DAG, which in addition to directed edges contains undirected edges\footnote{Undirected edges may be oriented either way in BNs of the Markov equivalence class (in the set of all valid combinations), while directed and missing edges are shared among all equivalent BNs.}. 

\subsection{Classes of BN learning algorithms}
BN learning algorithms are typically constraint-based, score-based or hybrid. Constraint-based learning algorithms, such as PC\footnote{PC stands for Peter and Clark, named after Peter Spirtes and Clark Glymour, the names of the two researchers who invented it.} \citep{spirtes1991} and FCI \citep{spirtes2000} employ conditional independence (CI) tests to discover the structure of the network (skeleton), and then orient the edges by repetitively applying orientation rules. On the contrary, score-based methods \citep{cooper1992,heckerman1995,chickering2002}, assign a score on the whole network and perform a search in the space of BNs to identify a high-scoring network. Hybrid algorithms, such as MMHC \citep{tsamardinos2006} and PCHC \citep{tsagris2021}, combine both aforementioned methods; they first perform CI tests to discover the skeleton of the BN and then employ a scoring method to direct the edges in the space of BNs. 

\subsection{PCHC and MMHC algorithms}
PCHC's skeleton identification phase is the same as that of the PC algorithm \citep{tsagris2021}. The phase commences with all pairwise unconditional associations and removes the edge between ordered pairs which are not statistically significantly associated. Subsequently, CI tests are performed with the cardinality of the conditioning set (denoted by k) increasing by 1 at a time. At every step, the conditioning set consists of subsets of the neighbours, adjacent to each variable $V_i$ ($adj(G, V_i)$). This process is repeated until no edge can be removed. 

\cite{spirtes2000} suggested three heuristics to select the pairs of variables and the order is crucial as it can yield erroneous results. The optimal is, for a given variable $V_i$, to first test those variables ${\bf V}$ that are least probabilistically dependent on $V_i$, conditional on those subsets of variables that are most probabilistically dependent on $V_i$. Note that the pairs are first ordered according to the third heuristic of \cite{spirtes2000} and so the order of selection of the pairs is deterministic. Hence, the skeleton identification phase is independent of the order at which the variables are located in the dataset \citep{tsagris2019b}. 

MMHC's skeleton identification phase performs a variable selection process for each variable (call it target variable, $V_i$), described as follows. A search for its statistically significantly associated variables $V_s$ is performed via unconditional statistical tests. The associations are stored and the variable with the highest association ($V_j$) is chosen, an edge is added between this $V_i$ and $V_j$ and all non statistically significant variables are excluded from further consideration. In the second step, all CI tests between the target variable and previously identified variables, conditional on the previously selected variable are performed $\left( V_i\perp\!\!\!\perp V_m | V_j, \ \ m \neq i,j \right)$ and the non statistically significant variables are neglected. The previously stored associations are updated, for each variable, the minimum between the old and the new variables is stored. The variable with the highest association (Max-Min heuristic) is next selected. In subsequent steps, while the set of the selected variables increases, the conditioning set does not, as its cardinality is at most equal to $k$\footnote{This algorithm resembles the classical forward variable selection in statistics with two distinct differences. At each step, non significant variables are excluded from future searches and instead of conditioning on all selected variables. Secondly, the CI test for the next variable conditions upon all possible subsets, up to a pre-specified cardinality, of the already selected variables.}. Upon completion, a backward phase, in the same spirit as the forward applies to remove falsely detected variables.

This variable selection process is repeated for all variables. The edges detected remain only if they were identified by all variables. If for example, $V_j$ was found to associated with $V_i$, but $V_i$ was not found to be associated with $V_j$, then no edge between $V_i$ and $V_j$ will be added. 

A slightly modified version of MMHC's skeleton identification phase is implemented in the \textit{R} package \textit{pchc}. The backward phase is not performed in order to make the algorithm faster. To distinguish between them, \textit{bnlearn}'s implementation will be denoted by MMHC-1 and \textit{pchc}'s implementation will be denoted by MMHC-2 hereafter.

The orientation of the discovered edges takes place in the second, Hill Climbing (HC) scoring, phase of PCHC and MMHC and is the same phase employed by FEDHC as well.

\section{The FEDHC BN learning algorithm} \label{fedhc}
Similarly to PCHC and MMHC, the skeleton identification phase of FEDHC relies on a variable selection algorithm. Thus, prior to introducing the FEDHC algorithm the Forward Backward with Early Dropping (FBED) variable selection algorithm \citep{borboudakis2019} is briefly presented.

\subsection{The FBED variable selection algorithm}
In the classical forward selection algorithm all available predictor variables are constantly used and their statistical significance is tested at each step. Assuming that out of $10,000$ predictor variables only $10$ are selected. This implies that almost $10,000 \times 10$ regression models must be fitted and the same amount of statistical tests must be executed. The computational cost is tremendous rendering this computationally expensive algorithm impractical and hence prohibitive. 

\cite{borboudakis2019} introduced the FBED algorithm as a speed-up modification of the traditional forward selection algorithm coupled with the backward selection algorithm \citep{draper1998}. FBED relies on the Early Dropping heuristic to speed up the forward selection. The heuristic drops the non statistically significant predictor variables at each step, thus removes them from further consideration resulting in a computationally dramatically cheaper algorithm, that is presented in Algorithm \ref{algorithm1}. 

\makeatletter
\def\BState{\State\hskip-\ALG@thistlm}
\makeatother

\begin{algorithm}
\begin{algorithmic}[1]
\State \textbf{Input}: A response variable $y$ and a set of $D$ predictor variables $\bf V$.
\State Let ${\bf S}=\emptyset$ denote the set of selected variables. 
\State Perform all regression models of $y$ on each $V_i$, $i=1,\ldots,D$, $y \sim f(V_i)$, where $f$ denotes 
a function of $V_i$, e.g. a linear model $y=a+bV_i + e$, and retain only the statistically 
significant predictor variables ${\bf V}_{sig}$. 
\State Choose $V_j$ from ${\bf V}_{sig}$ that has the highest association, add it in
${\bf S}$ and use that to 
perform all regression models of $y$ on the $V_j$ and each $V_{\ell}$, $y\sim f(V_j,V_{\ell})$, where $\ell \in {\bf V}$, 
with $\ell \neq j$ and again retain only the statistically significant predictor variables, thus 
reducing $| {\bf V}_{sig}|$ and increasing $| {\bf S}|$. 
\State Repeat until no predictor variable is left, i.e. ${\bf V}_{sig}=\emptyset$. 
\State This process can be repeated $k$ times, using all neglected predictor variables, where $k$ is a pre-defined number, until $|{\bf S}|$ cannot further increase. 
\State Perform a backward selection phase attempting to remove the non statistically significant 
predictor variables. 
\State \textbf{Return} ${\bf S}$.
\end{algorithmic}
\caption{The FBED variable selection algorithm} \label{algorithm1}
\end{algorithm}

\subsection{Skeleton identification phase of the FEDHC algorithm}
The skeleton identification phase of the FEDHC algorithm is the one presented in Algorithm \ref{algorithm2}, but it must be stressed that the backward phase of FBED is not performed so as to reduce the computational cost. The FBED algorithm (Algortihm \ref{algorithm1}) for each variable (call it target variable, $V_i$). This variable selection process is repeated for all variables. The edges detected remain only if they were identified by all variables. If for example, $V_j$ was found to associated with $V_i$, but $V_i$ was not found to be associated with $V_j$, then no edge between $V_i$ and $V_j$ will be added.

\makeatletter
\def\BState{\State\hskip-\ALG@thistlm}
\makeatother

\begin{algorithm}[H]
\centering
\begin{algorithmic}[1]
\State \textbf{Input}: Data set on a set of $D$ variables $\bf V$.
%\text{Form the complete undirected graph G on the variables set $\bf V$.} \\
\State Let the adjacency matrix $G$ be full of zeros. 
\State \textbf{Repeat} for all variables $V_i$, $i=1,\ldots,D$ 
\State Perform the FBED algorithm in Algorithm \ref{algorithm1}, excluding the backward phase, and return \hskip 0.4cm ${\bf S}_i$. 
\State Set $G_{ij}=1$ for all $j \in {\bf S}_i$. 
\State \textbf{If} $G_{ij} \neq G_{ji}$ set $G_{ij}=G_{ji}=0$. 
\State \textbf{Return} $G$. 
\end{algorithmic}
\caption{Skeleton identification phase of the FEDHC algorithm} \label{algorithm2}
\end{algorithm}

\subsection{Hill Climbing phase of the FEDHC algortihm}
The first phase of FEDHC, MMHC and of PCHC is to discover any possible edges between the nodes using CI tests. In the second phase, a search for the optimal DAG is performed, where edges turn to arrows or are deleted towards maximisation of a score metric. This scoring phase performs a greedy HC search in the space of BNs, commencing with an empty graph \citep{tsamardinos2006}. The edge deletion or direction reversal that leads to the largest increase in score, in the space of BNs\footnote{This implies that every time an edge removal, or arrow direction is implemented, a check for cycles is performed. If cycles are created, the operation is cancelled regardless if it increases the score.}, is applied and the search continues recursively. The fundamental difference from standard greedy search is that the search is constrained to the orientation of the edges discovered by the skeleton identification phase\footnote{For more information see \cite{tsamardinos2006}.}.

Tabu search is such an iterative local searching procedure adopted by \cite{tsamardinos2006} for this purpose. Its performance is enhanced by using a list where the last 100 structures explored are stored, while searching in the neighborhood of each solution. The search is also capable of escaping from local optima, in which normal local search techniques often get stuck. Instead of applying the best local change, the best local change that results in a structure not on the list is performed in an attempt to escape local maxima \citep{tsamardinos2006}. This change may actually reduce the score. When a number of changes (10-15) occur without an increase in the maximum score ever encountered during search, the algorithm terminates. The overall best scoring structure is then returned. 

The Bayesian Information Criterion (BIC) \citep{schwarz1978} is a frequent score used for continuous data, while other options include the multivariate normal log-likelihood, the Akaike Information Criterion (AIC) and the Bayesian Gaussian equivalent\footnote{The term "\textit{equivalent}" refers to their attractive property of giving the same score to equivalent structures (Markov equivalent BNs) i.e., structures that are statistically indistinguishable \citep{tsamardinos2006}.} \citep{geiger1994} score. The Bayesian Dirichlet equivalent (BDE) \citep{buntine1991}, the BDe uniform score (BDeu) \citep{heckerman1995}, the multinomial log-likelihood score \citep{bouckaert1995} and the MDL score \citep{suzuki1993,lam1994} are four scoring options for discrete data. 

The combination of the FBED algorithm during the skeleton identification phase with the HC scoring method forms the FEDHC algorithm. Interestingly enough, the skeleton identification phase of FEDHC performs substantially fewer statistical tests than PCHC and MMHC. 

\subsection{Prior knowledge} \label{prior}
All BN learning algorithms are agnostic of true relationships among the input data. It is customary though for practitioners and researchers to have prior knowledge of the necessary directions (forbidden or not) of some of the relationships among the variables. For instance, variables such as sex or age cannot be caused by any economic or demographic variables. Economic theory (or theory from any other field) can further assist in improving the quality of the fitted BN by imposing or forbidding directions among some variables. All the prior information can be inserted into the scoring phase of the aforementioned BN learning algorithms leading to less errors and more realistic BNs.  

\subsection{Theoretical properties of FEDHC}
The theoretical properties and guarantees of MMHC and PCHC can be found in \cite{tsamardinos2006} and \cite{tsagris2021}, respectively. As for the FEDHC, while there is no theoretical guarantee of the skeleton identification phase of FEDHC, \cite{borboudakis2019} showed that running FBED with two repeats recovers the MB of the response variable if the joint distribution of the response and the predictor variables can be faithfully represented by a BN. When used for BN learning though, FBED need not be be run more than once for each variable. In this case FBED, similarly to MMHC, will identify the children and parents of a variable $V_i$, but not the spouses of the children \citep{borboudakis2019} as this is not necessary during the skeleton identification phase. When FBED is run on the children of the variable $V_i$ it will again identify the children's parents who are the spouses of the variable $V_i$. Hence, upon completion of the FEDHC algorithm will have identified the MB of each variable. 

Additionally, the early dropping heuristic does not only reduce the computational time but also reduces the problem of multiple testing, in some sense \citep{borboudakis2019}. When FBED is run only once (as in the current situation), in the worst-case scenario, it is expected to select about $\alpha \cdot D$ variables (where $\alpha$ is the significance level) since all other variables will be dropped in the first (filtering) phase. However, their simulation studies showed that FBED was selecting fewer false positives than expected and the authors' recommendation is to reduce the number of runs to further limit the number of falsely selected variables, a strategy FEDHC follows by default.

Similar to MMHC, the FEDHC is a local learning algorithm, and hence during the HC phase the overall score is decomposed \citep{tsamardinos2006} exploiting the Markov property of BNs (\ref{markov}). The local learning has several advantages (see \cite{tsamardinos2006}) and the scores (BDe, BIC., etc.) are locally consistent \citep{chickering2002}.

\subsection{Robustification of the FEDHC algorithm for continuous data}
It is not uncommon for economic datasets to contain outliers, observations with values far from the rest of the data. Income is such an example that contains outliers, but if outliers appear only in that variable their effect will be minor. The effect of outliers is propagated when they exist in more variables and in order to mitigate their effect, they must be identified in the multivariate space. If these outliers are not detected or not treated properly, BN learning algorithms will yield erroneous results. FEDHC will employ the Reweighted Minimum Covariance Determinant (RMCD) \citep{rousseeuw1985,rousseeuw1999} as a means to identify outliers and remove them\footnote{The reason why one cannot use the robust correlation matrix directly is because the independence property between two variables no longer holds true. The robust correlation between any two variables depends on the other variables, so adding or removing a variable modifies the correlation structure \citep{raymaekers2021}.}.

The RMCD estimator is computed in two stages. In the first stage, a subset of observations $h$ ($n/2 ≤ h < n$) is selected such that the covariance matrix has the smallest determinant and a robust mean vector is also computed. The second stage, is a re-weighting scheme that increases the efficiency of the estimator, while preserving its high-breakdown properties. A weight $w_i=0$ is given to observations whose first-stage robust distance exceeds a threshold value, otherwise the weight is $w_i = 1$ ($i=1,\ldots,n$) is given. Using the re-weighted robust covariance matrix and mean vector, robust Mahalanobis distances are computed $d_{i(RMCD)}^2=\left({\bf x}_i - \tilde{\pmb{\mu}}_{(RMCD)}\right)^T\tilde{\pmb{\Sigma}}^{-1}_{(RMCD)}\left({\bf x}_i - \tilde{\pmb{\mu}}_{(RMCD)}\right)$ and proper cut-off values are required to detect the outliers. Those cut-off values are based on the following accurate approximations \citep{cerioli2010,cerchiello2016}
\begin{eqnarray*}
d_{i(RMCD)}^2 \begin{array}{ccc}
\sim & \frac{\left(w-1\right)^2}{w}Be\left(\frac{D}{2}, \frac{w-D-1}{2}\right) & \text{if} \ \ w_i=1 \\
\sim & \frac{w+1}{w}\frac{(w-1)D}{w-D}F_{D,w-D}& \text{if} \ \ w_i=0,
\end{array}
\end{eqnarray*}
where $w=\sum_{i=1}^nw_i$, and $Be$ and $F$ denote the Beta and F distributions respectively. 

The observations whose Mahalananobis distance $d_{i(RMCD)}^2$ exceeds the $97.5\%$ quantile of either distribution ($Be$ or $F$) are considered to be outliers and are hence removed from the dataset. The remainder of the dataset, assumed to be outlier free, will be used by FEDHC to learn the BN. 

The default value for $h$ is $[(n+p+1)/2]$, where $[.]$ denotes the largest integer. This value was proven to have the highest breakdown point \citep{hubert2010}, but low efficiency on the other hand. Changing $h$ yields an estimator with lower robustness properties and increases the computational cost of the RMCD estimator. For these reasons, this is the default value used inside the robustification process of the FEDHC algorithm.

The case of $n < p$ and $n \ll p$ (very high dimensional case) in general can be treated in a similar way, by replacing the RMCD estimator with the high dimensional MCD approach of \cite{ro2015}. 

\section{Monte Carlo simulation studies} \label{mc}
Extensive experiments were conducted on simulated data to investigate the quality of estimation of FEDHC compared to PCHC and MMHC-2. MMHC-1 participated in the simulation studies only with categorical data and not with continuous data because it is prohibitively expensive. The continuous data were generated by synthetic BNs that contained a various number of nodes, $p = (20, 30, 50)$, with an average of 3 and 5 neighbours (edges) for each variable. For each case 50 random BNs were generated with Gaussian data and various sample sizes. Categorical data were generated utilising two famous (in the BN learning community) realistic BNs and the sample sizes were left varying. The \textit{R} packages \textit{MXM} \citep{tsagris2019a} and \textit{bnlearn} were used for the BN generation, and the \textit{R} packages \textit{pchc} and \textit{bnlearn} were utilised to apply the FEDHC, PCHC, MMHC-2 and the MMHC-1 algorithms respectively. All simulations were implemented in a desktop computer with Intel Core i5-9500 CPU at 3.00GHz with 48 GB RAM and SSD installed. 

The metrics of quality of the learned BNs were the structural Hamming distance (SHD) \citep{tsamardinos2006} of the estimated DAG from the true DAG\footnote{This is defined as the number of operations required to make the estimated graph equal to the true graph. Instead of the true DAG, the Markov equivalence graph of the true BN; that is some edges have been un-oriented as their direction cannot be statistically decided. The transformation from the DAG to the Markov equivalence graph was carried out using Chickering's algorithm \citep{chickering1995}.}, the computational cost and the number of tests performed during the skeleton identification phase and the total duration of the algorithm. PCHC and the MMHC-1 algorithms have been implemented in \textit{C++}, henceforth the comparison of the execution times is not really fair at the programming language level. FEDHC and MMHC-2 have been implemented in \text{R} (skeleton identification phase) and \textit{C++} (scoring phase).

\subsection{Synthetic BNs with continuous data} \label{datgen}
The procedure used to generate data for $X$ is summarised in the steps below. Let $X$ be a variable in $G$ and $Pa(X)$ be the parents of $X$ in $G$.
\begin{enumerate}
\item Sample the coefficients $\beta$ of $f\left(Pa(X)\right)$ uniformly at random from $[-1,-0.1] \cup [0.1,1]$.
\item In case $Pa(X)$ is empty, $X$ is sampled from the standard normal distribution. Otherwise, $X = f\left(Pa(X)\right) = \beta_0 + \sum_i \beta_i Pa_i(X) + \epsilon_X$, a linear function\footnote{In general this can represent any (non-linear) function.} depending on $X$, where $\epsilon_X$ is generated from a standard normal distribution.
\end{enumerate}

The average number of connecting edges (neighbours) was set to 3 and 5. The higher the number of edges is, the denser the network is and the harder the inference becomes. The sample sizes considered were $n = (100, 500, 1,000, 5000,  1\times 10^4, 3\times 10^4, 5\times 10^4, 1\times 10^5, 3\times 10^5, 5\times 10^5, 1\times 10^6, 3\times 10^6, 5\times 10^6)$.

\begin{figure}[!ht]
\centering
\begin{tabular}{cc}
\includegraphics[scale = 0.38, trim = 70 0 0 0]{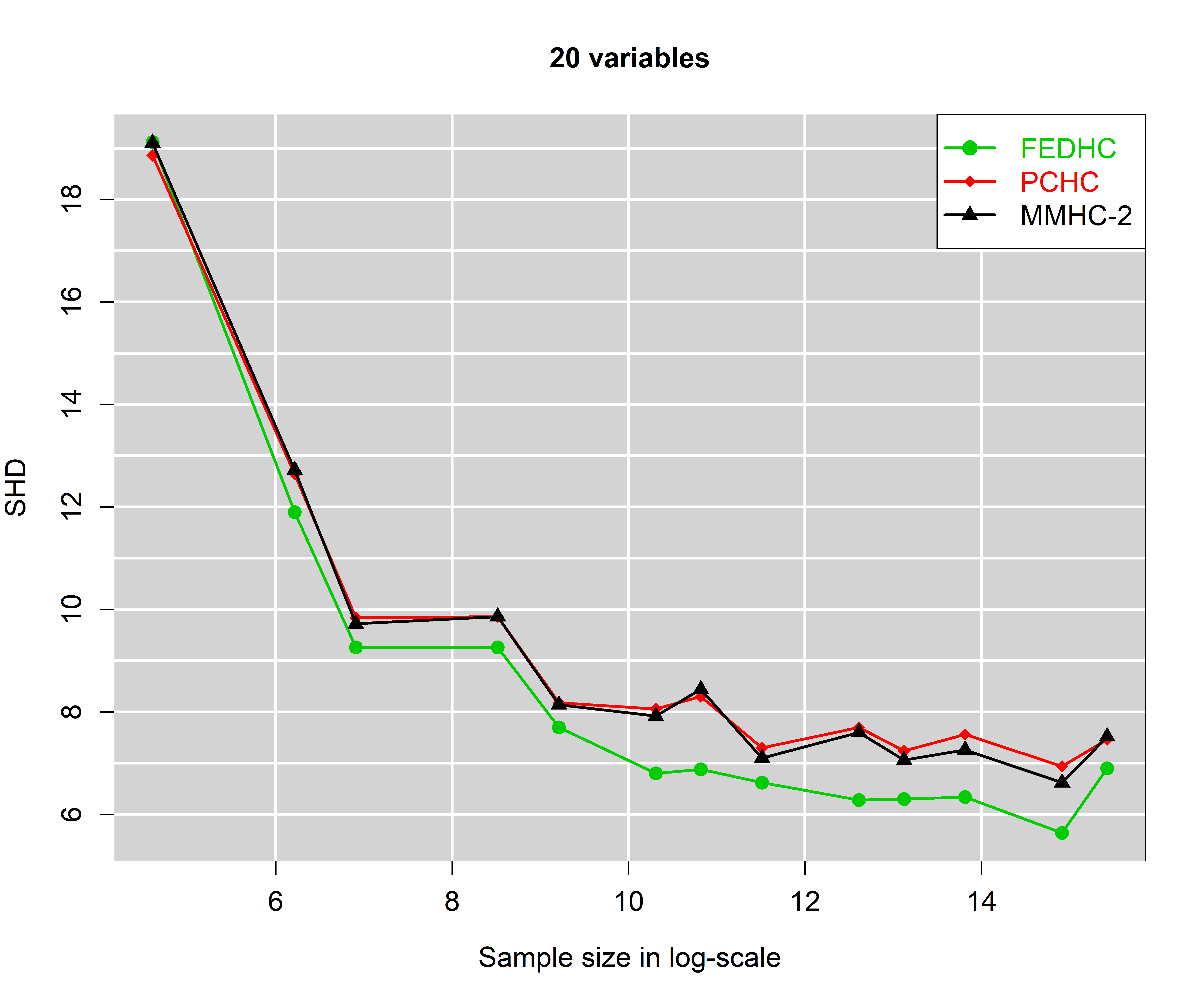}  &
\includegraphics[scale = 0.38, trim = 50 0 0 0]{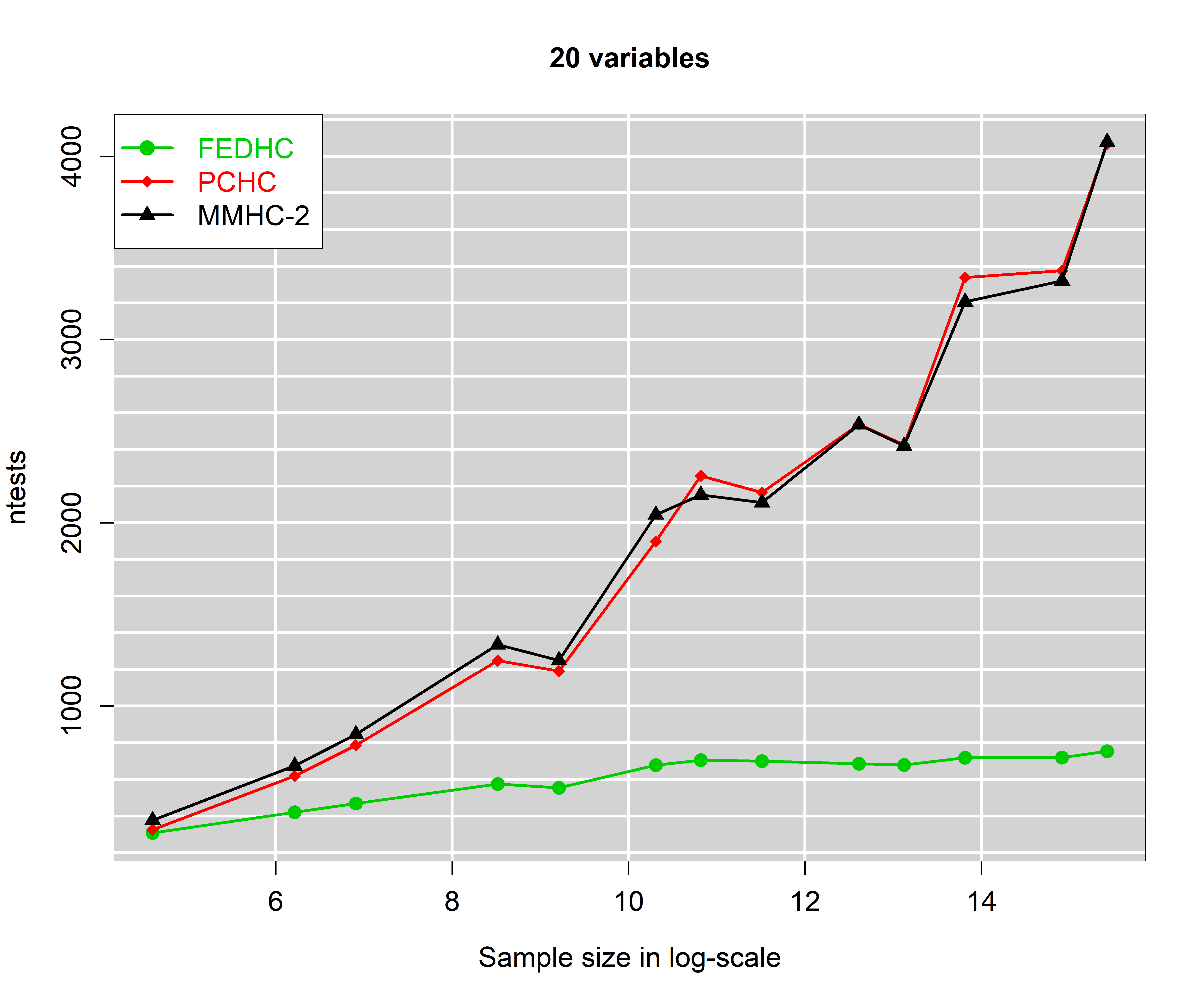}  \\
(a) SHD vs log of sample size. &  (b) Number of CI tests vs log of sample size.  \\
\includegraphics[scale = 0.38, trim = 70 0 0 0]{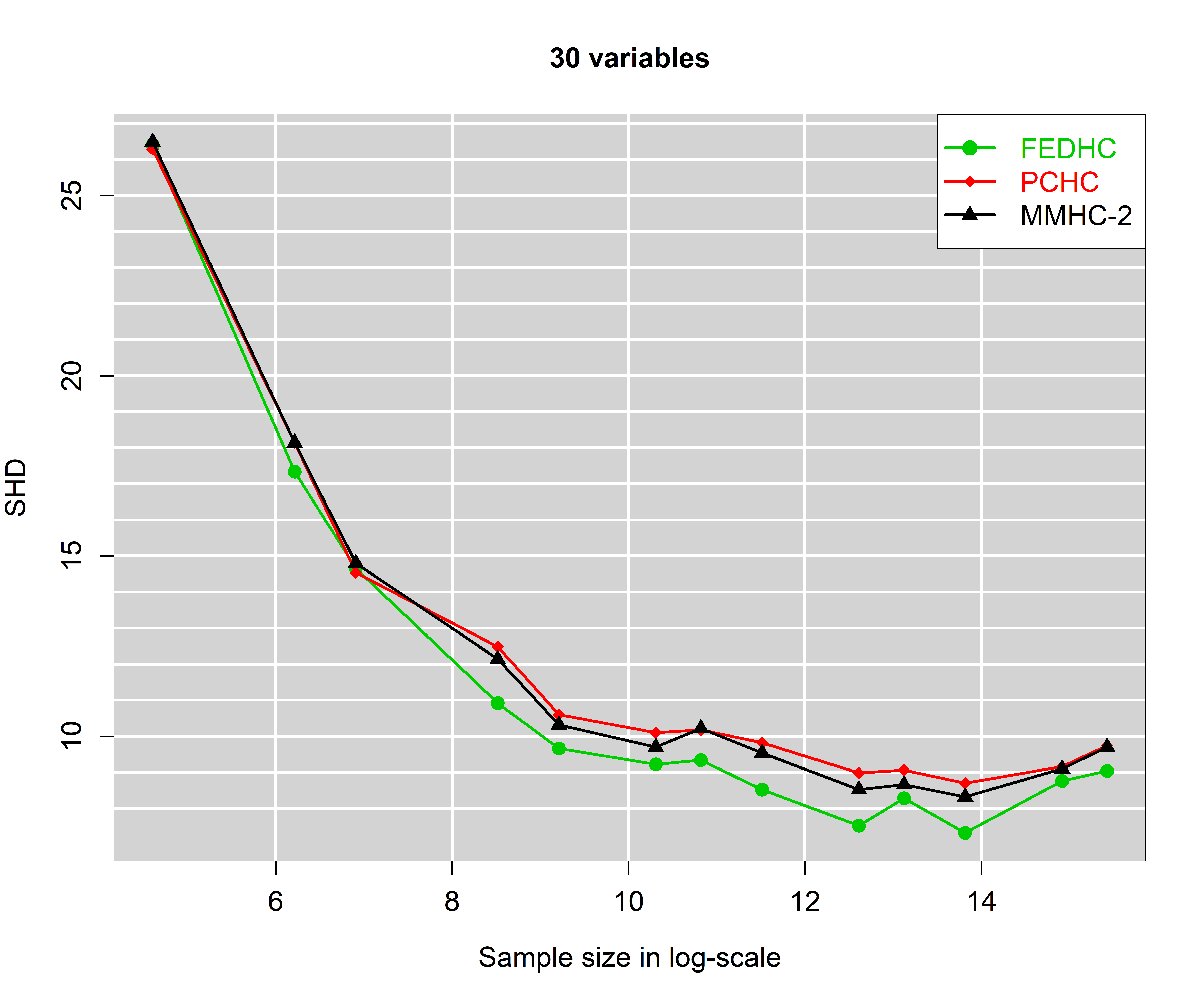}  &
\includegraphics[scale = 0.38, trim = 50 0 0 0]{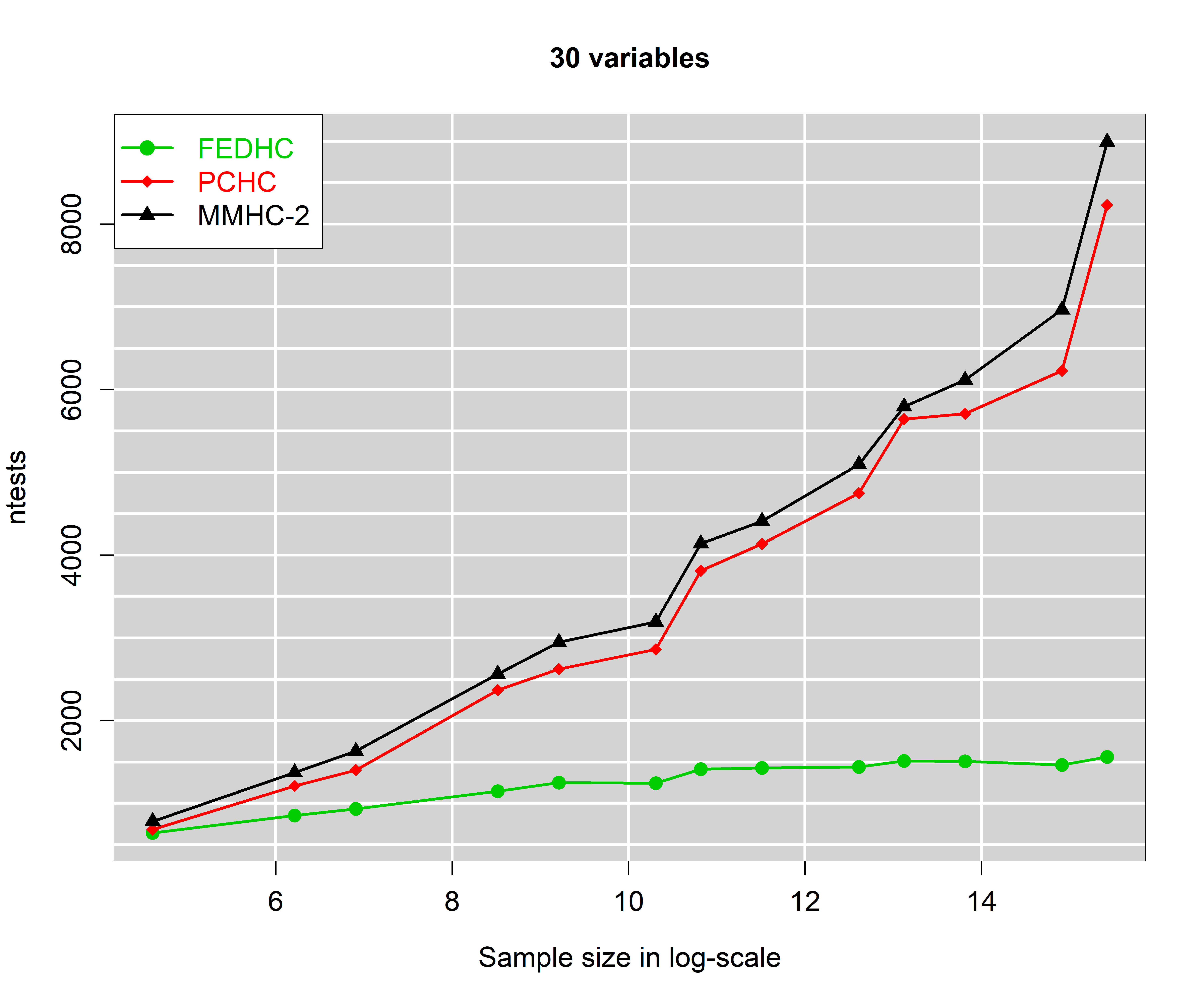}  \\
(c) SHD vs log of sample size. &  (d) Number of CI tests vs log of sample size.  \\
\end{tabular}
\caption{SHD and number of CI tests against log of sample size for 20 and 30 dimensions with \textbf{3 neighbours} on average. \label{synthetic_3a} }
\end{figure}

\begin{figure}[!ht]
\centering
\begin{tabular}{cc}
\includegraphics[scale = 0.38, trim = 70 0 0 0]{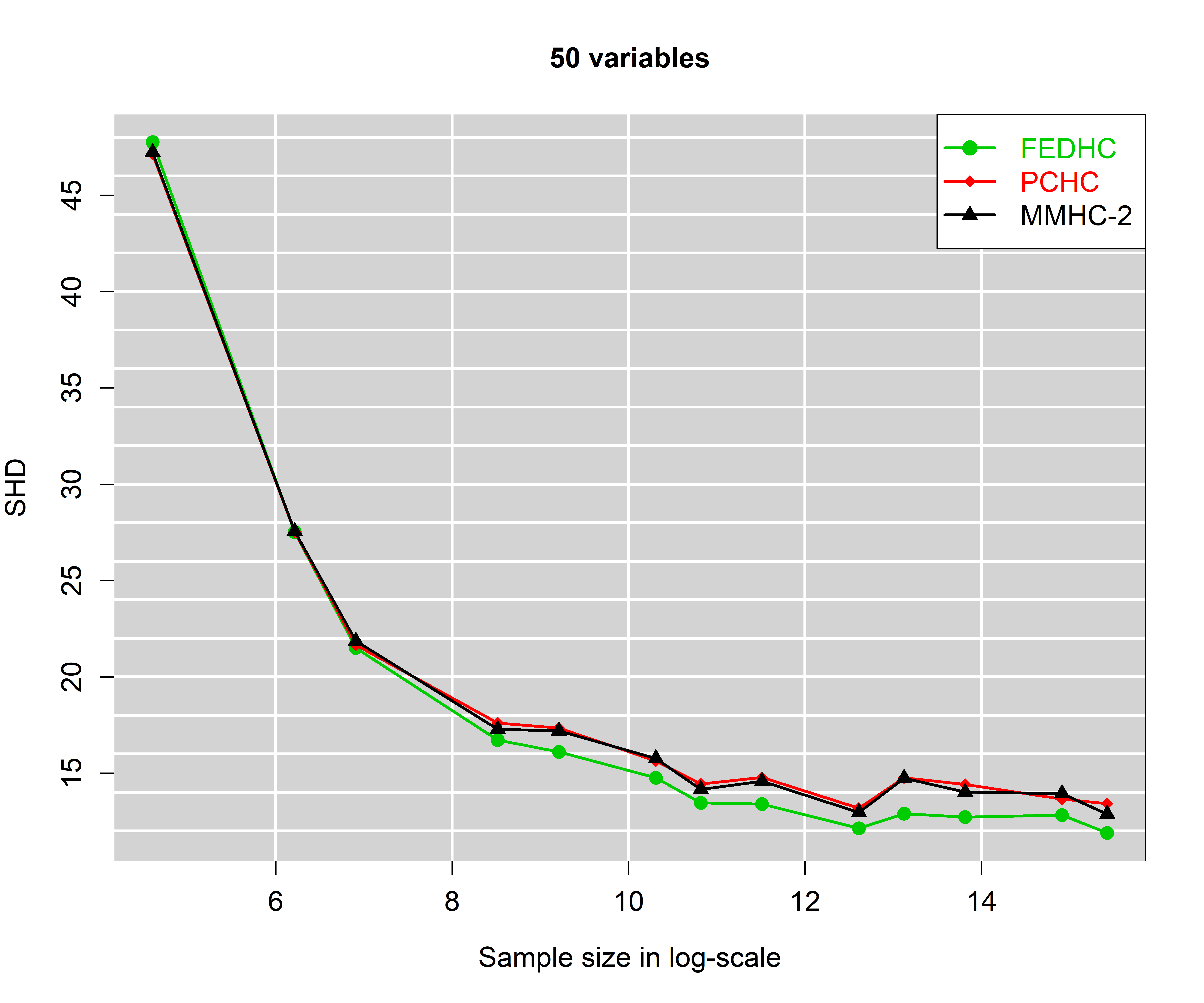}  &
\includegraphics[scale = 0.38, trim = 50 0 0 0]{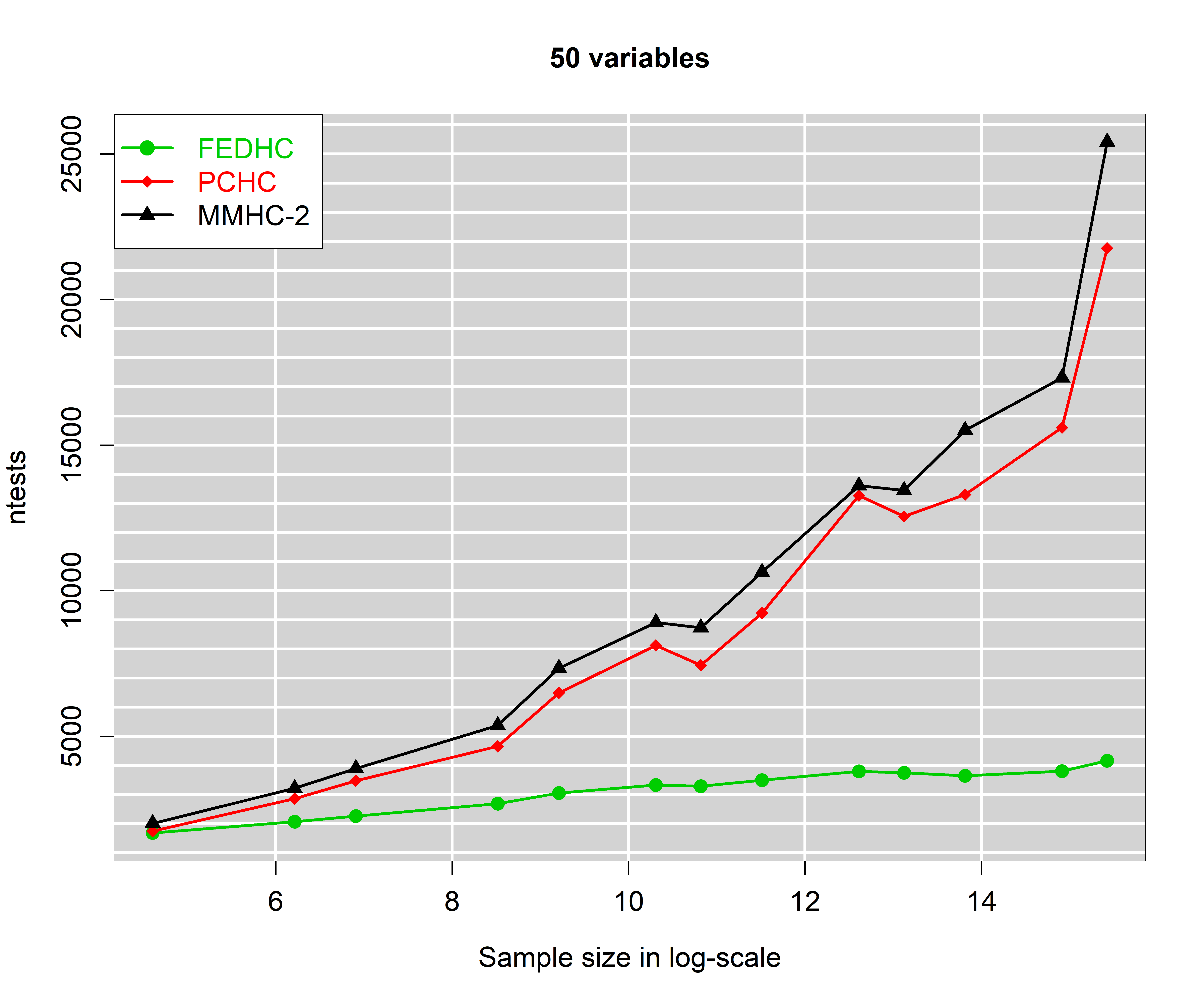}  \\
(a) SHD vs log of sample size. &  (b) Number of CI tests vs log of sample size.  \\
\includegraphics[scale = 0.38, trim = 70 0 0 0]{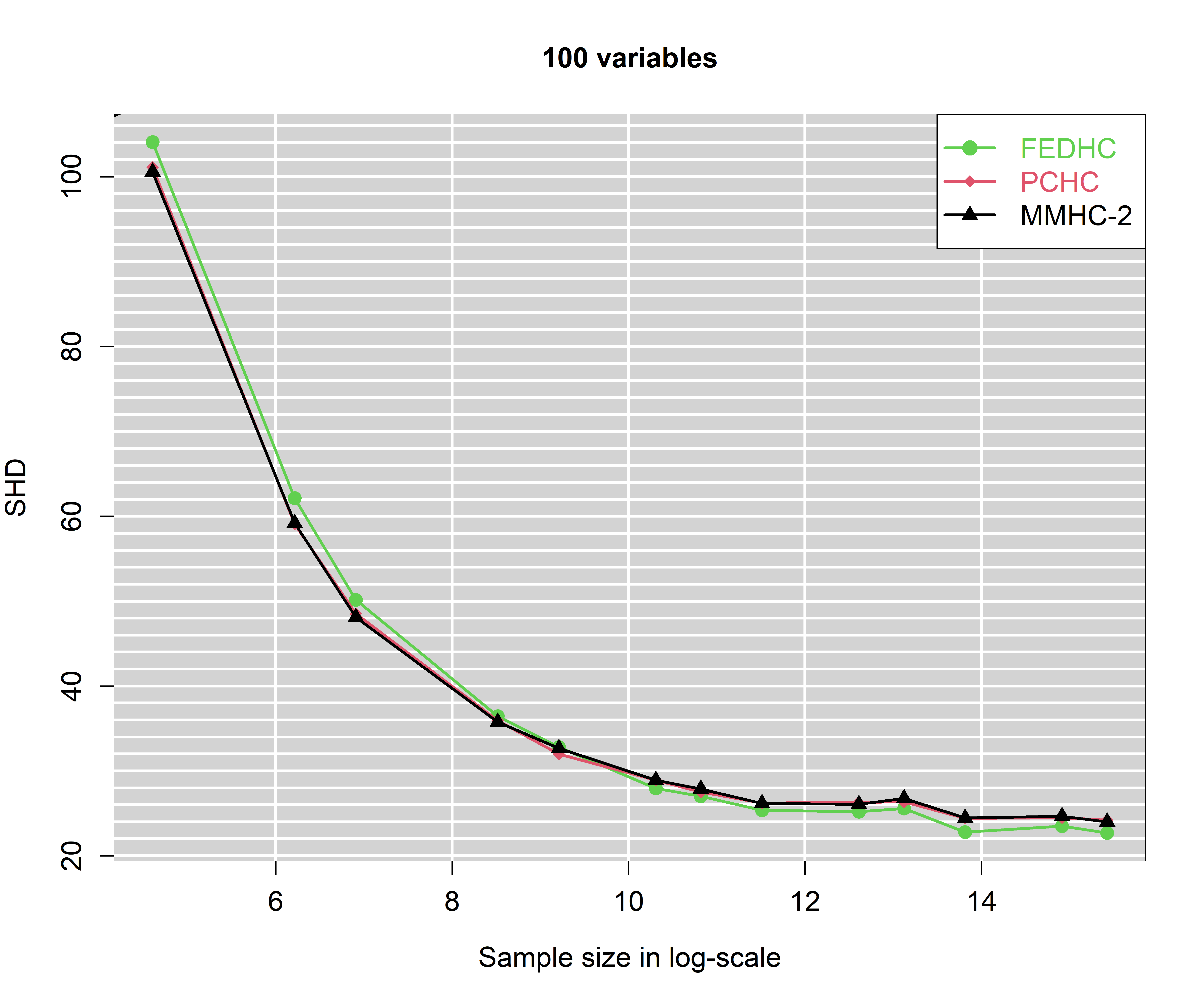}  &
\includegraphics[scale = 0.38, trim = 50 0 0 0]{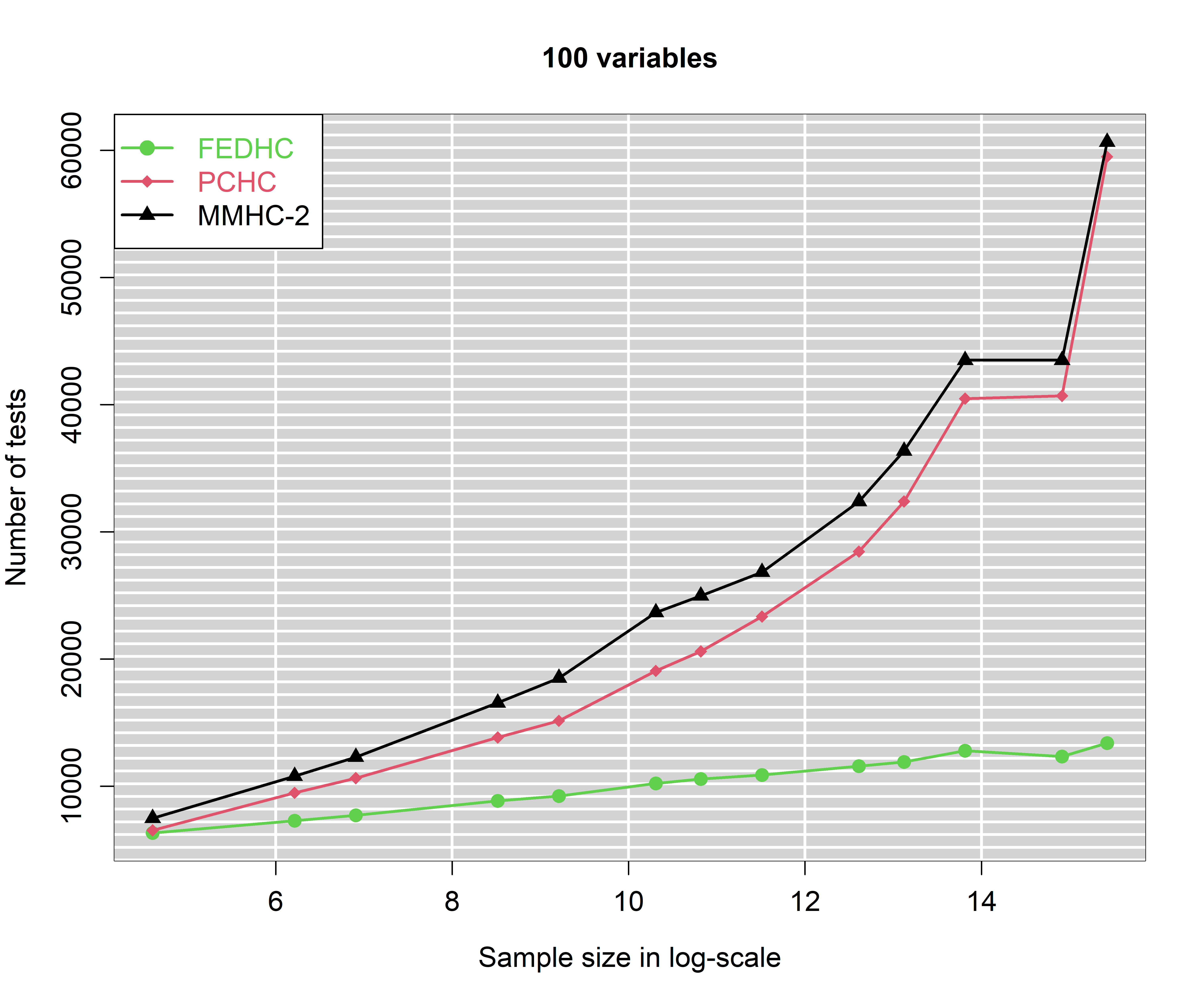}  \\
(c) SHD vs log of sample size. &  (d) Number of CI tests vs log of sample size.  \\
\end{tabular}
\caption{SHD and number of CI tests against log of sample size for 50 and 100 dimensions with \textbf{3 neighbours} on average. \label{synthetic_3b} }
\end{figure}

\begin{figure}[!ht]
\centering
\begin{tabular}{cc}
\includegraphics[scale = 0.38, trim = 70 0 0 0]{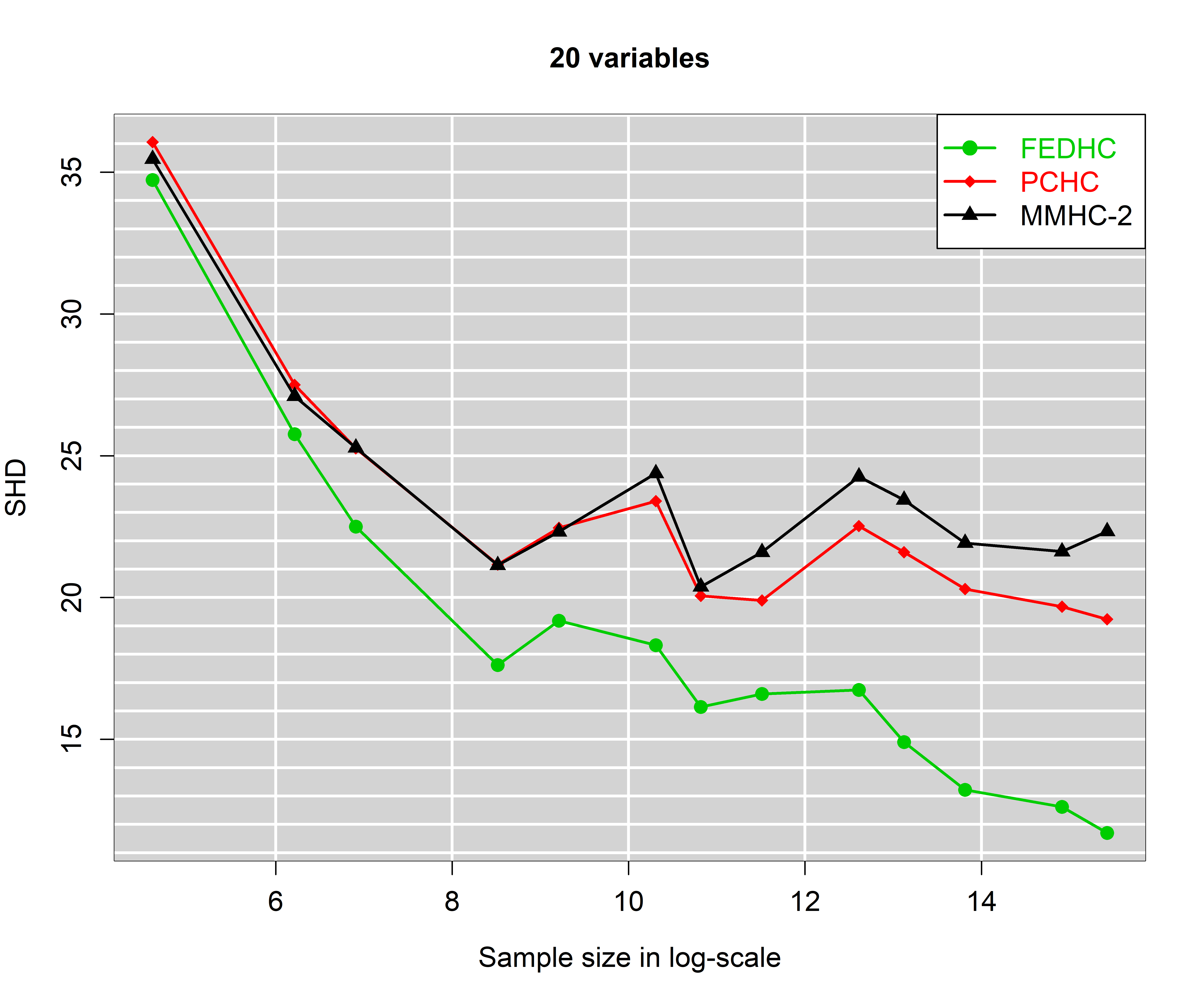}  &
\includegraphics[scale = 0.38, trim = 50 0 0 0]{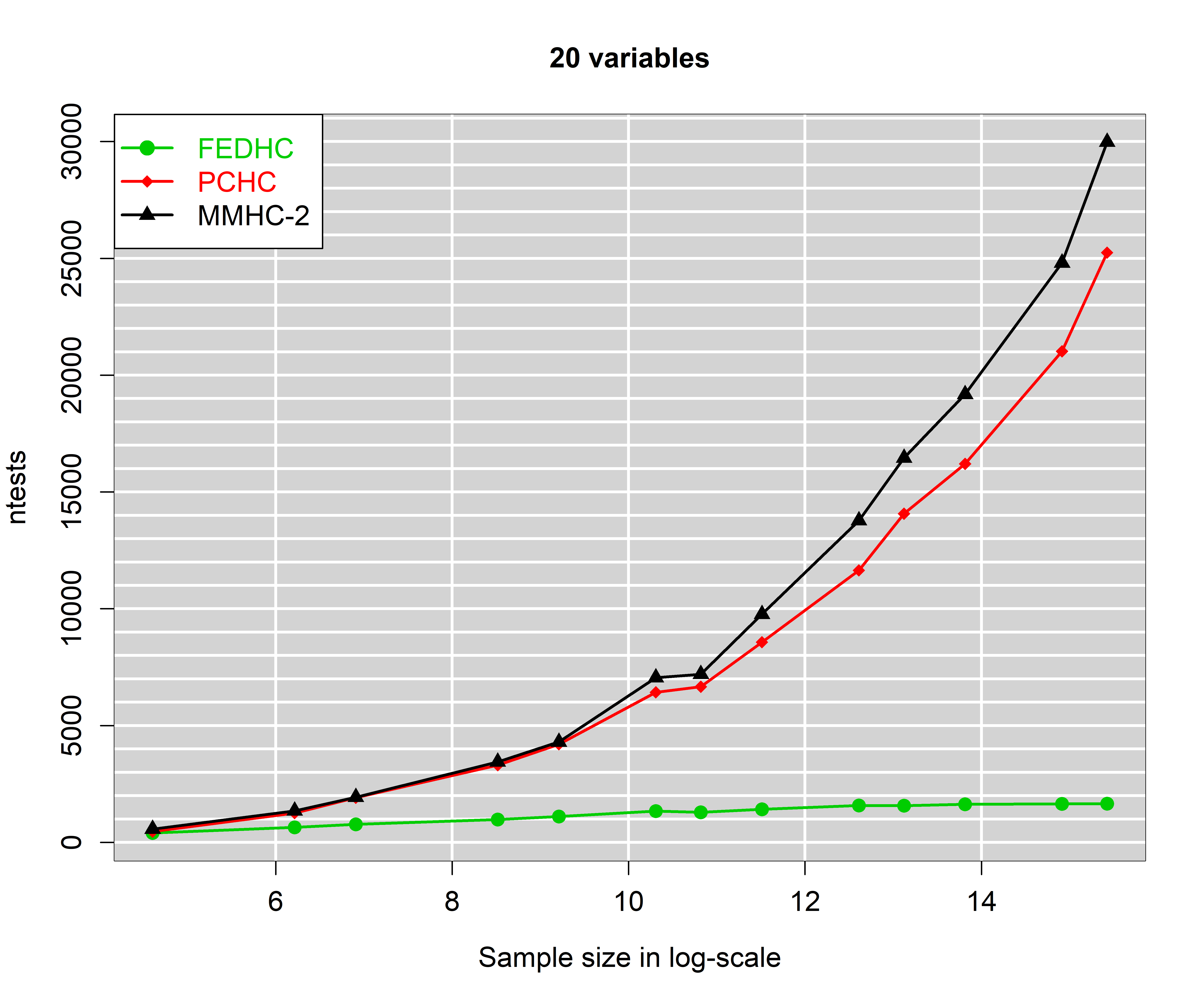}  \\
(a) SHD vs log of sample size. &  (b) Number of CI tests vs log of sample size.  \\
\includegraphics[scale = 0.38, trim = 70 0 0 0]{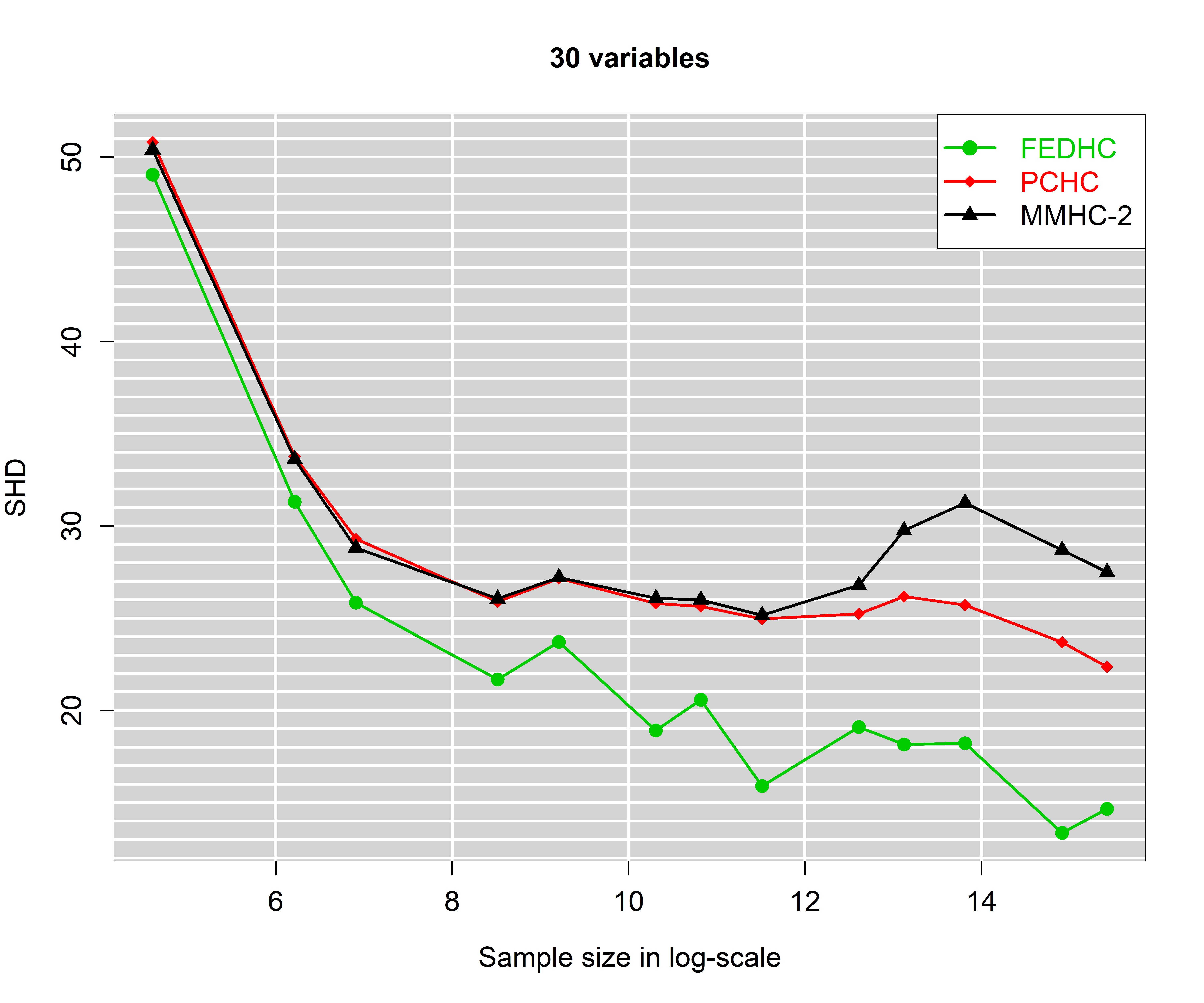}  &
\includegraphics[scale = 0.38, trim = 50 0 0 0]{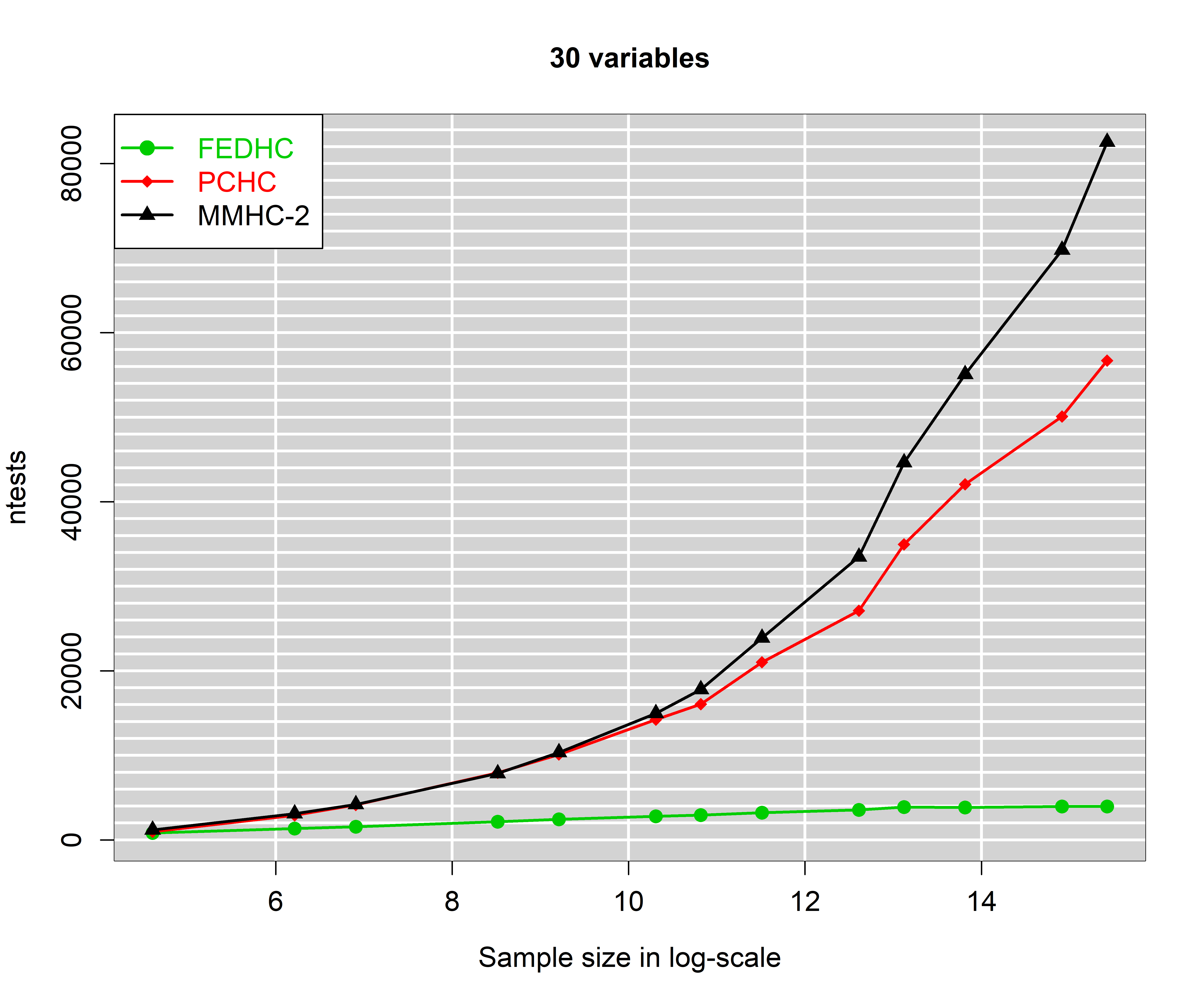}  \\
(c) SHD vs log of sample size. &  (d) Number of CI tests vs log of sample size.  \\
\end{tabular}
\caption{SHD and number of CI tests against log of sample size for 20 and 30 dimensions with \textbf{5 neighbours} on average. \label{synthetic_5a} }
\end{figure}

\begin{figure}[!ht]
\centering
\begin{tabular}{cc}
\includegraphics[scale = 0.38, trim = 70 0 0 0]{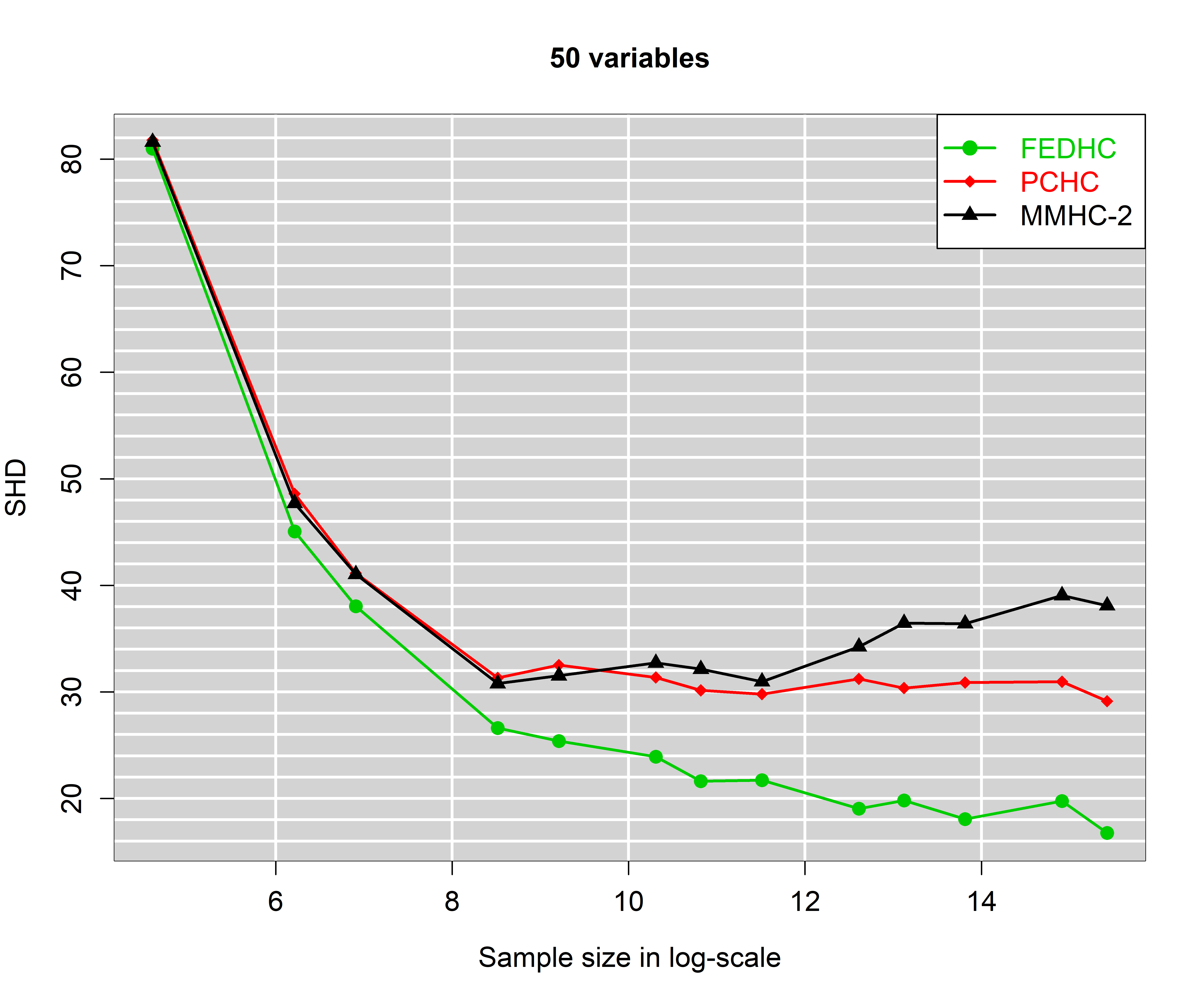}  &
\includegraphics[scale = 0.38, trim = 50 0 0 0]{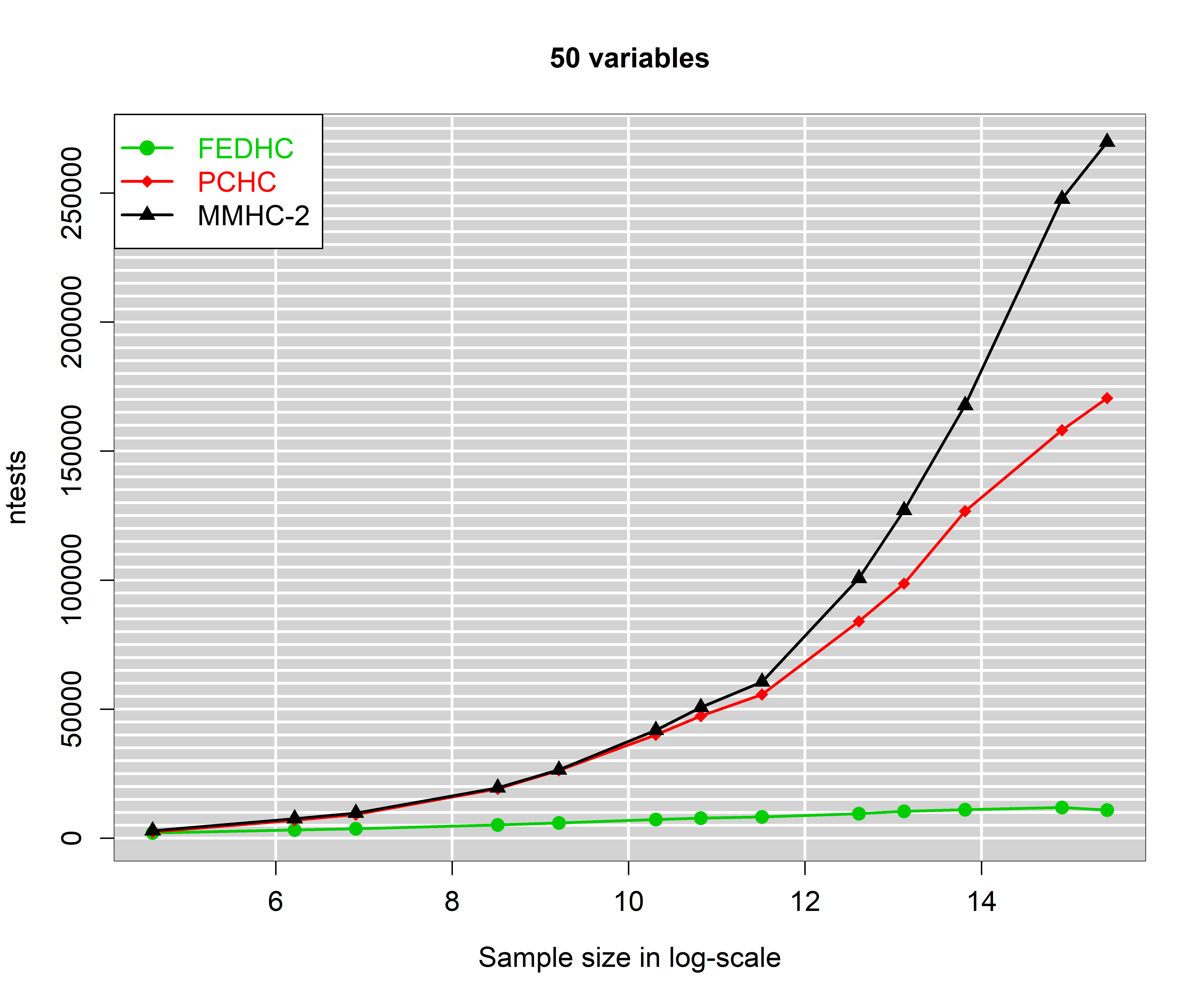}  \\
(a) SHD vs log of sample size. &  (b) Number of CI tests vs log of sample size.  \\
\includegraphics[scale = 0.38, trim = 70 0 0 0]{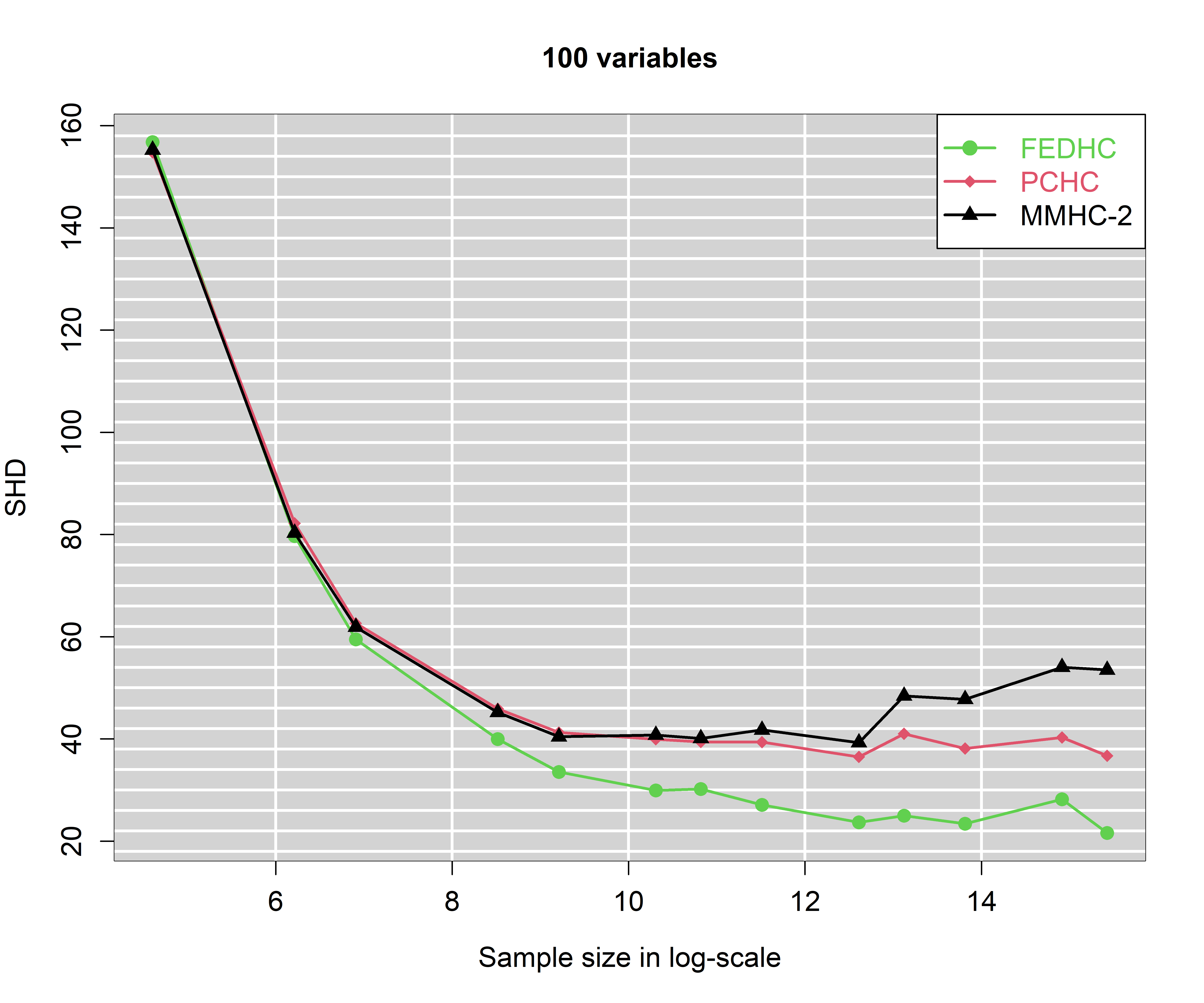}  &
\includegraphics[scale = 0.38, trim = 50 0 0 0]{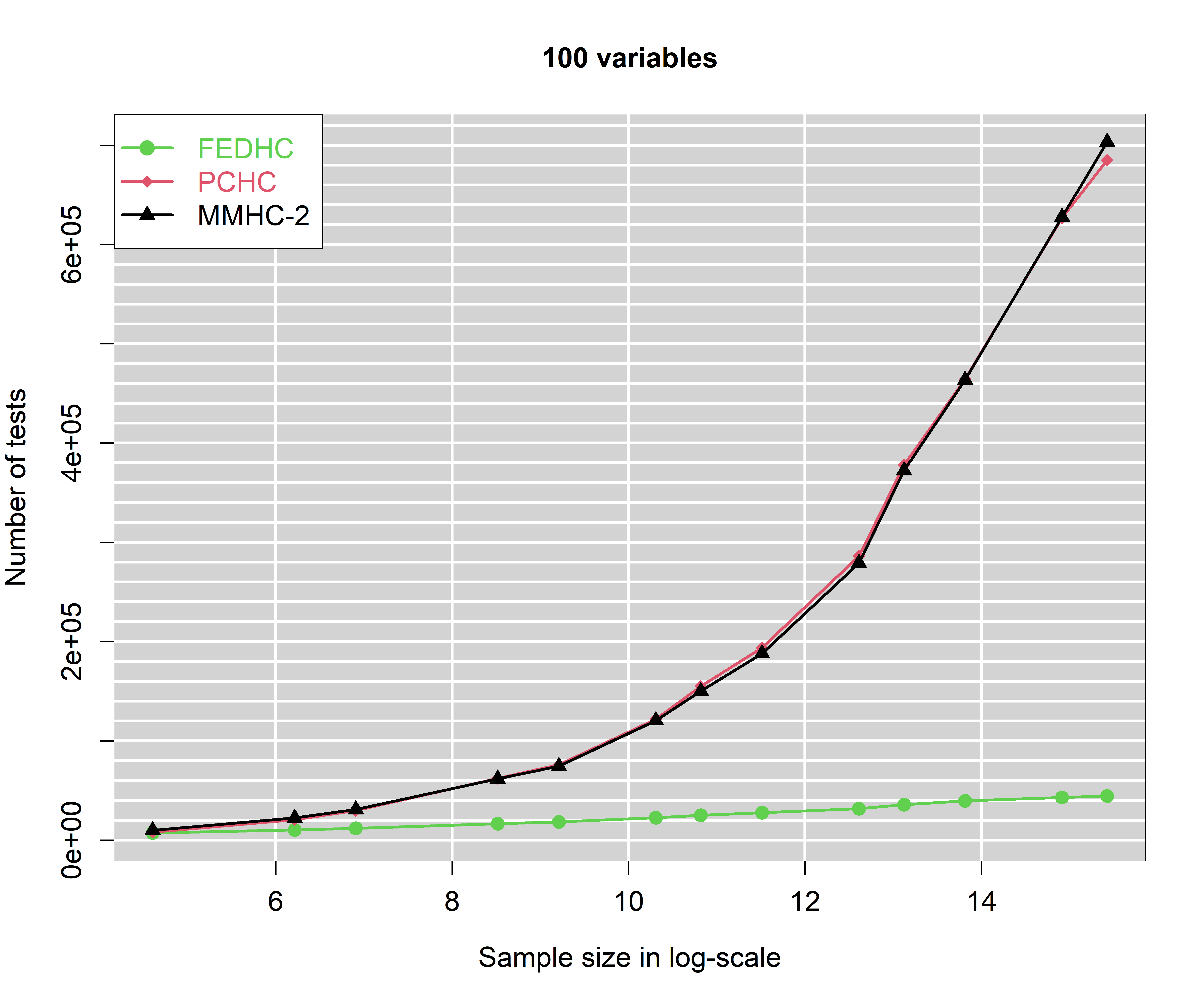}  \\
(c) SHD vs log of sample size. &  (d) Number of CI tests vs log of sample size.  \\
\end{tabular}
\caption{SHD and number of CI tests against log of sample size for various dimensions with \textbf{5 neighbours} on average. \label{synthetic_5b} }
\end{figure}

Figures \ref{synthetic_3a}, \ref{synthetic_3b}, \ref{synthetic_5a} and \ref{synthetic_5b} present the SHD and the number of CI tests performed of each algorithm for a range of sample sizes (in log-scale). With 3 neighbours on average per node, the differences in the SHD are noticeably rather small, yet FEDHC achieves lower numbers. With 5 neighbours on average though the differences are more significant and increasing with increasing sample sizes. As for the number of CI tests executed during the skeleton identification phase, FEDHC is the evident winner as it executes up to 6 times less tests, regardless of the average neighbours.  

\subsection{Robustified FEDHC for continuous data}
Examination of the robustified FEDHC contains two axes of comparison; the outlier-free and the outliers present cases. At first the performances of the raw and the robustified FEDHC in the outlier-free case are evaluated. 

\begin{figure}[!ht]
\centering
\begin{tabular}{cc}
\includegraphics[scale = 0.38, trim = 70 0 0 0]{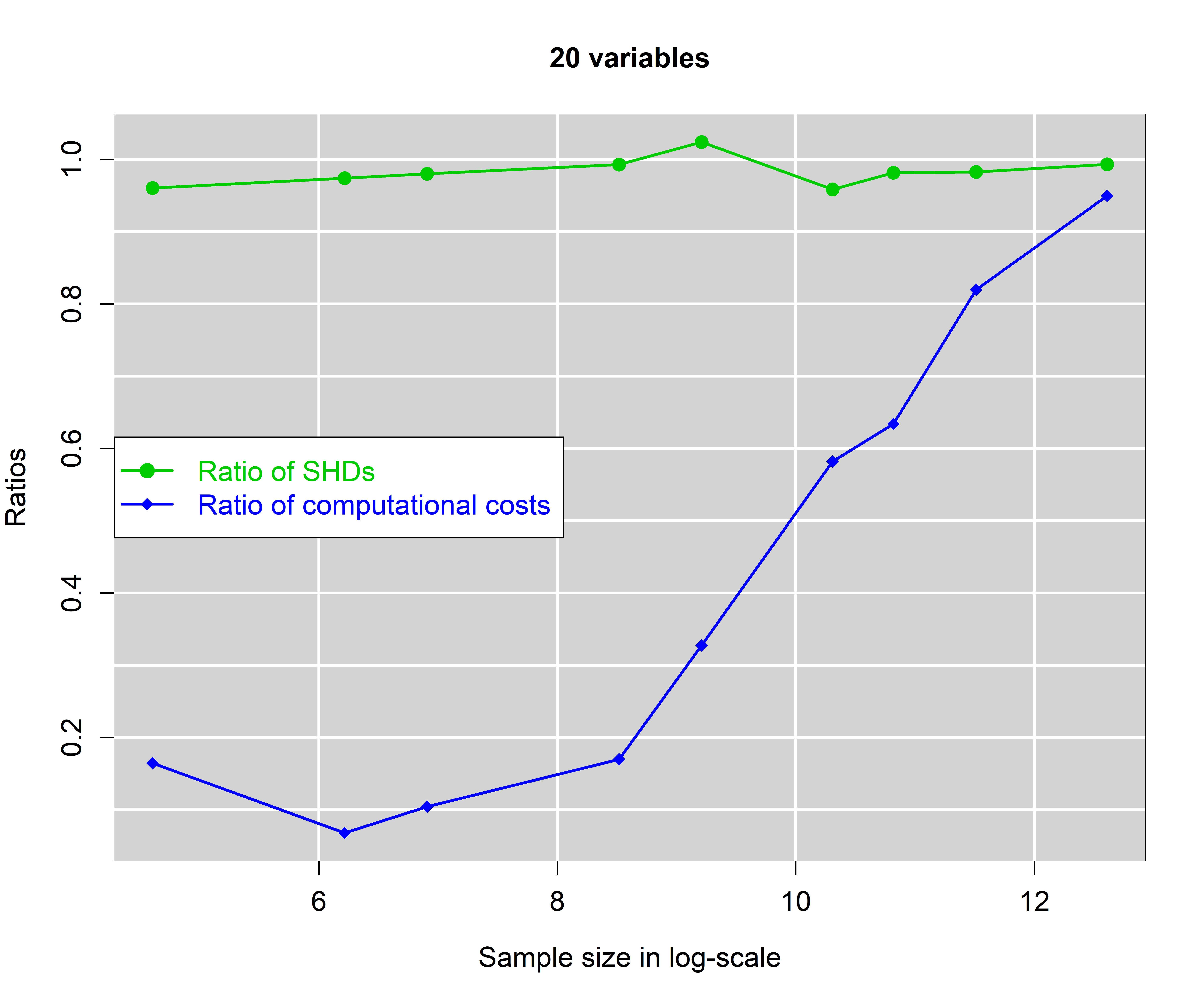}  &
\includegraphics[scale = 0.38, trim = 50 0 0 0]{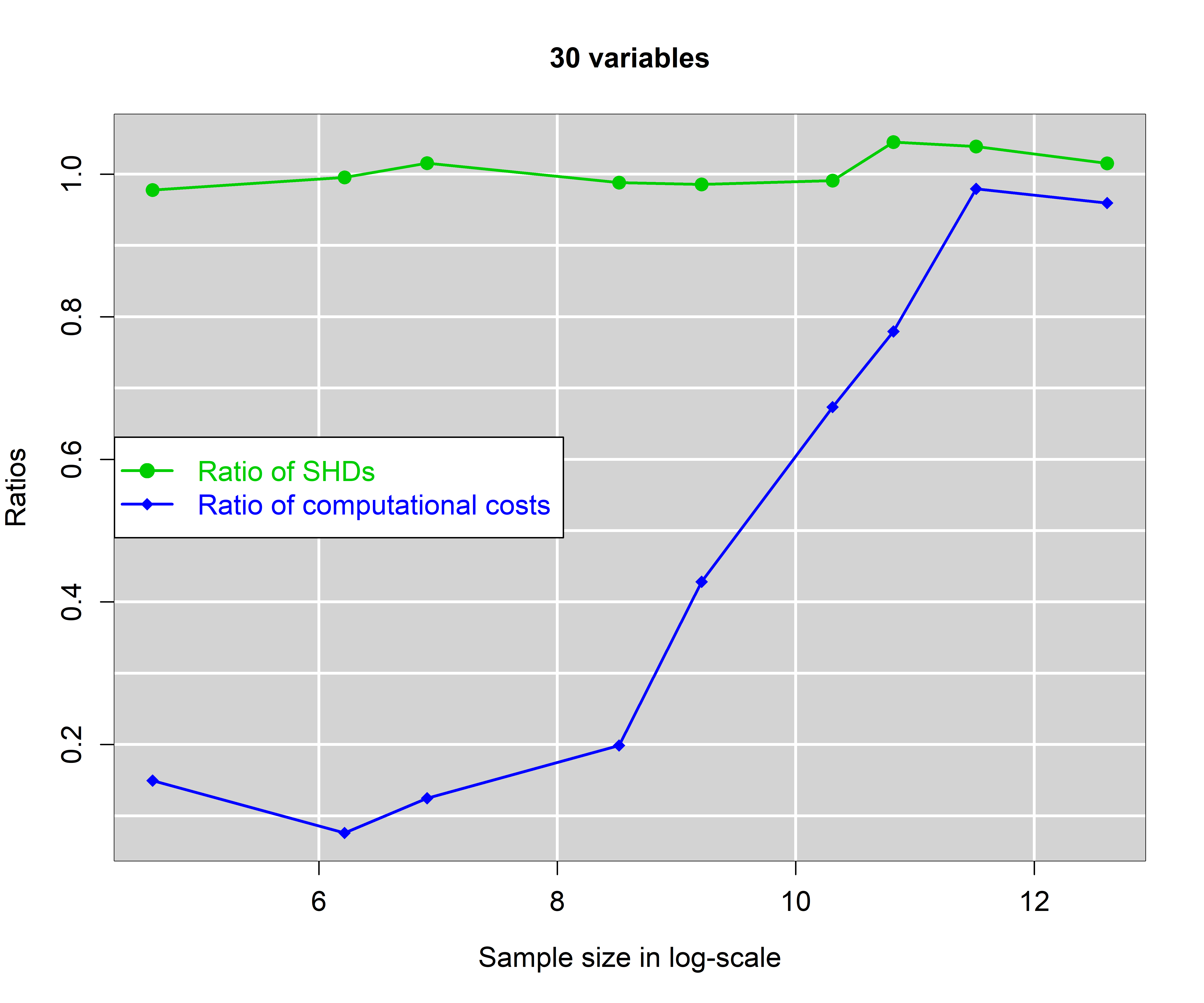}  \\
(a) 3 neighbours. &  (b) 5 neighbours.    \\
\includegraphics[scale = 0.38, trim = 70 0 0 0]{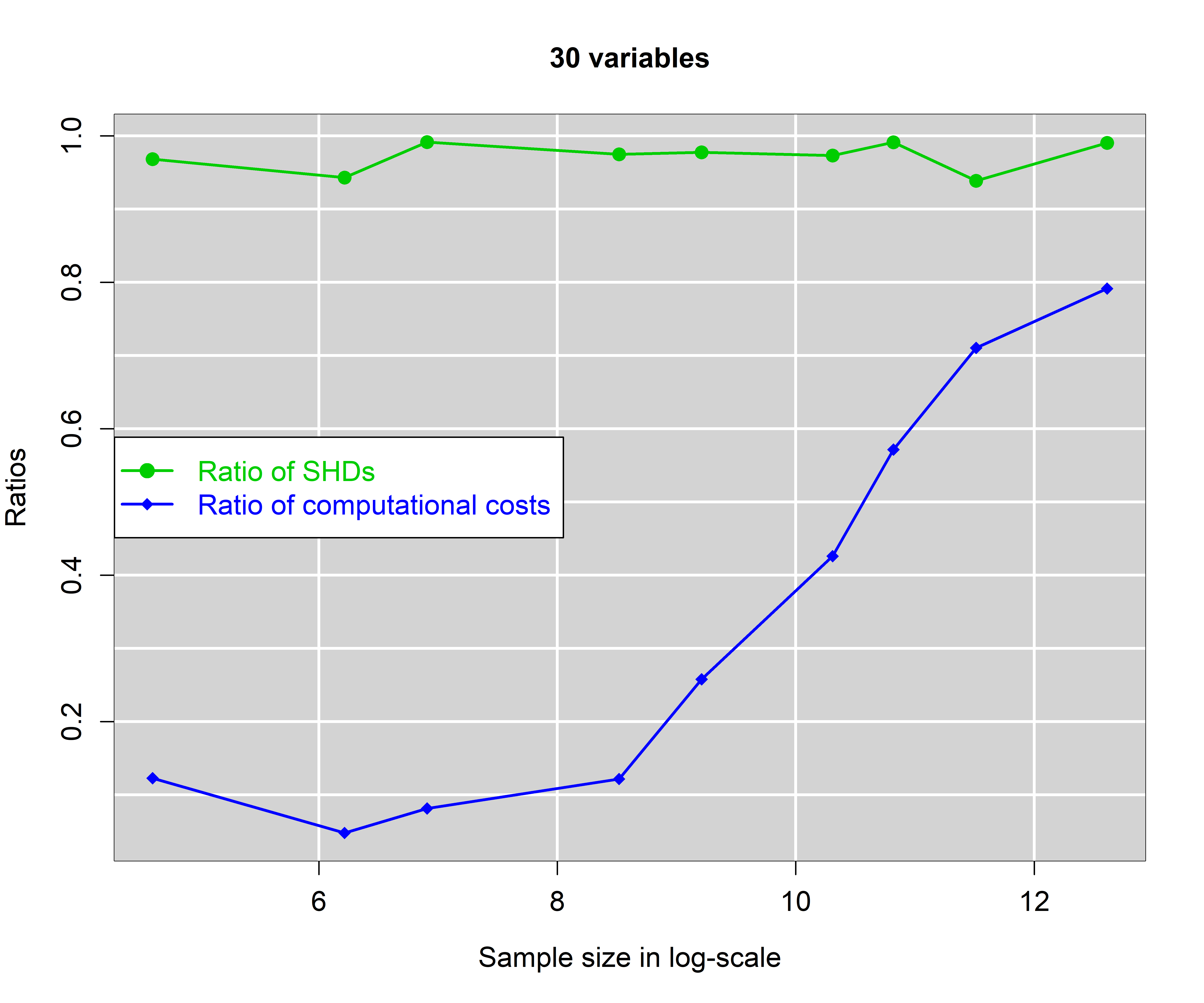}  &
\includegraphics[scale = 0.38, trim = 50 0 0 0]{5_30_robust.png}  \\
(c) 3 neighbours.  &  (d) 5 neighbours.    \\
\end{tabular}
\caption{Ratios of SHD and computational cost against log of sample size for various dimensions with \textbf{3 neighbours} and \textbf{5 neighbours} on average. The ratios depict the errors and computational cost of the raw FEDHC relatively to the robustified FEDHC with NO outliers. \label{robust_no_1} }
\end{figure}

\begin{figure}[!ht]
\centering
\begin{tabular}{cc}
\includegraphics[scale = 0.38, trim = 70 0 0 0]{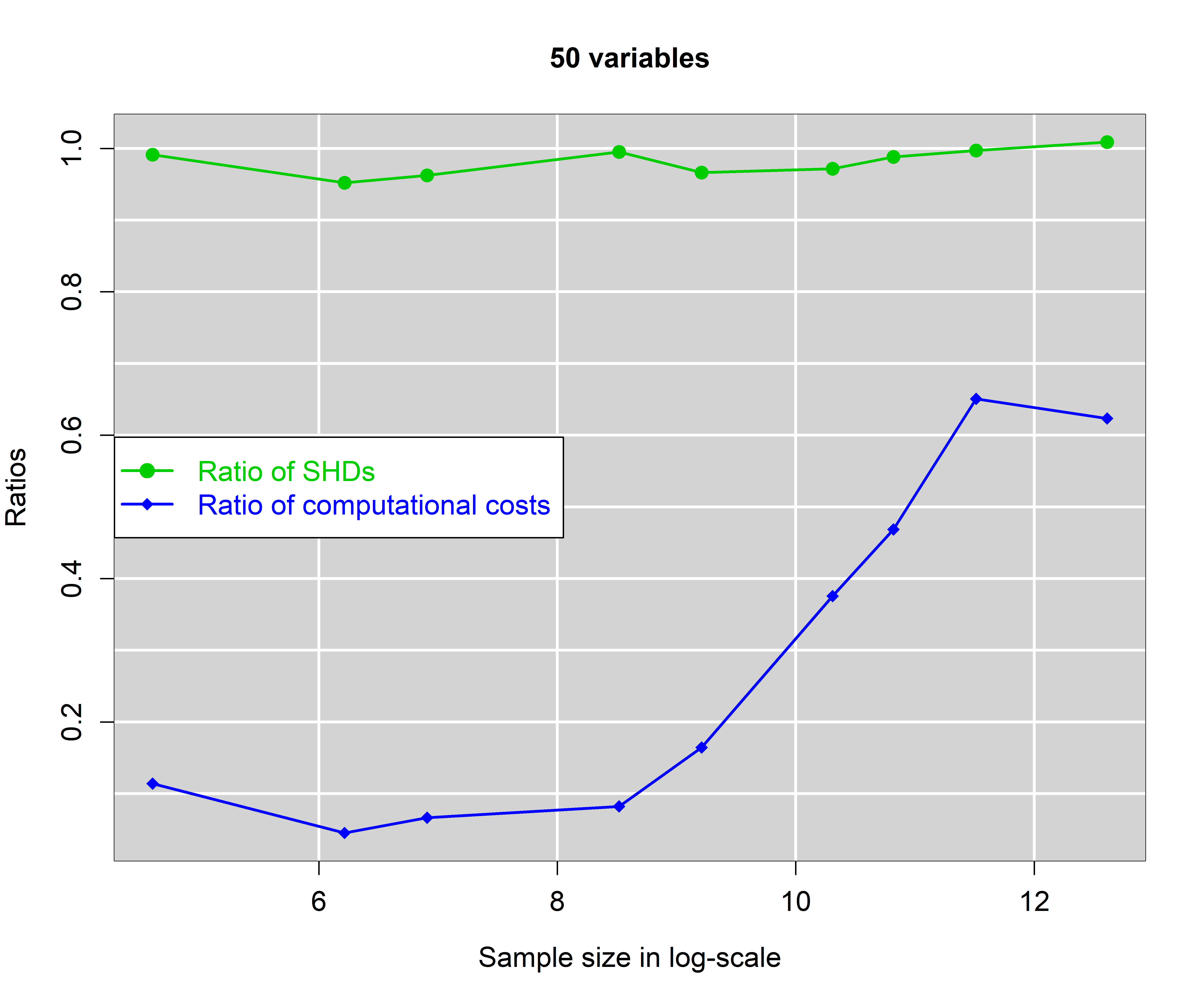}  &
\includegraphics[scale = 0.38, trim = 50 0 0 0]{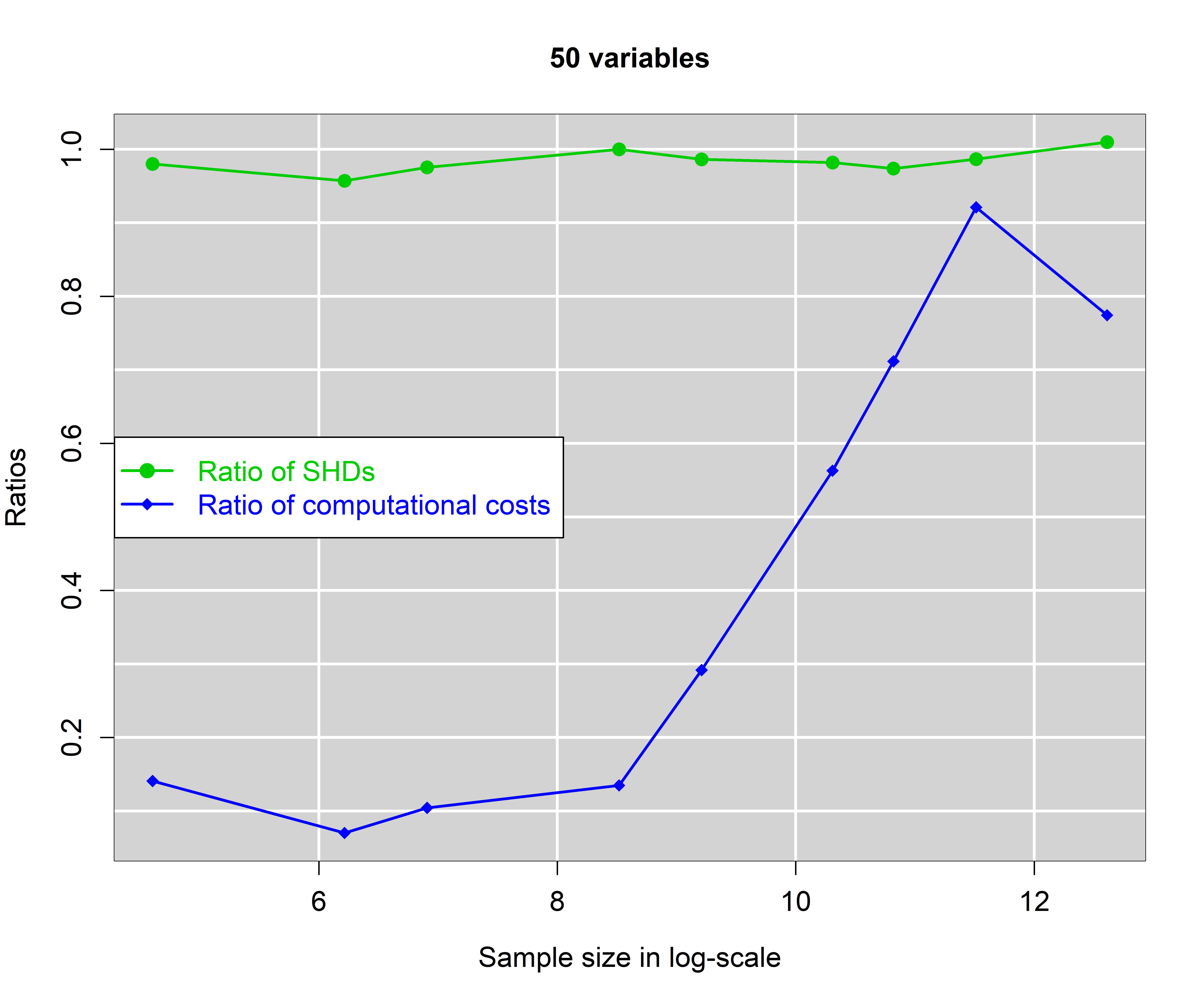}  \\
(a) 3 neighbours. &  (b) 5 neighbours.    \\
\includegraphics[scale = 0.38, trim = 70 0 0 0]{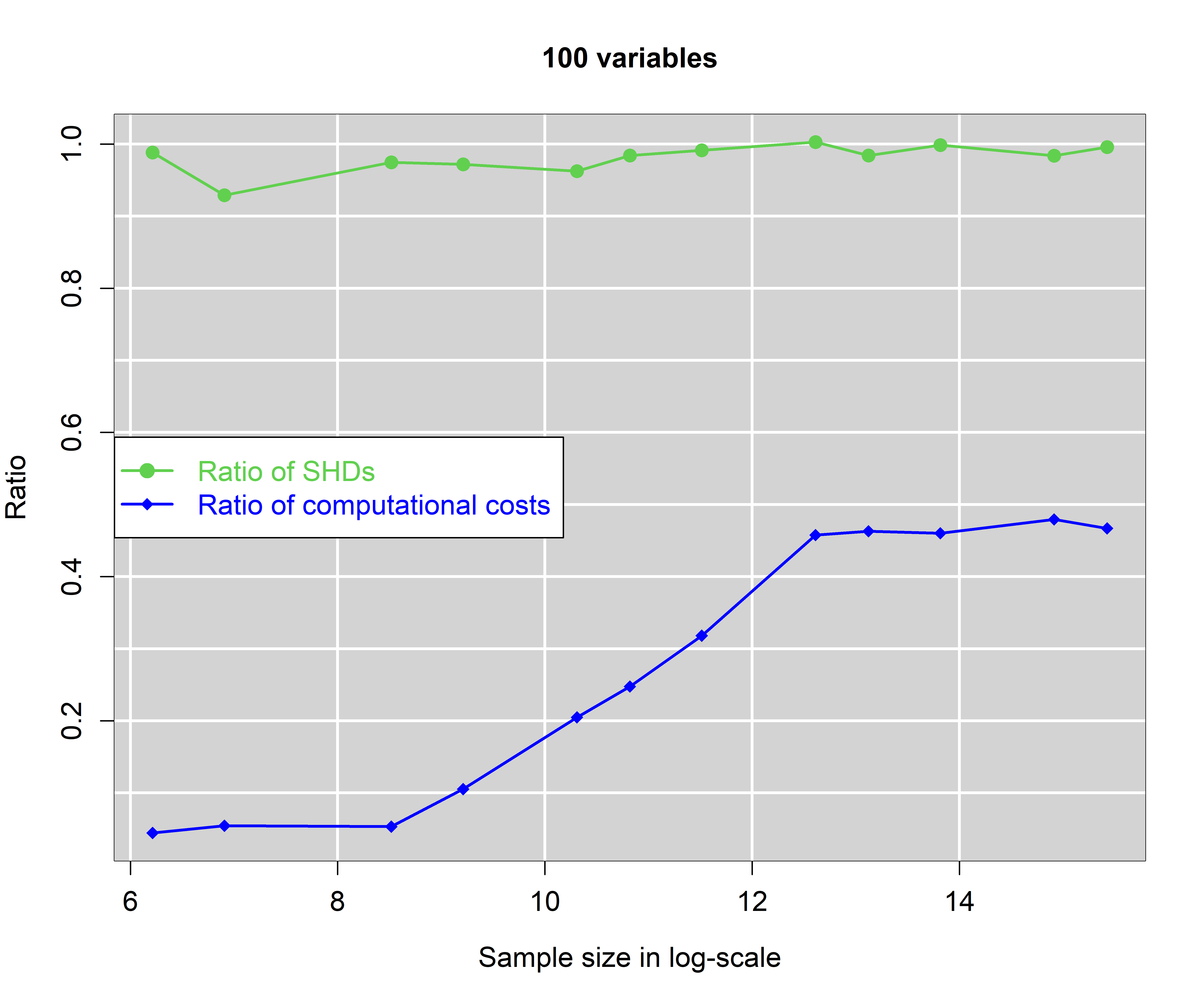}  &
\includegraphics[scale = 0.38, trim = 50 0 0 0]{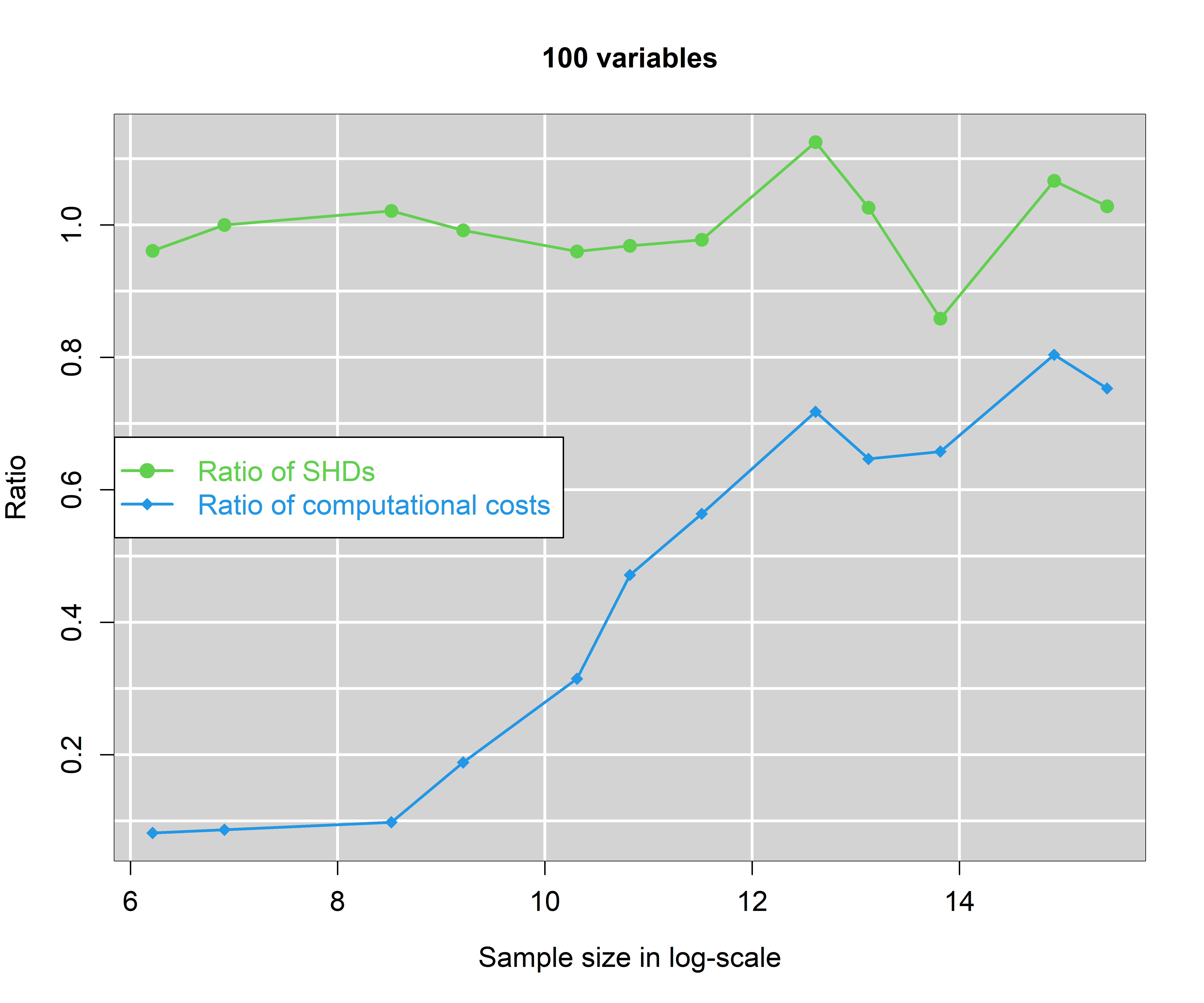}  \\
(c) 3 neighbours.  &  (d) 5 neighbours.    
\end{tabular}
\caption{Ratios of SHD and computational cost against log of sample size for various dimensions with \textbf{3 neighbours} and \textbf{5 neighbours} on average. The ratios depict the errors and computational cost of the raw FEDHC relatively to the robustified FEDHC with NO outliers. \label{robust_no_2} }
\end{figure}

Figures \ref{robust_no_1} and \ref{robust_no_2} signify that there is no loss in the accuracy when using the robustified FEDHC over the raw FEDHC. Computationally speaking though, the raw FEDHC is significantly faster than the robustified FEDHC. For small sample sizes the robustified FEDHC can be 10 times higher, while for large sample sizes its cost can be double or only 5\% higher than that of the raw FEDHC.
 
The performances of the raw FEDHC and of the robustified FEDHC in the presence of extreme outliers are evaluated next. The BN generation scheme is the one described in Section \ref{datgen} with the exception of having added a  5\% of outlying observations. The considered sample sizes are smaller, as although FEDHC is computationally efficient, it becomes really slow in the presence of outliers. 

\begin{figure}[!ht]
\centering
\begin{tabular}{cc}
\includegraphics[scale = 0.38, trim = 70 0 0 0]{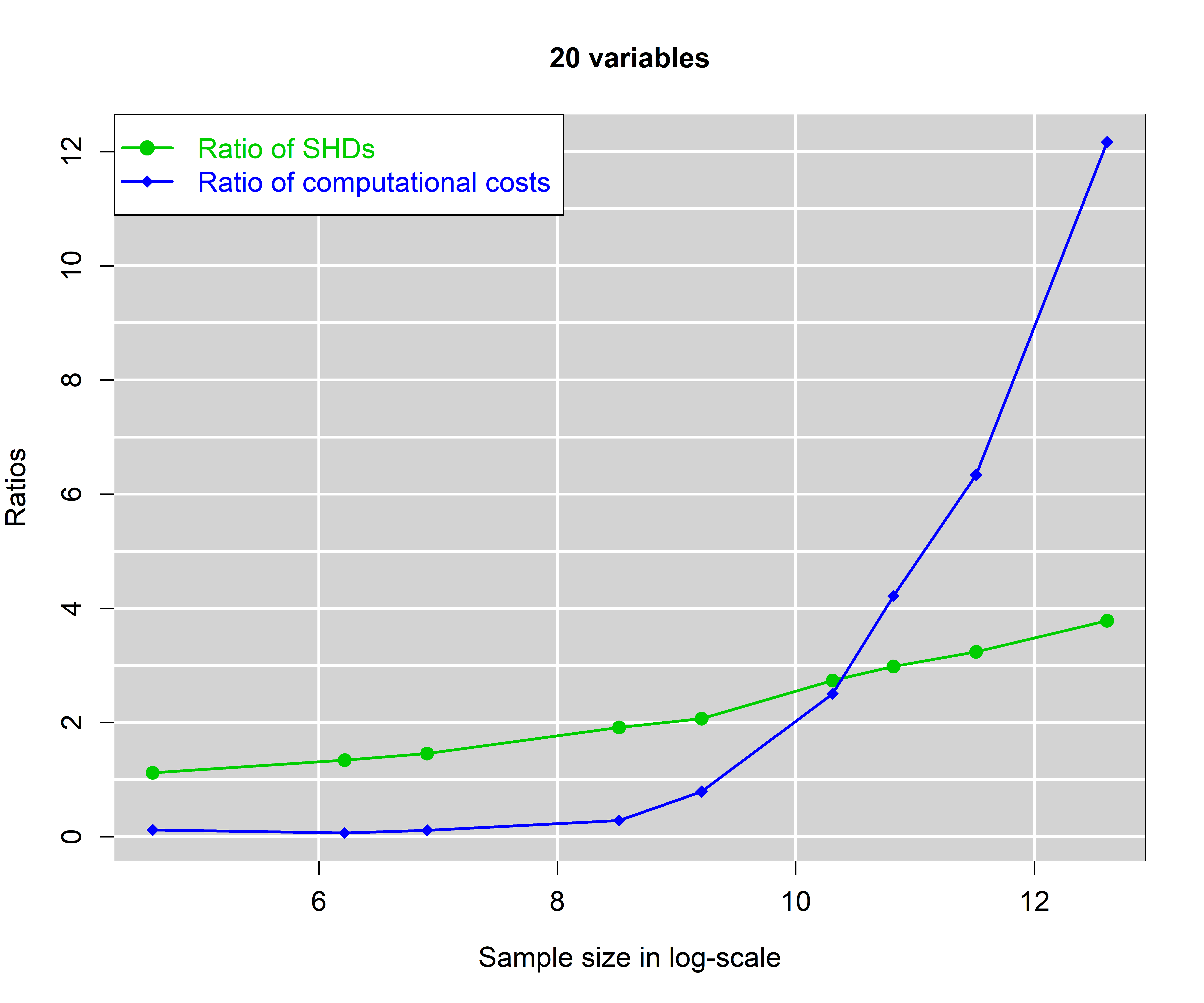}  &
\includegraphics[scale = 0.38, trim = 50 0 0 0]{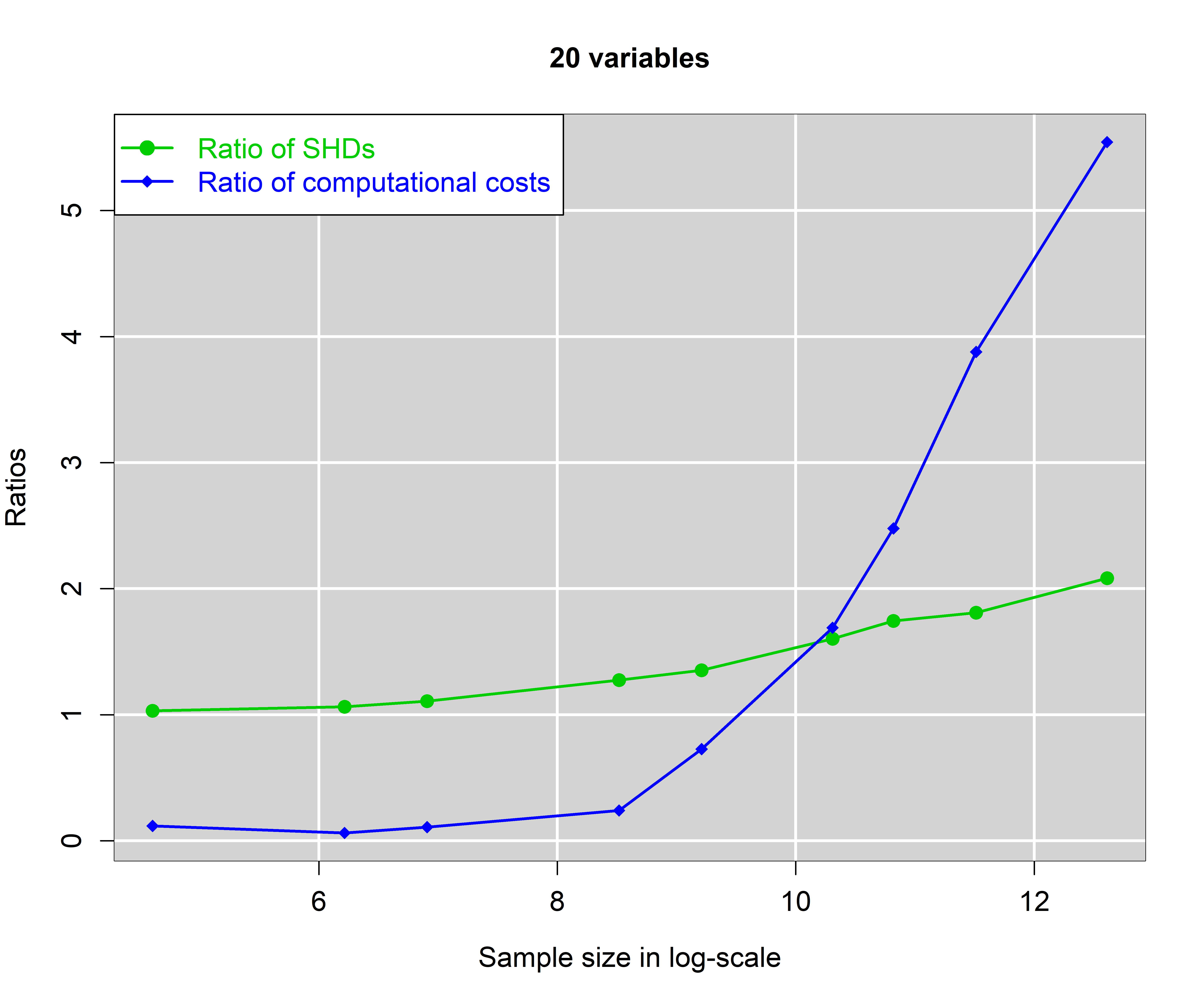}  \\
(a) 3 neighbours. &  (b) 5 neighbours.    \\
\includegraphics[scale = 0.38, trim = 70 0 0 0]{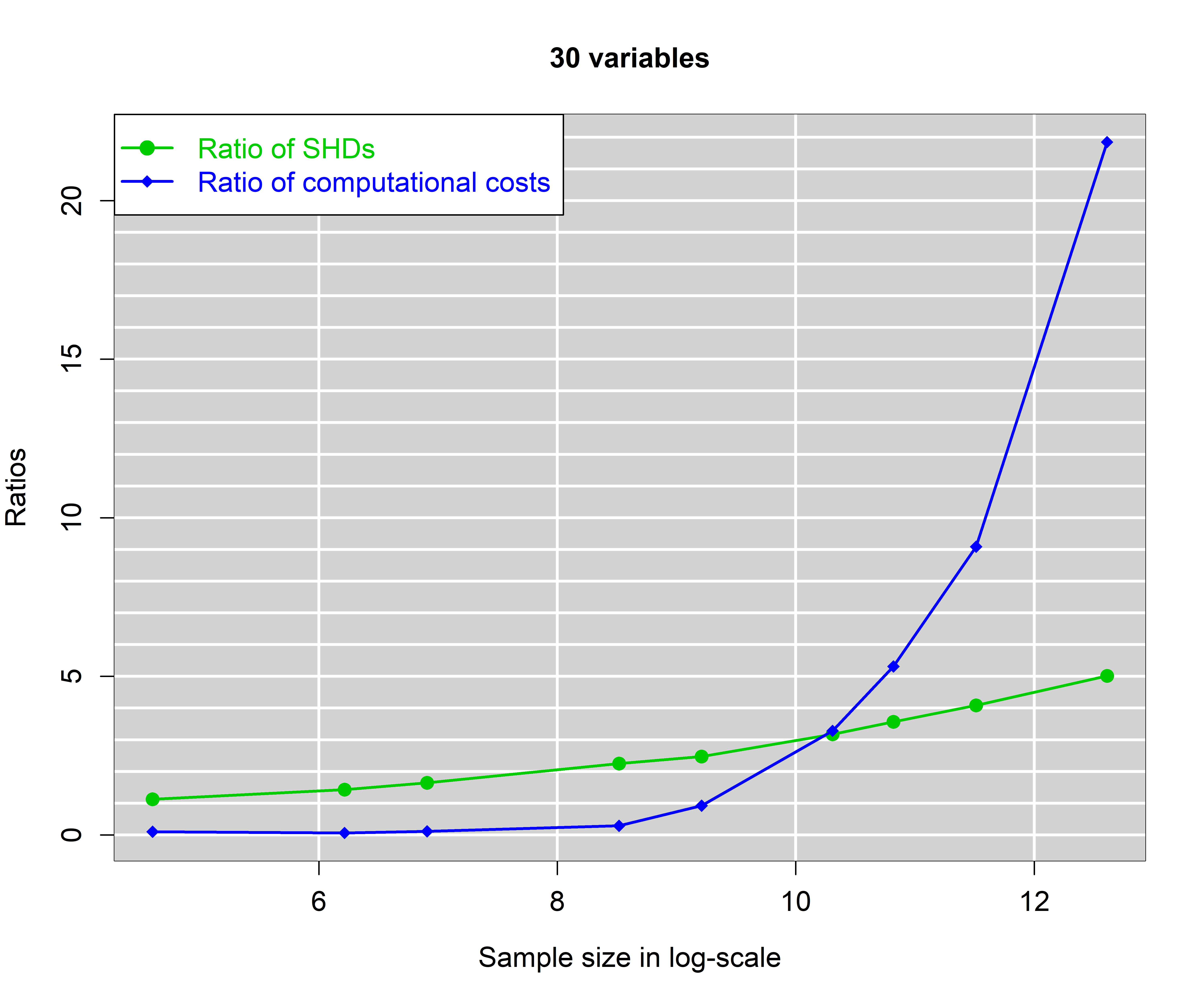}  &
\includegraphics[scale = 0.38, trim = 50 0 0 0]{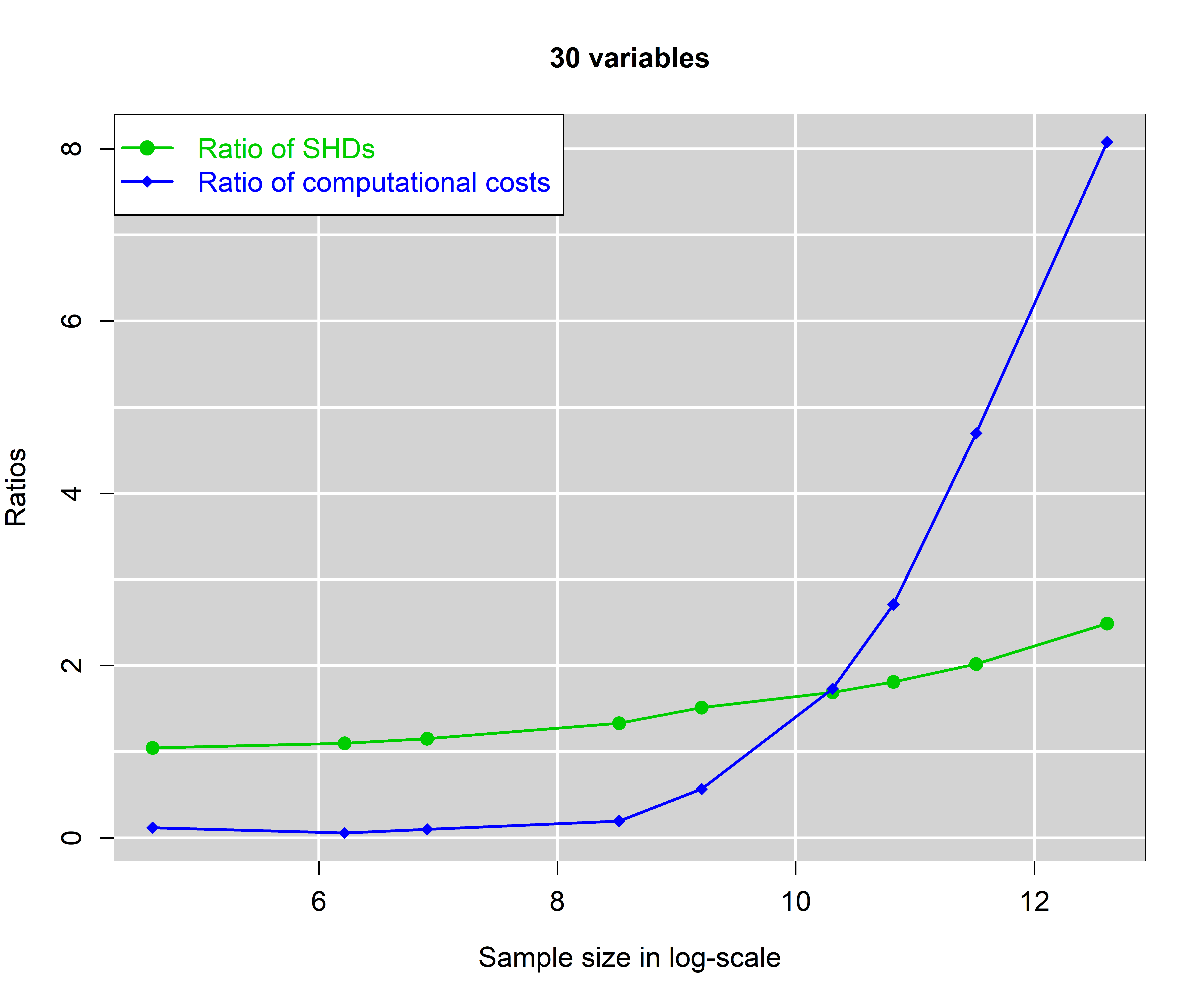}  \\
(c) 3 neighbours.  &  (d) 5 neighbours.    \\
\end{tabular}
\caption{Ratios of SHD and computational cost against log of sample size for 20 and 30 dimensions with \textbf{3 neighbours} and \textbf{5 neighbours} on average. The ratios depict the errors and computational cost of the raw FEDHC relatively to the robustified FEDHC with 5\% outliers. \label{robust1} }
\end{figure}

\begin{figure}[!ht]
\centering
\begin{tabular}{cc}
\includegraphics[scale = 0.38, trim = 70 0 0 0]{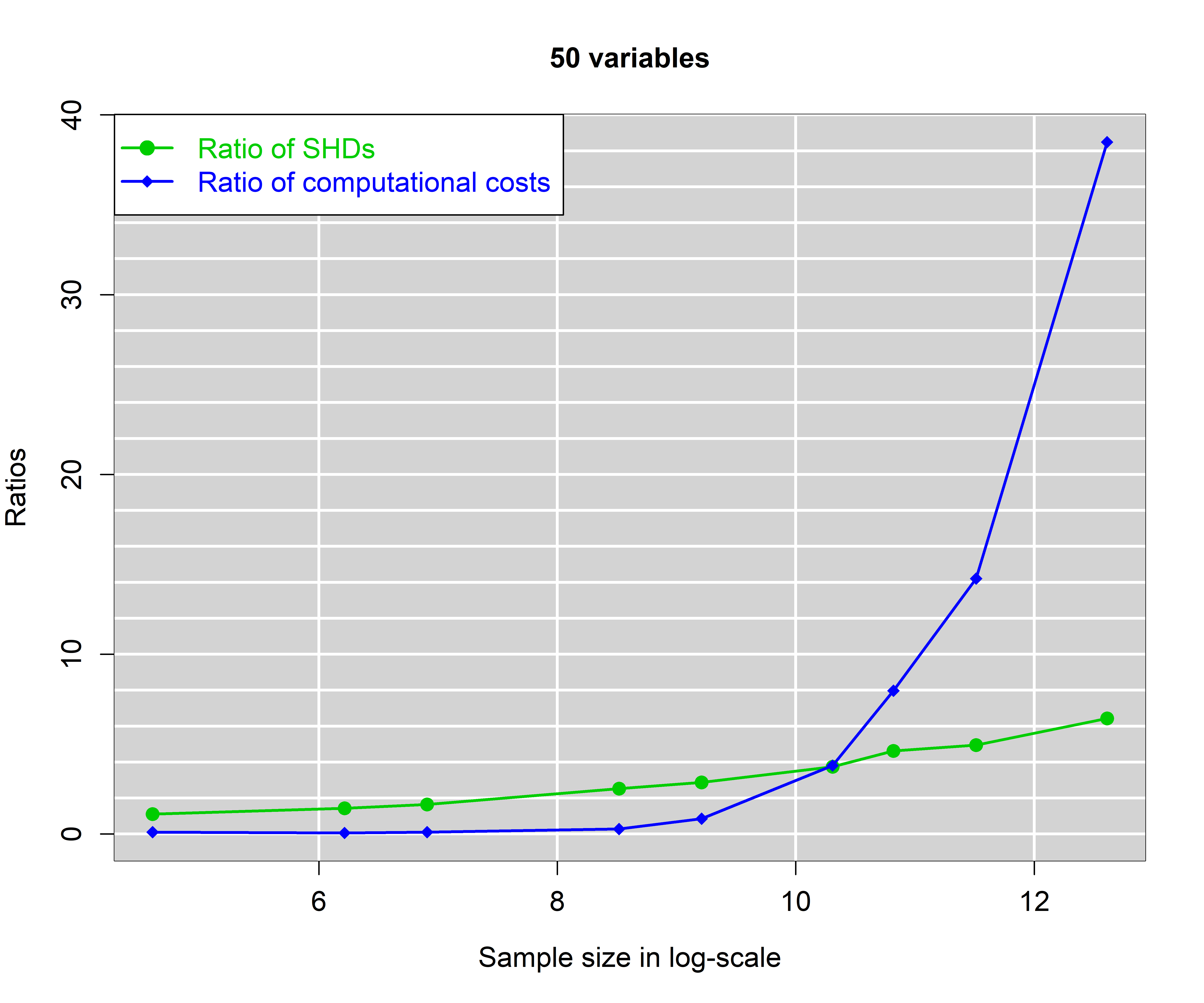}  &
\includegraphics[scale = 0.38, trim = 50 0 0 0]{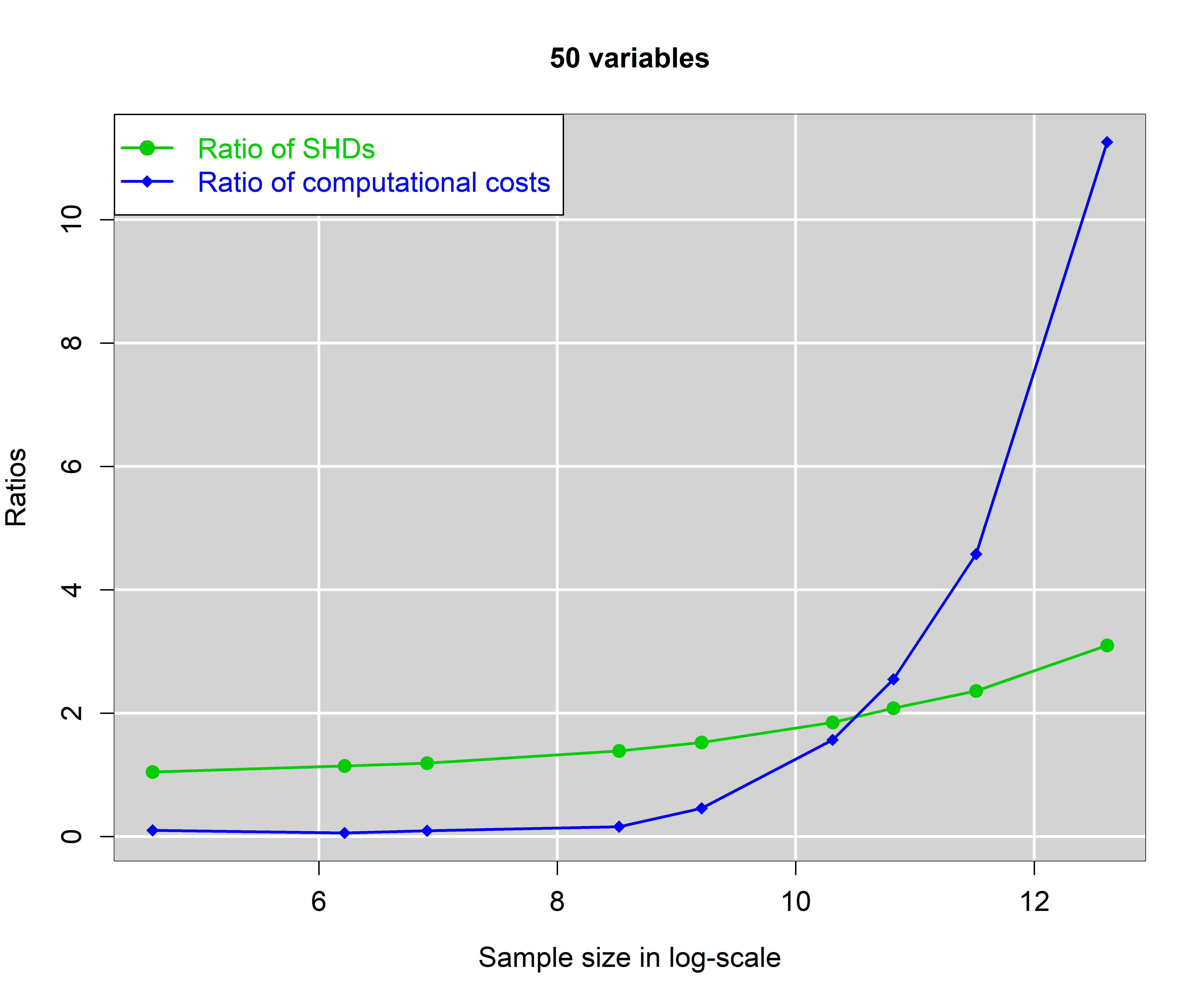}  \\
(a) 3 neighbours. &  (b) 5 neighbours.  \\
\includegraphics[scale = 0.38, trim = 70 0 0 0]{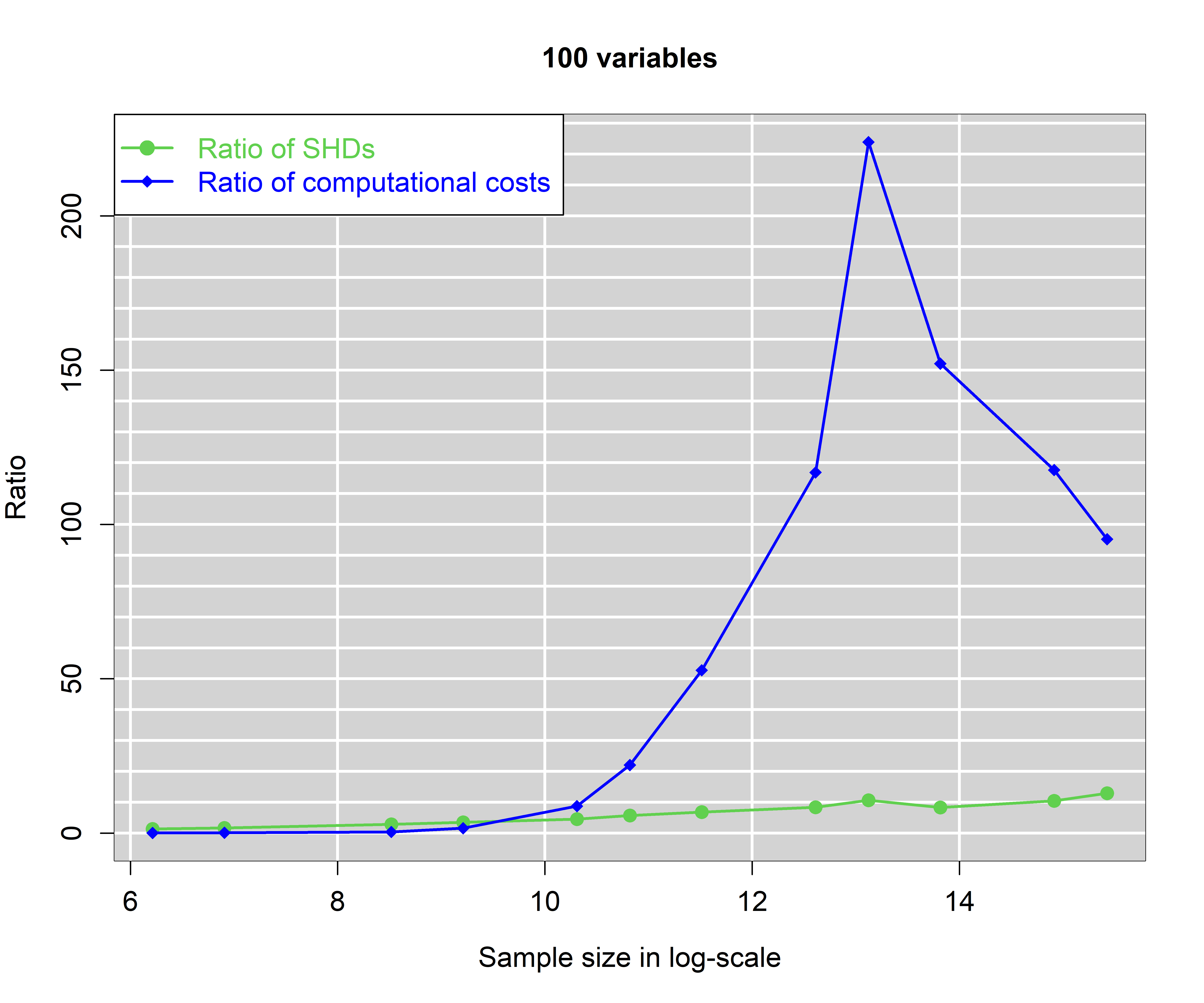}  &
\includegraphics[scale = 0.38, trim = 50 0 0 0]{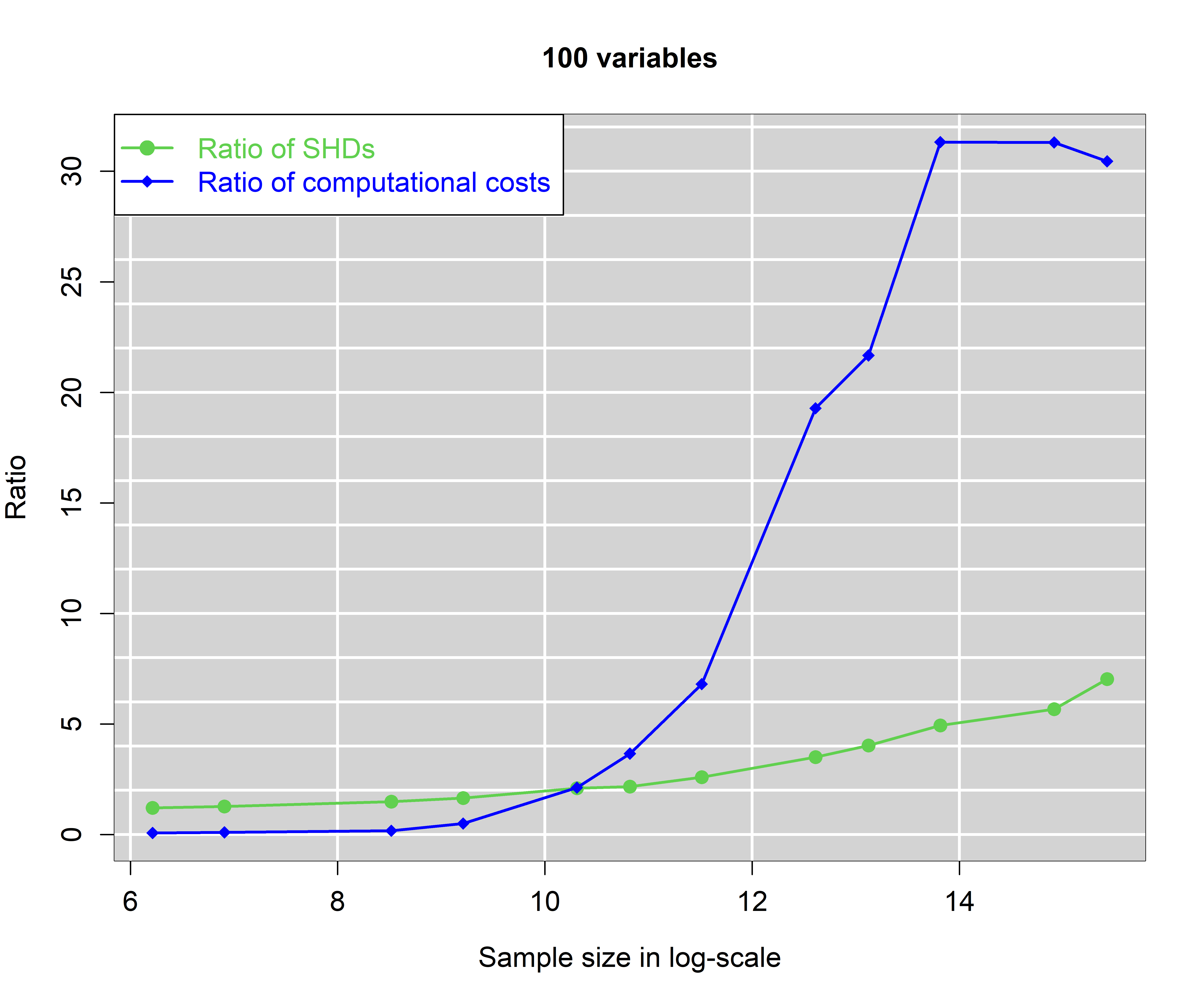}  \\
(c) 3 neighbours. &  (d) 5 neighbours.  
\end{tabular}
\caption{Ratios of SHD and computational cost against log of sample size for 50 and 100 dimensions with \textbf{3 neighbours} and \textbf{5 neighbours} on average. The ratios depict the errors and computational cost of the raw FEDHC relatively to the robustified FEDHC with 5\% outliers. \label{robust2} }
\end{figure}

The results presented in Figures \ref{robust1} and \ref{robust2} evidently show the gain when using the robustified FEDHC over the raw FEDHC. The SHD of the raw FEDHC increases by as little as 100\% up to 700\% with 3 neighbours on average and 50 variables. The duration of the raw FEDHC increases substantially\footnote{Similar patterns were observed for the duration of the skeleton learning phase and for the number of CI tests}. The raw FEDHC becomes up to more than 200 times slower than the robustified version with hundreds of thousands of observations and 50 variables. This is attributed to the HC phase of the raw FEDHC which consumes a tremendous amount of time. The outliers produce noise that escalates the labour of the HC phase.  

\subsection{Realistic BNs with categorical data}
The $f\left(Pa(X)\right)$ function utilised in the continuous data case relies on the $\beta$ coefficients. The larger the magnitude of their values, the stronger the association between the edges becomes and hence it becomes easier to identify them. For BNs with categorical data one could apply the same generation technique and then discretize the simulated data. To avoid biased or optimistic estimates favoring one or the other method, two real BNs with categorical data were utilised to simulate data. These are a) the \textit{Insurance} BN, used for evaluating car insurance risks \citep{beinlich1989}, that consists of 27 variable (nodes) and 52 (directed) edges and b) the \textit{Alarm} BN, designed to provide an alarm message system for patient monitoring and consists of 37 variables and 45 (directed) edges. The \textit{R} package \textit{bnlearn} contains a few thousand categorical instantiations from these BNs, but for the purpose of the simulation studies more instantiations were generated using the same package. The sample sizes considered were $n=(1\times 10^4, 2\times 10^4, 5\times 10^4, 1\times 10^5, 2\times 10^5, 5\times 10^5, 1\times 10^6, 2\times 10^6, 5\times 10^6)$.

\begin{figure}[!ht]
\centering
\begin{tabular}{cc}
\includegraphics[scale = 0.38, trim = 70 0 0 0]{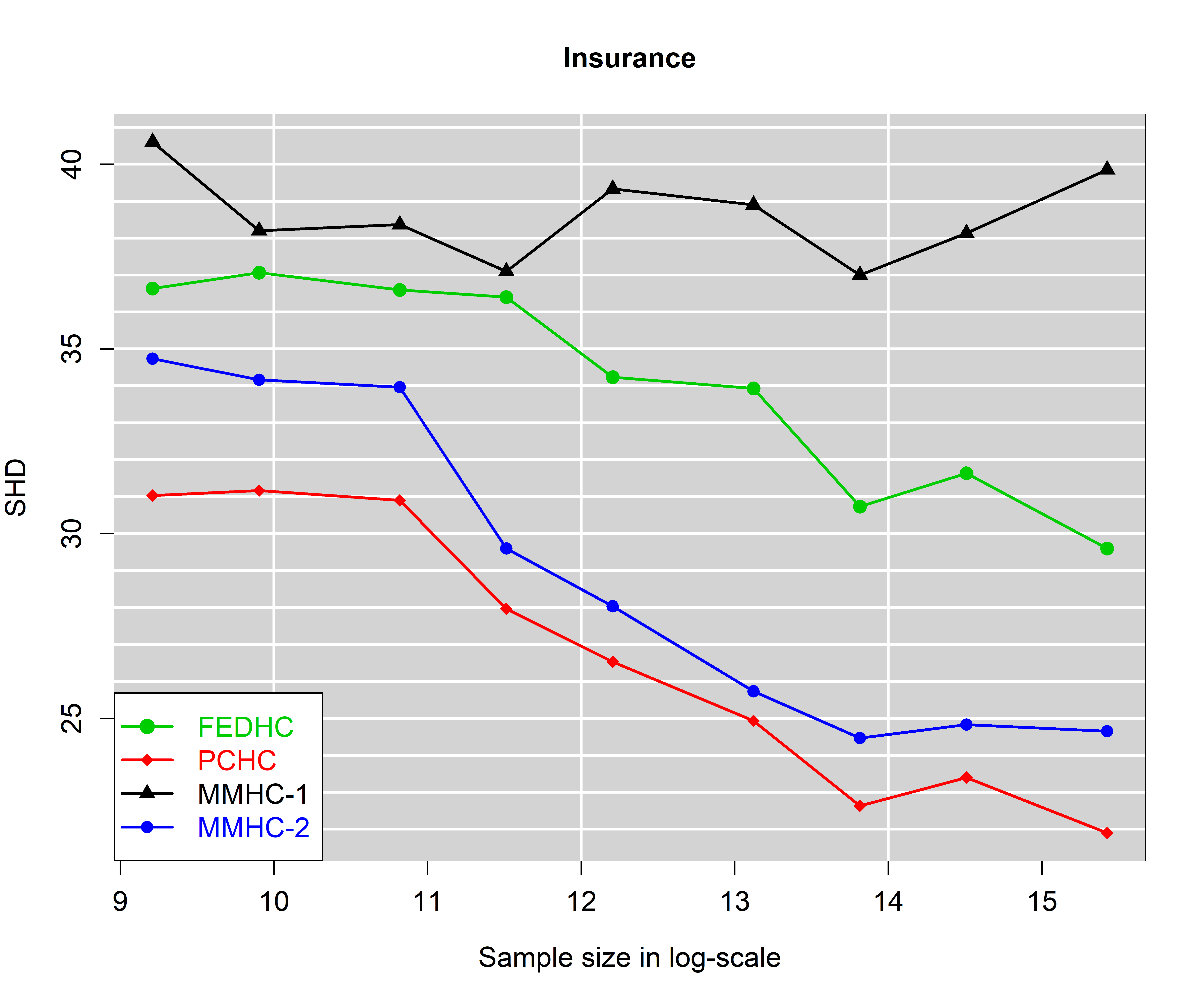}  &
\includegraphics[scale = 0.38, trim = 50 0 0 0]{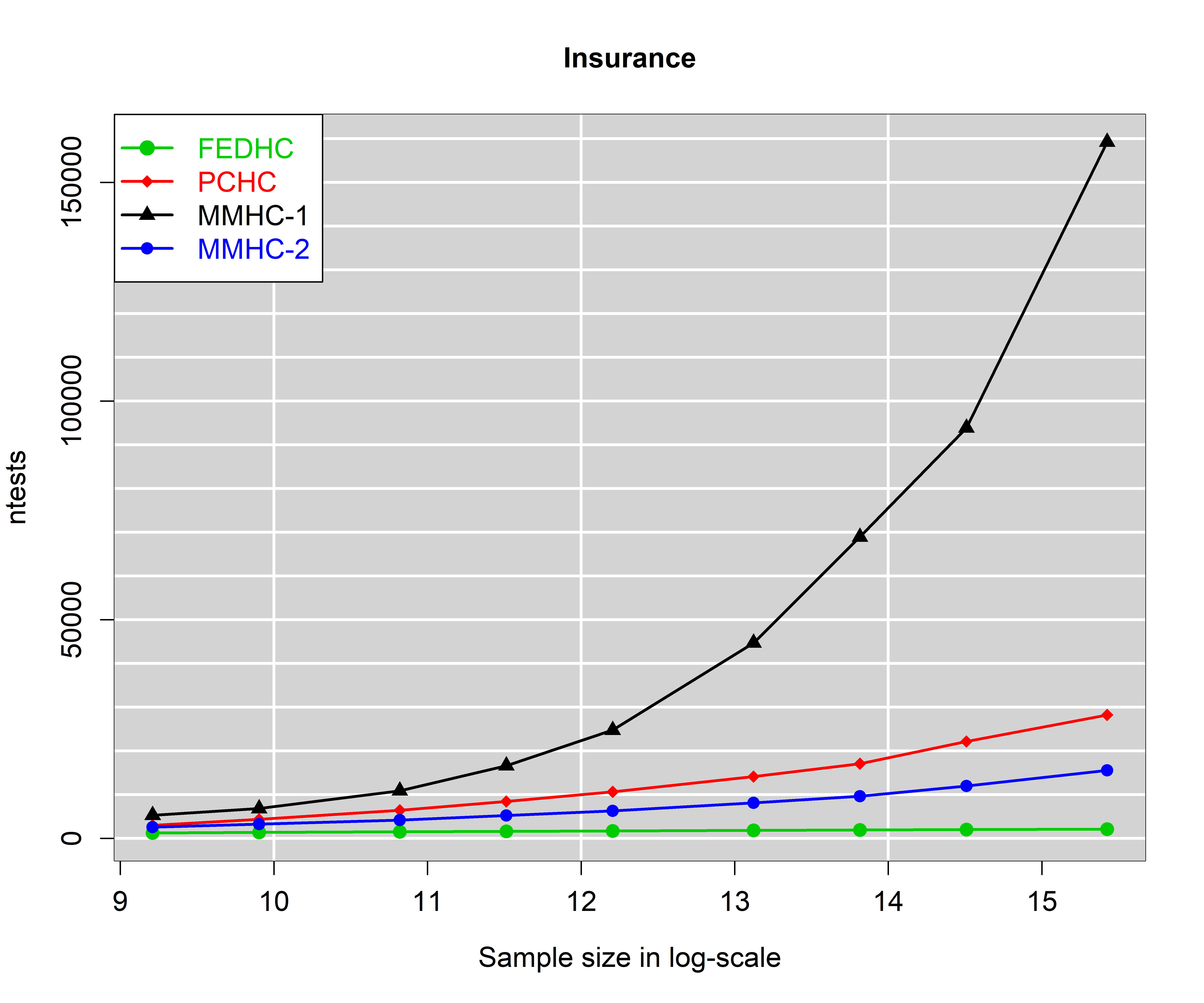}  \\
(a) SHD vs log of sample size. &  (b) Number of CI tests vs log of sample size.  \\
\includegraphics[scale = 0.38, trim = 70 0 0 0]{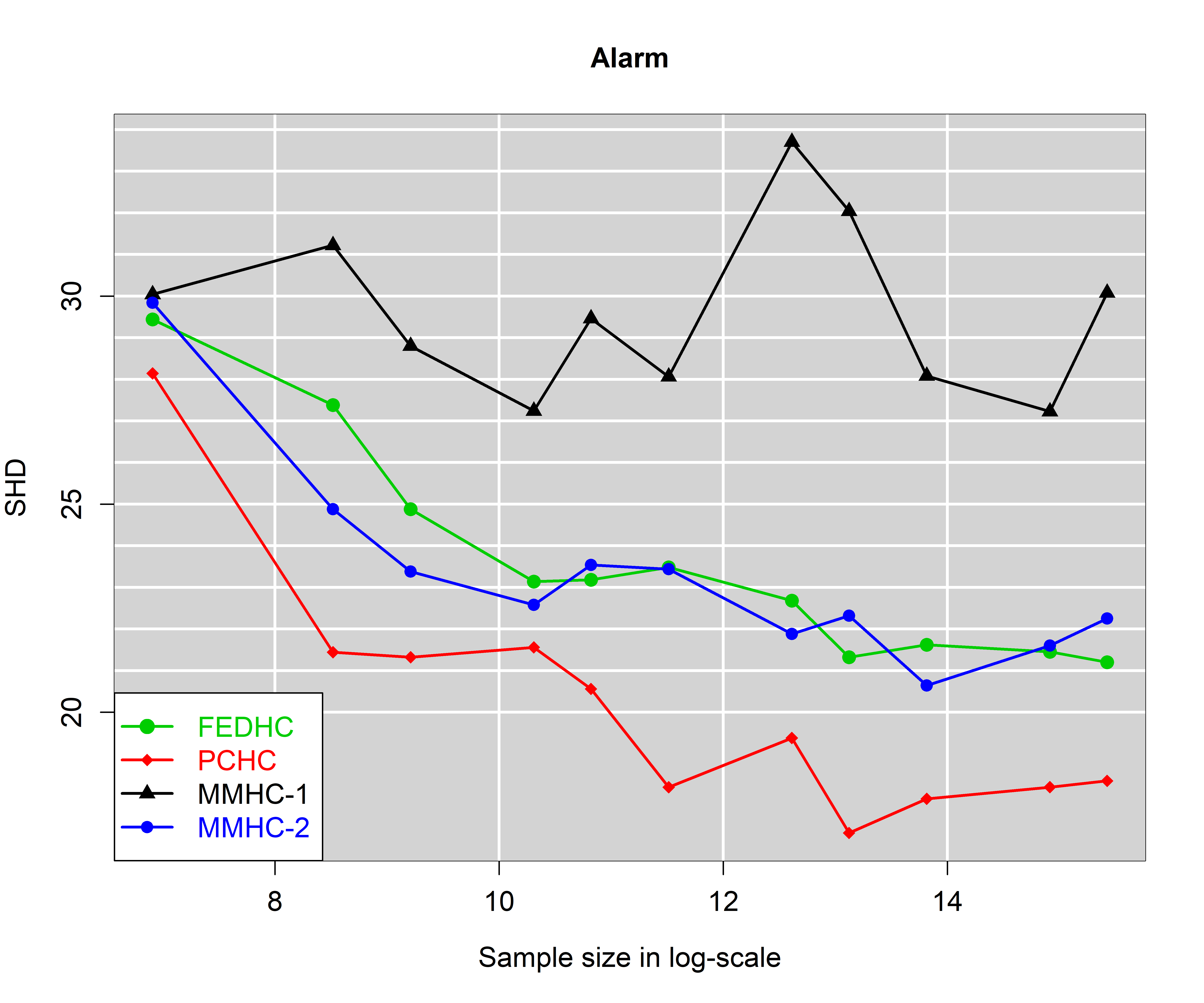}  &
\includegraphics[scale = 0.38, trim = 50 0 0 0]{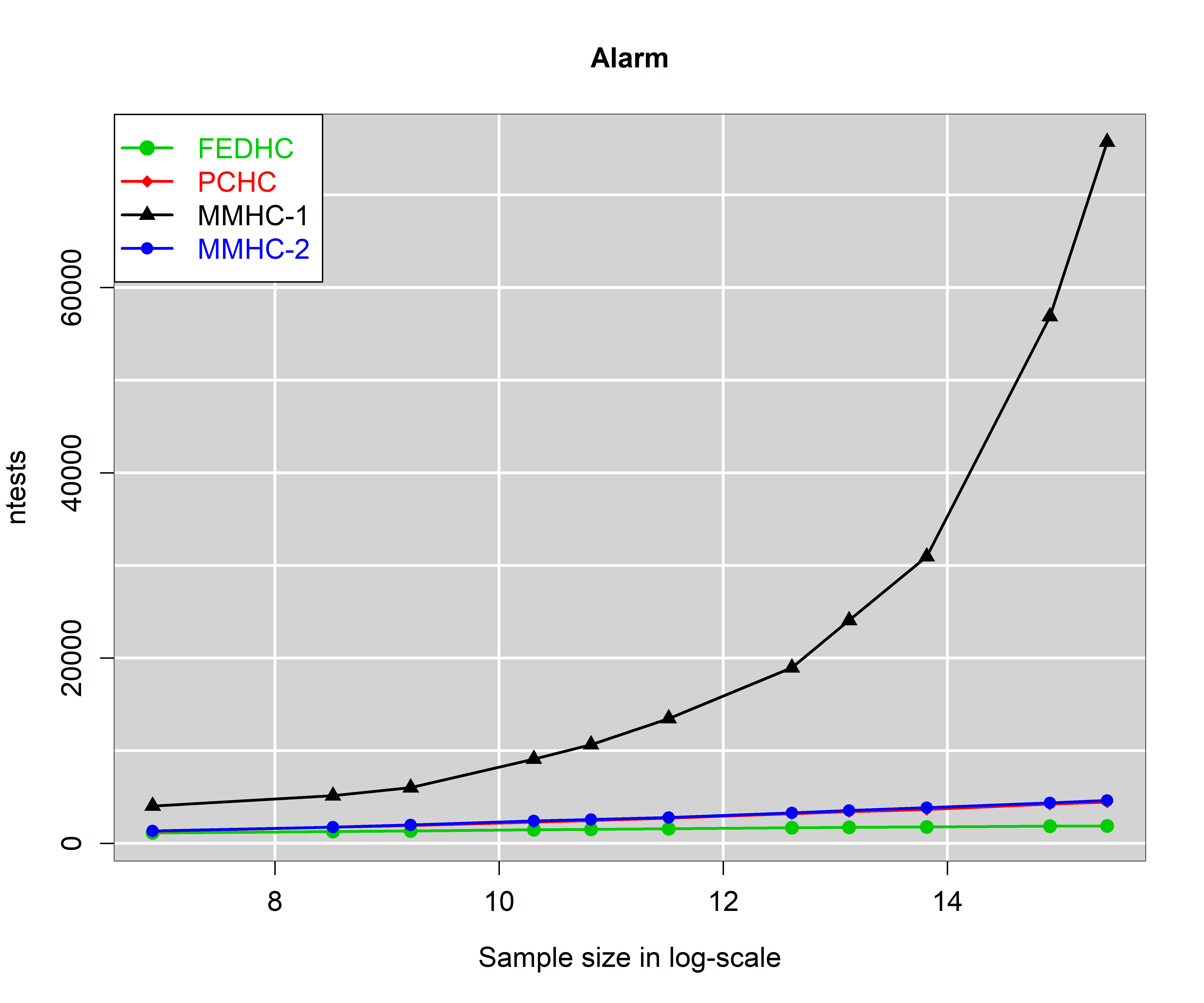}  \\
(c) SHD vs log of sample size. &  (d) Number of CI tests vs log of sample size.  \\
\end{tabular}
\caption{SHD and number of CI tests against log of sample size with \textbf{categorical} data. \label{realistic} }
\end{figure}

Figure \ref{realistic} shows the SHD and the number of CI tests executed by each algorithm against the sample size. The MMHC-1 has evidently the poorest performance in both axes of comparison. Our implementation (MMHC-2) performs substantially better but the overall winner is the PCHC. FEDHC on the other hand performs better than MMHC-1 yet is the second best option. 

\subsection{Scalability of FEDHC} \label{scalability}
\begin{figure}[!ht]
\centering
\begin{tabular}{cc}
\includegraphics[scale = 0.38, trim = 70 0 0 0]{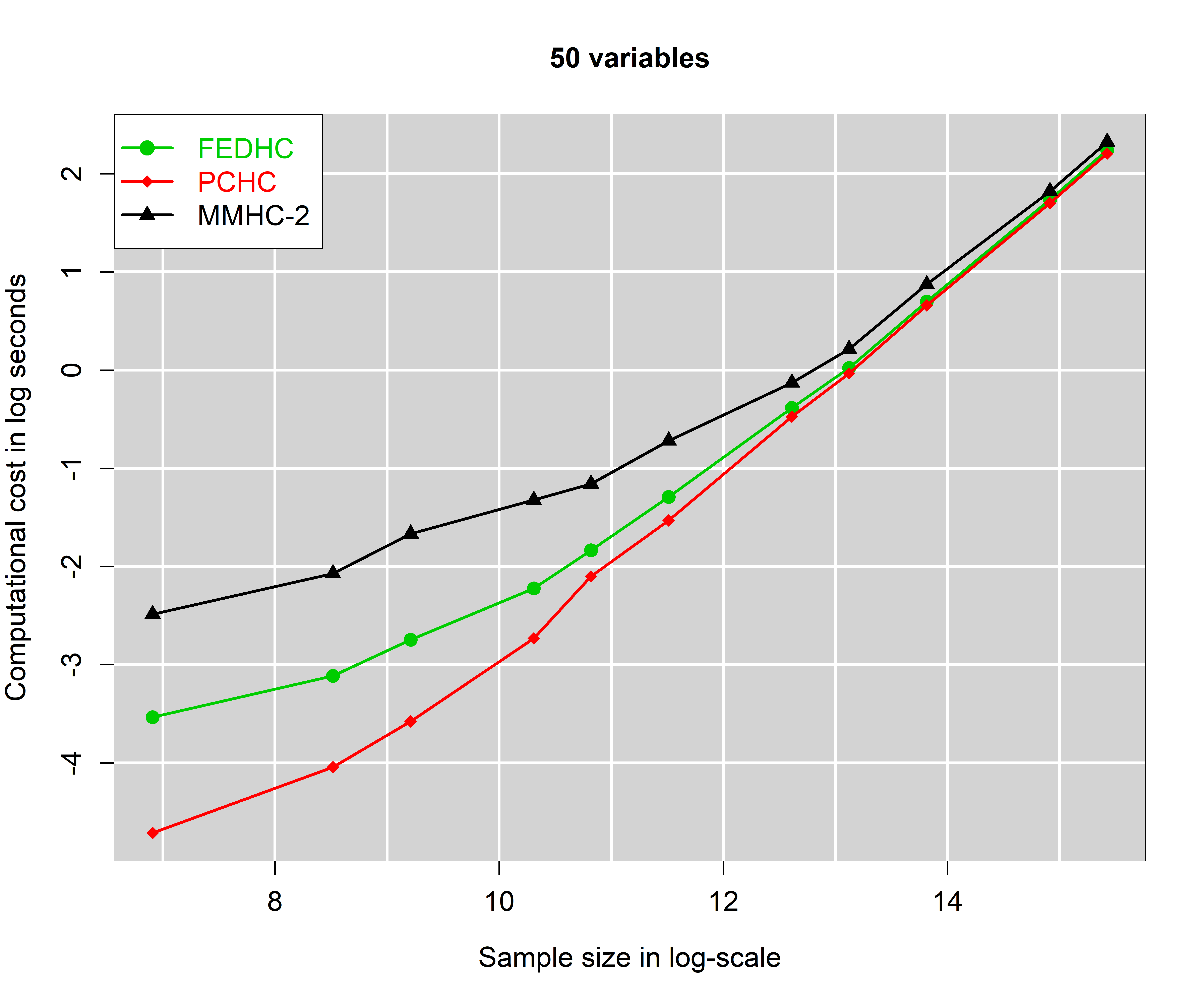}  &
\includegraphics[scale = 0.38, trim = 50 0 0 0]{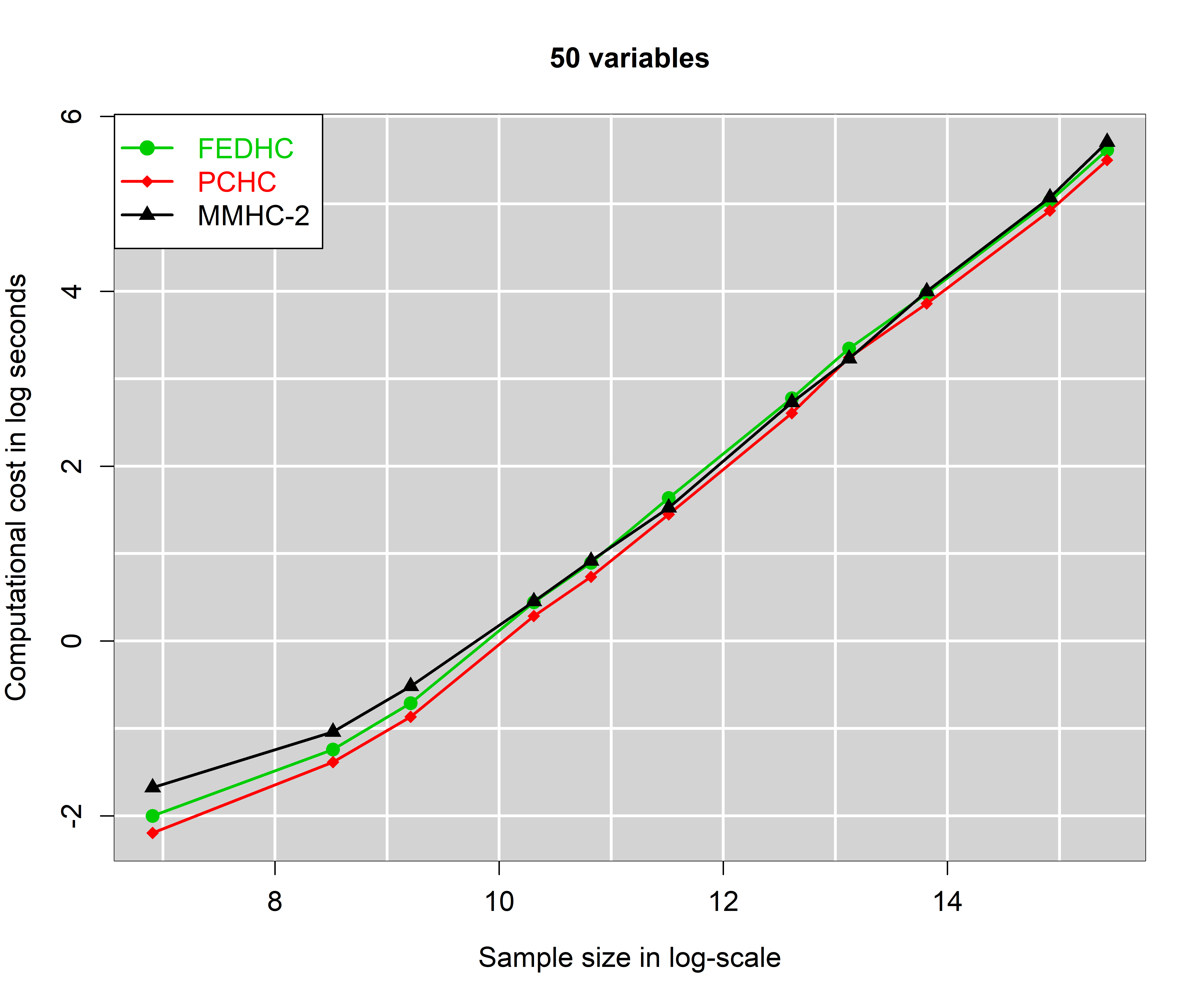}  \\
\multicolumn{2}{c}{\textbf{Continuous data with 3 neighbours on average.}} \\
\includegraphics[scale = 0.38, trim = 50 10 0 0]{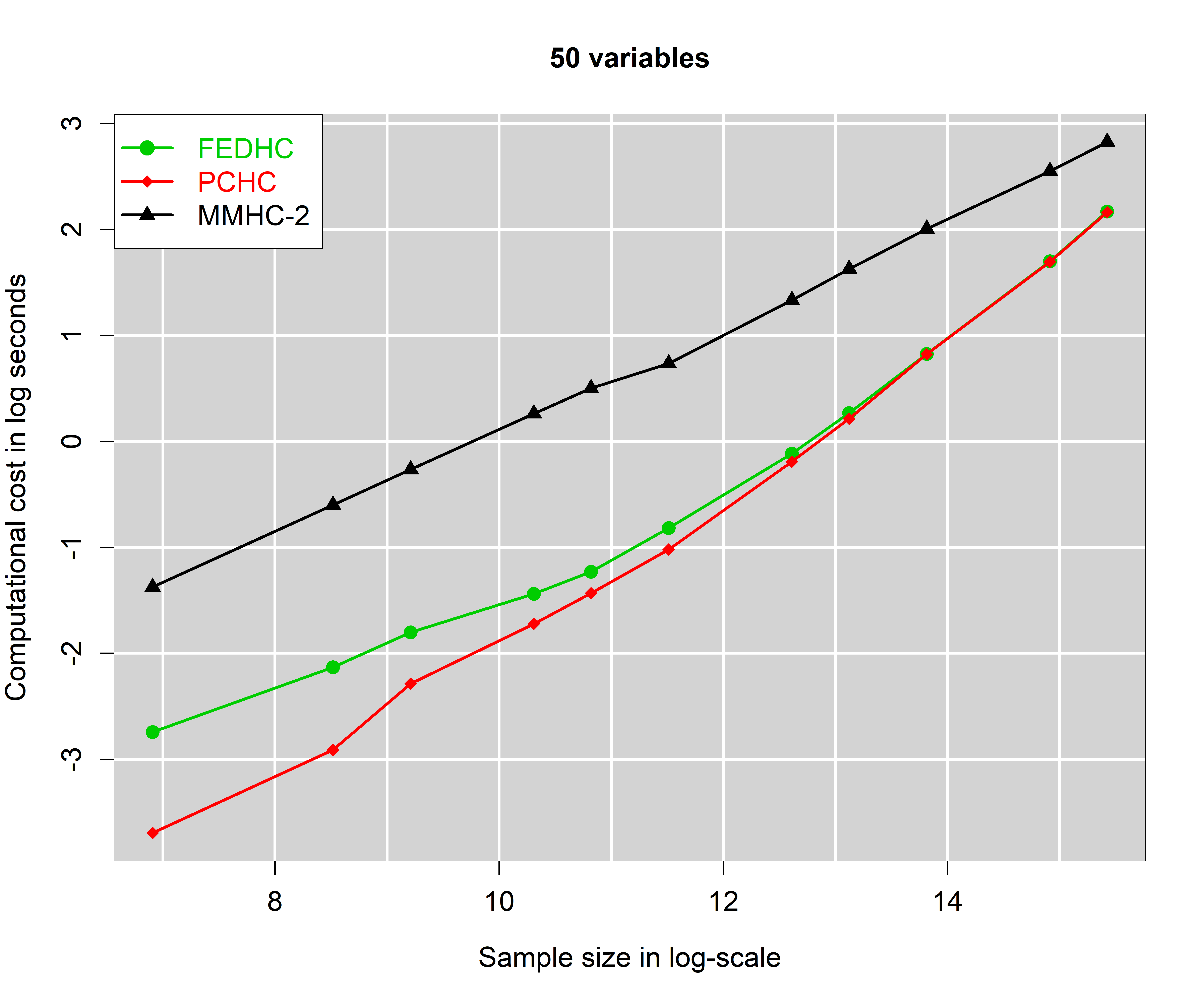}  &
\includegraphics[scale = 0.38, trim = 50 10 0 0]{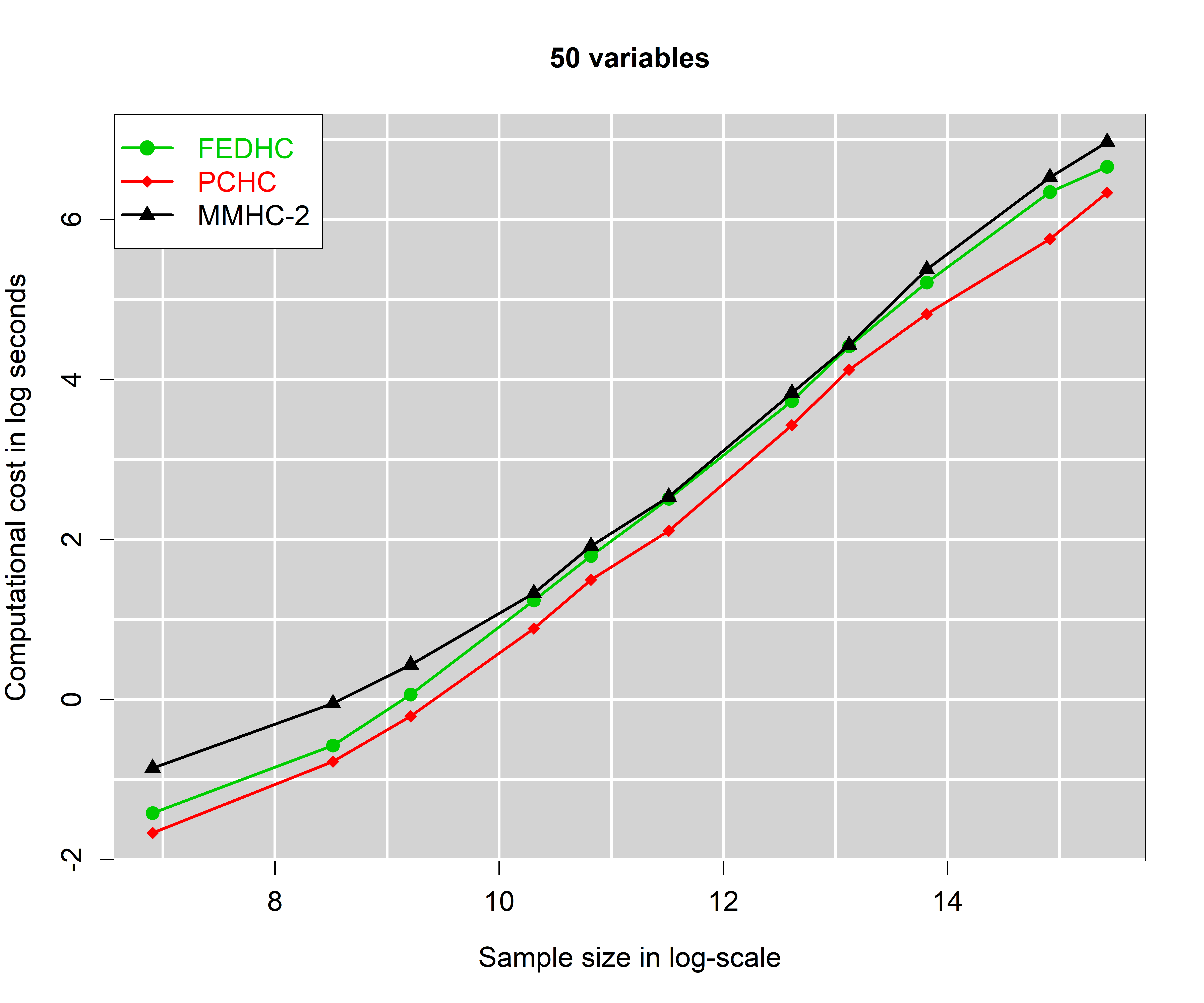}  \\
\multicolumn{2}{c}{\textbf{Continuous data with 5 neighbours on average.}} \\
\includegraphics[scale = 0.38, trim = 70 10 0 0]{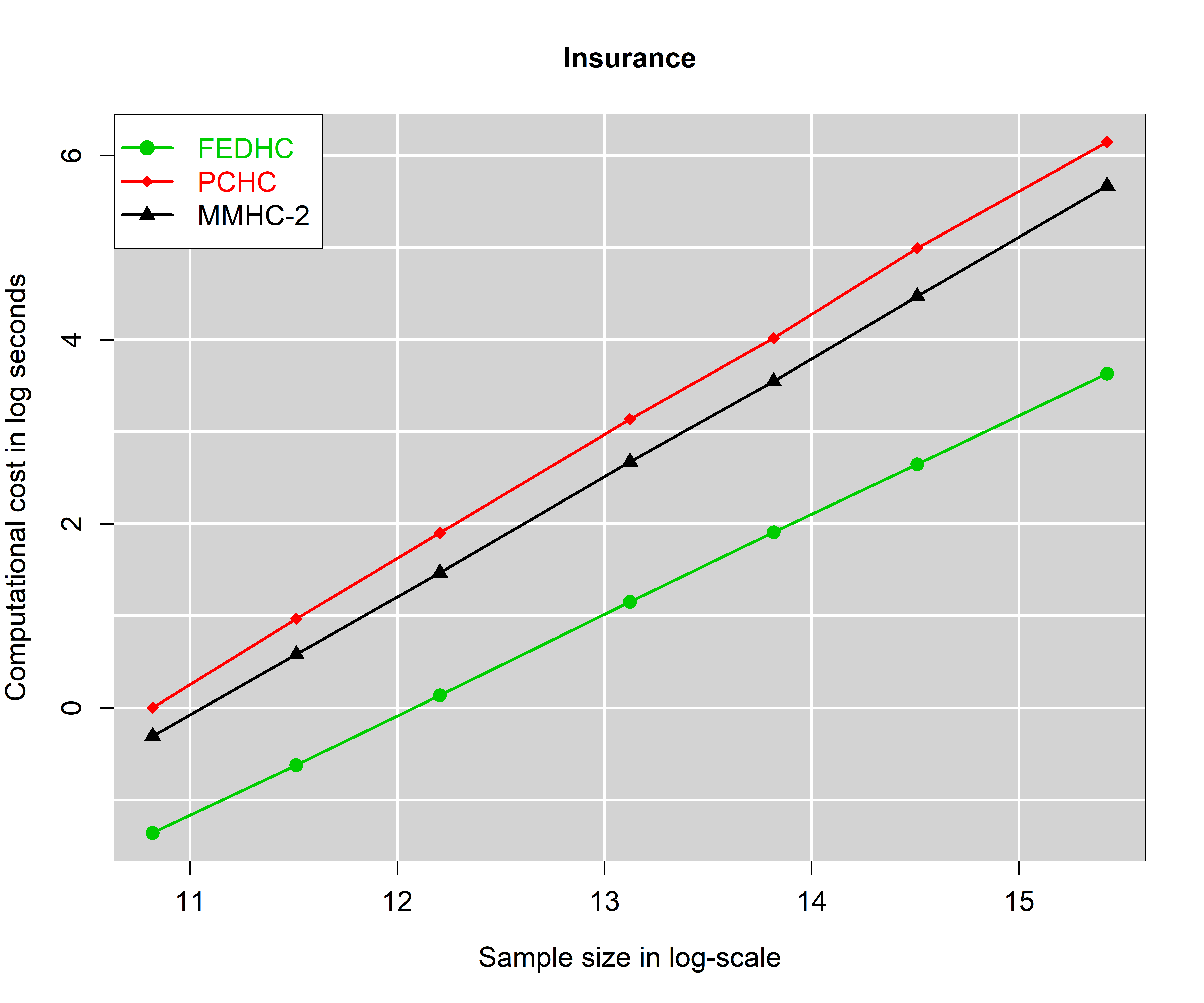}  &
\includegraphics[scale = 0.38, trim = 50 10 0 0]{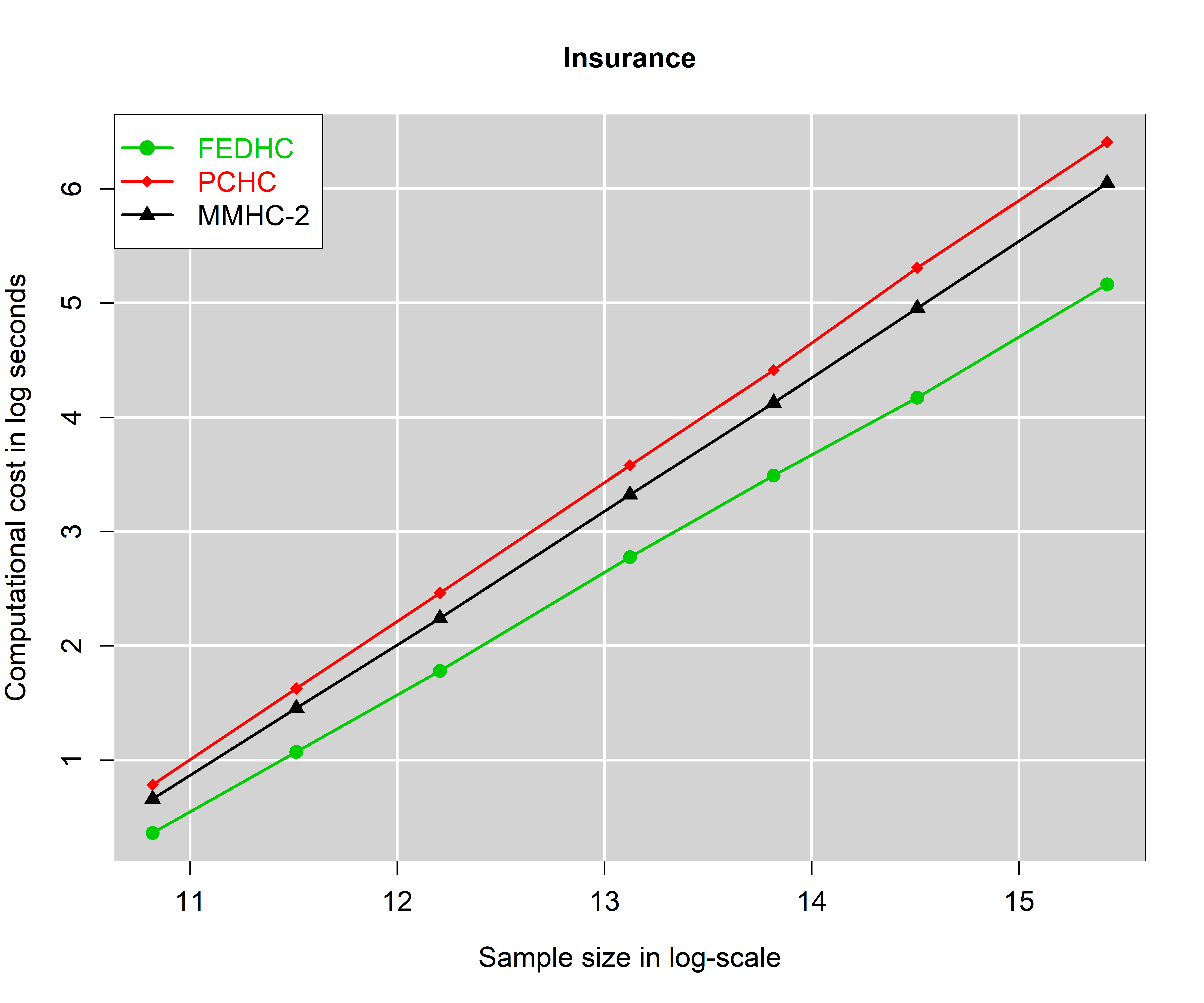}  \\
\multicolumn{2}{c}{\textbf{Categorical data.}} 
\end{tabular}
\caption{Scalability of the algorithms with respect to the sample size for some selected cases. The results for the other cases convey the same message and are thus not presented. The left column refers to the skeleton identification phase whereas the right column to both phases. \label{scalab} }
\end{figure}

The computational cost of each algorithm was also measured appearing in Figure \ref{scalab} as a function of the sample size. The empirical slopes of all lines in Figure \ref{scalab} are nearly equal to 1 indicating that the scalability of FEDHC, PCHC, and MMHC-2 is linear in the sample size. Hence, the computational cost of all algorithms increases linearly with respect to the sample size. For any percentage-wise increase in the sample size, the time increases by the same percentage. Surprisingly enough, the computational cost of MMHC-1 was not evaluated for categorical data case because similarly to the continuous data case it was too high to evaluate.

It is surprising though that the computational cost of FEDHC is similar to that of PCHC and MMHC-2. In fact the skeleton identification phase requires about the same amount of time and it requires only 8 seconds with 5 million observations. The scoring phase is the most expensive phase of the algorithms, absorbing 73\%-99\% of the total computation time. Regarding FEDHC and MMHC-2, since the initial phase of both has been implemented in \textit{R}, this can be attributed to the fact that the calculations of the partial correlation in FEDHC are heavier than those in MMHC-2 because the conditioning set in the former can grow larger than the conditioning set in MMHC-2 which is always bounded by a pre-specified value $k$. Thus, MMHC-2 performs more but computationally lighter calculations than FEDHC.

\section{Illustration of the algorithms on real economics data using the \textit{R} package \textit{pchc} } \label{real}
The \textit{R} package \textit{pchc} is first presented and then two examples with the datasets used in \cite{tsagris2021} illustrate the performance of FEDHC against its competitors, PCHC, MMHC-1, MMHC-2. The advantages of BNs have already been discussed in \cite{tsagris2021} and hence the examples focus on the comparison of FEDHC to PCHC, MMHC-1 and MMHC-2. 

\subsection{Expenditure data}
The first example concerns a dataset with continuous variables containing information on the monthly credit card expenditure of individuals was used. It is the \textbf{Expenditure} dataset \citep{greene2003} and is publicly accessible via the \textit{R} package \textit{AER} \citep{aer2008}. The dataset contains information about 1,319 observations (10\% of the original data set) on the following 12 variables. Whether the application for credit card was accepted or not (\textbf{Card}), the number of major derogatory reports, (\textbf{Reports}), the age in years plus twelfths of a year (\textbf{Age}), the yearly income in \$10,000 (\textbf{Income}), the ratio of monthly credit card expenditure to yearly income (\textbf{Share}), the average monthly credit card expenditure (\textbf{Expenditure}), whether the person owns their home or they rent (\textbf{Owner}), whether the person is self employed or not (\textbf{Selfemp}), the number of dependents + 1 (\textbf{Dependents}), the number of months living at current address (\textbf{Months}), the number of major credit cards held (\textbf{Majorcards}) and the number of active credit accounts (\textbf{Active}).

The \textit{R} package \textit{AER} contains the data and must be loaded and be processed for the algorithms to run. \\
\begin{verbatim}
> library(AER)  ## CreditCard are available 
> library(bnlearn)  ## To run MMHC-1
> data(CreditCard)  ## load the data
> x <- CreditCard
> colnames(x) <- c( "Card", "Reports", "Age", "Income", "Share", "Expenditure",
+             "Owner", "Selfemp", "Dependents", "Months",  "Majorcards", "Active" )
## Prepare the data 
> for (i in 1:12)  x[, i] <- as.numeric(x[, i]) 
> x <- as.matrix(x)
> x[, c(1, 7, 8)] <- x[, c(1, 7, 8)] - 1
## Run all 4 algorithms
> a1 <- bnlearn::mmhc( as.data.frame(x), restrict.args = 
+                      list(alpha = 0.05, test = "zf") )
> a2 <- pchc::mmhc(x, alpha = 0.05)
> a3 <- pchc::pchc(x, alpha = 0.05)
> a4 <- pchc::fedhc(x, alpha = 0.05)
\end{verbatim}

In order to plot the fitted BNs of each algorithm the following commands were used. 
\begin{verbatim}
> pchc::bnplot(a1)
> pchc::bnplot(a2$dag)
> pchc::bnplot(a3$dag)
> pchc::bnplot(a4\$dag)
\end{verbatim}

This example shows the practical usefulness of the BNs. Evidently this small scale experiment shows that  companies can customize their products according to the key factors that determine the consumers’ behaviour. Instead of selecting one variable only, a researcher/practitioner can identify the relationships among all variables by estimating the causal mechanistic system that generated the data. The BN can further reveal information about the variables that are statistically significantly related. 

\begin{figure}[!ht]
\centering
\begin{tabular}{cc}
\includegraphics[scale = 0.48, trim = 70 0 0 0]{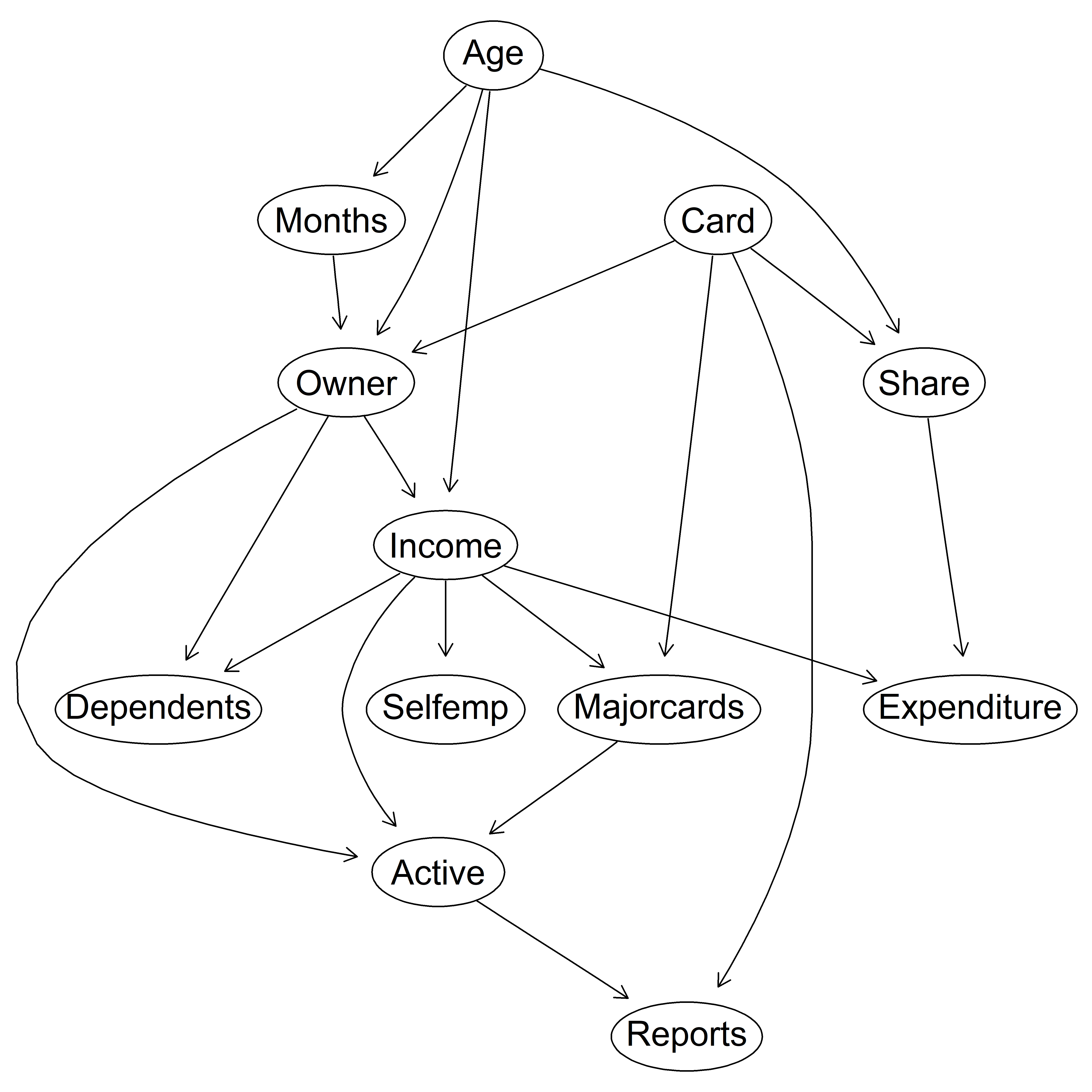}  &
\includegraphics[scale = 0.48, trim = 20 0 0 0]{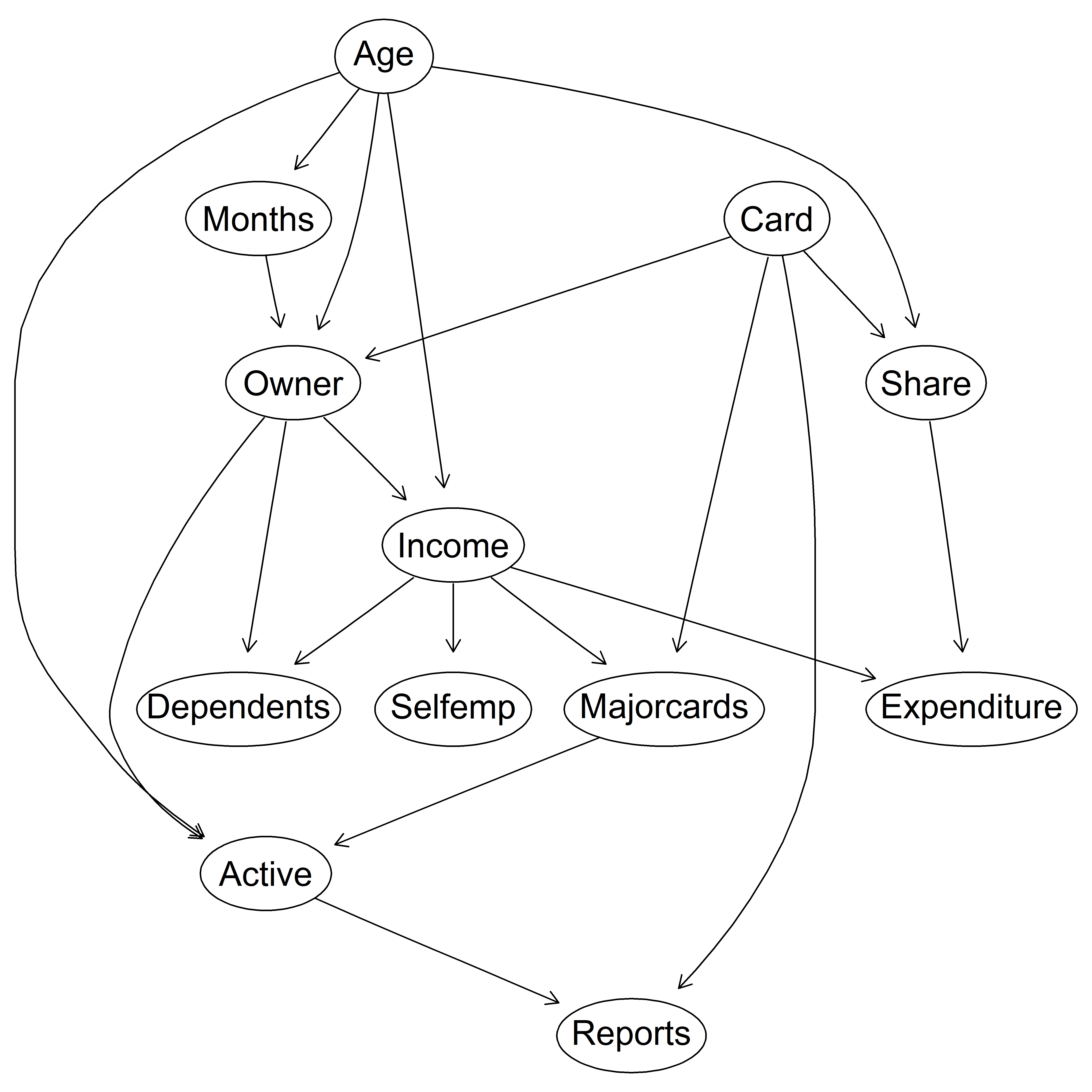}  \\
(a) FEDHC.  &  (b) PCHC.  \\
\includegraphics[scale = 0.48, trim = 70 0 0 0]{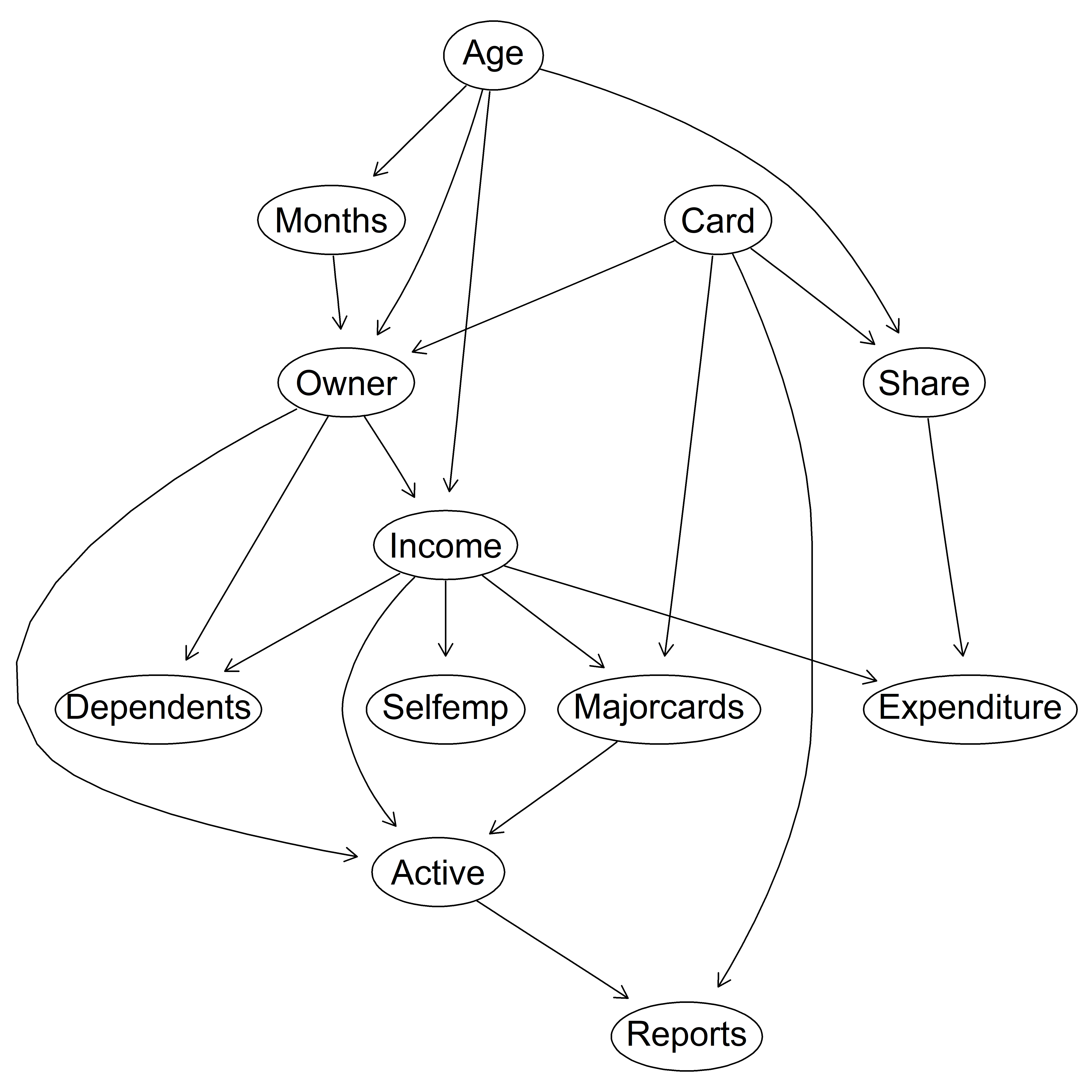}  &
\includegraphics[scale = 0.48, trim = 20 0 0 0]{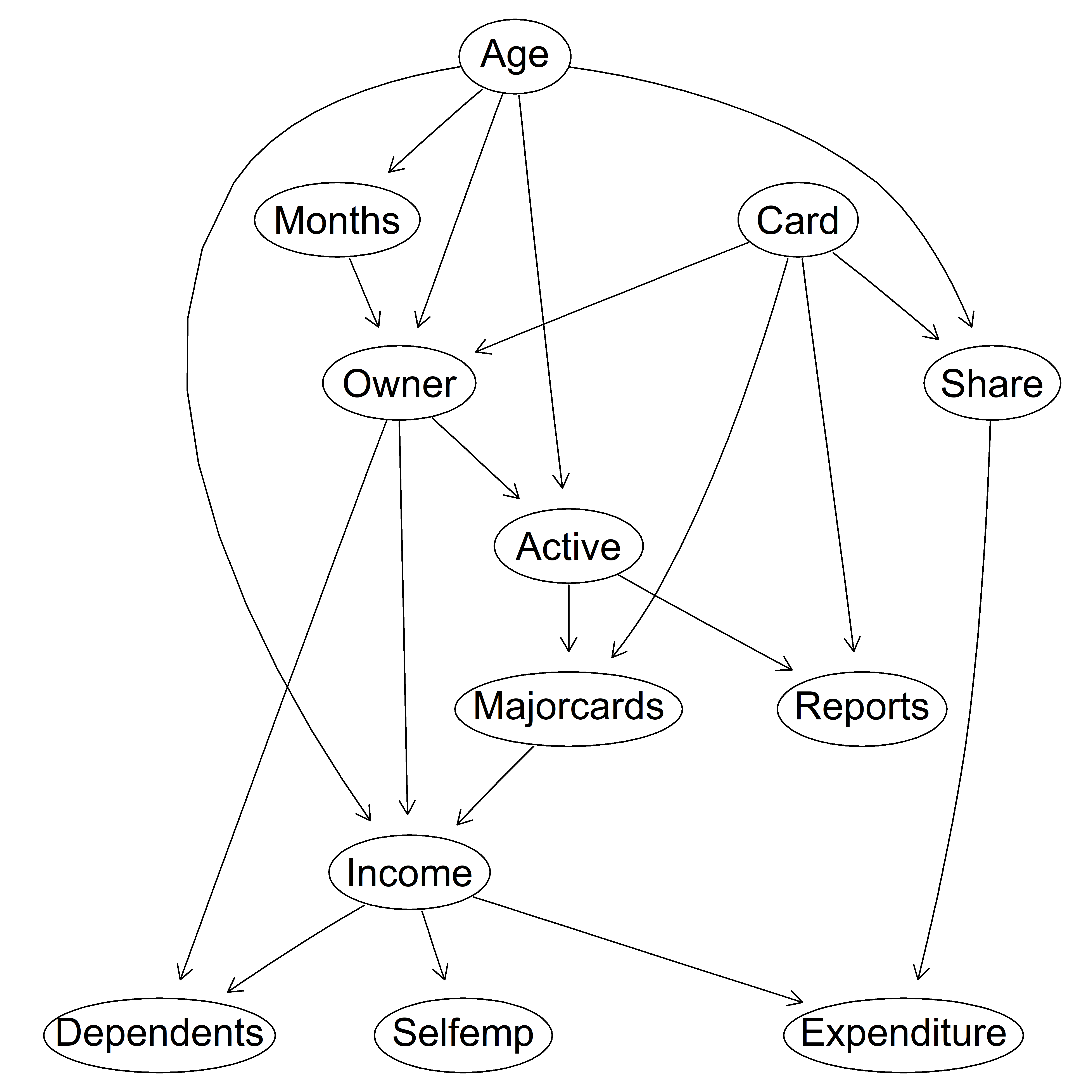}  \\
(c)  MMHC-1.  &  (d) MMHC-2.  
\end{tabular}
\caption{\textbf{Expenditure data}. Estimated BNs using (a) FEDHC, (b) PCHC, (c) MMHC-1 and (d) MMHC-2. \label{fig_expenditure} }
\end{figure}

According to FEDHC (Figure \ref{fig_expenditure})(a), the age of the individual affects their income, the number of months they have been living at their current address, whether they own their home or not, and the ratio of their monthly credit card expenditure to their yearly income. The only variables associated with the number of major derogatory reports (Reports) are whether the consumer's application for credit card was accepted or not (Card) and the number of active credit accounts (Active). In fact these two variables are parents of Reports as the arrows are directed towards it.  A third advantage of BNs is that they provide a solution to the variable selection problem. The parents of the variable Majorcards (number of major credit cards held) are Card (whether the application for credit card was accepted or not) and  Income (yearly income in \$10,000), its only child is Active (number of active credit accounts) and and it only spouse (parent of Active) is Owner (whether the consumer owns their home). The collection of those parents, children and spouses form the Majorcards' MB. That is, any other variable does not increase the  information on the number of major credit cards held by the consumer. For any given variable one can straightforwardly obtain (and visualise) its MB which can be used for the construction of the appropriate linear regression model.

Figure \ref{fig_expenditure} contains the BNs using both implementations of MMHC, the PCHC and the FEDHC algorithms fitted to the expenditure data with the variables sorted in a topological order \citep{chickering1995}, a tree-like structure. The BIC of the BN learned by MMHC-1 and MMHC-2 are equal to $-32171.75$ and $-32171.22$, and for the PCHC and FEDHC are both equal to $-32171.75$. This is an indication that all four algorithms produced BNs of nearly the same quality. On a closer examination of the graphs one can detect some differences between the algorithms. For instance \textbf{Age} is directly related to \textbf{Active} according to PCHC and MMHC-2 but not according to FEDHC and MMHC-1. Further, all algorithms have identified the \textbf{Owner} as the parent of \textbf{Income} and not vice-versa. This is related to the prior knowledge discussed in Section \ref{prior} and will be examined in the next categorical example dataset.

\subsection{Income data}
The second example data set contains categorical variables and originates from an example in the book "The Elements of Statistical Learning" \citep{friedman2001} and is publicly available from the \textit{R} package \textit{arules} \citep{arules2011}. It consists of $6,876$ instances (obtained from the original data set with $9,409$ instances, by removing observations with missing annual income) and a mixture of 13 categorical and continuous demographic variables. The continuous variables were discretised as suggested by the authors of the package. The continuous variables (age, education, income, years in bay area, number in household, and number of children) were discretised based on their median values. \textbf{Income}:  "\$0-\$40,000" or "\$40,000+", \textbf{Sex}: "male" or "female", \textbf{Marriage}: "married", "cohabitation", "divorced", "widowed" or "single", \textbf{Age}: "14-34" or "35+", \textbf{Education}: "college graduate" or "no college graduate", \textbf{Occupation}, "professional/managerial", "sales", "laborer", "clerical/service", "homemaker", "student", "military", "retired" or "unemployed", \textbf{Bay} (number of years in bay area): "1-9" or "10+", \textbf{No of people} (number of of people living in the house): "1" or "2+", \textbf{Children}: "0" or "1+", \textbf{Rent}: "own", "rent" or "live with parents/family", \textbf{Type}: "house", "condominium", "apartment", "mobile home" or "other", \textbf{Ethnicity}: "American Indian", "Asian", "black", "east Indian", "hispanic", "white", "pacific islander" or "other" and \textbf{Language} ( language spoken at home): "english", "spanish" or "other".

The dataset is at first accessed via the \textit{R} package \textit{arules} and is prepossessed as suggested in \textit{arules}. \\

\begin{verbatim}
> library(arules)
> data(IncomeESL)
## remove incomplete cases
> IncomeESL <- IncomeESL[complete.cases(IncomeESL), ]
## preparing the data set
> IncomeESL[["income"]] <- factor((as.numeric(IncomeESL[["income"]]) > 6) + 1,
+ levels = 1 : 2 , labels = c("0-40,000", "40,000+"))
> IncomeESL[["age"]] <- factor((as.numeric(IncomeESL[["age"]]) > 3) + 1,
+ levels = 1 : 2 , labels = c("14-34", "35+"))
> IncomeESL[["education"]] <- factor((as.numeric(IncomeESL[["education"]]) > 4) + 
+ 1, levels = 1 : 2 , labels = c("no college graduate", "college graduate"))
> IncomeESL[["years in bay area"]] <- factor(
+ (as.numeric(IncomeESL[["years in bay area"]]) > 4) + 1,
+ levels = 1 : 2 , labels = c("1-9", "10+"))
> IncomeESL[["number in household"]] <- factor(
+ (as.numeric(IncomeESL[["number in household"]]) > 3) + 1,
+ levels = 1 : 2 , labels = c("1", "2+"))
> IncomeESL[["number of children"]] <- factor(
+ (as.numeric(IncomeESL[["number of children"]]) > 1) + 0,
+ levels = 0 : 1 , labels = c("0", "1+"))
\end{verbatim}

Some more steps are required prior to running the BN algorithms. \\

\begin{verbatim}
> x <- IncomeESL
> x <- x[, -8]
> colnames(x) <- c( "Income", "Sex", "Marriage", "Age", "Education", "Occupation", 
+     "Bay", "No of people", "Children", "Rent", "Type", "Ethnicity", "Language" )
> nam <- colnames(x)
\end{verbatim}

The importance of prior knowledge incorporation discussed in Section \ref{prior} becomes evident in this example. Prior knowledge can be added in the \textbf{blacklist} argument denoting the forbidden directions (arrows). \\

\begin{verbatim}
> black <- matrix(nrow = 26, ncol = 2)
> black <- as.data.frame(black)
> for (i in 1:13)  black[i, ] <- c(nam[i], nam[2])
> for (i in 1:13)  black[13 + i, ] <- c(nam[i], nam[4])
> black <- black[-c(2, 17), ]
> black <- rbind( black, c(nam[9], nam[3]) )
> black <- rbind( black, c(nam[3], nam[6]) )
> black <- rbind( black, c(nam[9], nam[6]) )
> black <- rbind( black, c(nam[6], nam[5]) )
> black <- rbind( black, c(nam[3], nam[1]) )
> black <- rbind( black, c(nam[1], nam[5]) )
> black <- rbind( black, c(nam[10], nam[1]) )
> black <- rbind( black, c(nam[10], nam[5]) )
> black <- rbind( black, c(nam[10], nam[6]) )
> black <- rbind( black, c(nam[13], nam[12]) )
> colnames(black) <- c("from", "to") 
\end{verbatim}

Finally, the 4 BN algorithms are applied to the Income data. \\

\begin{verbatim}
> b1 <- bnlearn::mmhc( x, blacklist = black, restrict.args = 
+                      list(alpha = 0.05, test = "mi") )
> b2 <- pchc::mmhc(x, method = "cat", alpha = 0.05, blacklist = black, 
+ score = "bic")
> b3 <- pchc::pchc(x, method = "cat", alpha = 0.05, blacklist = black, 
+ score = "bic")
> b4 <- pchc::fedhc(x, method = "cat", alpha = 0.05, blacklist = black, 
+ score = "bic")
\end{verbatim}

\begin{figure}[!ht]
\centering
\begin{tabular}{cc}
\includegraphics[scale = 0.48, trim = 60 0 0 0]{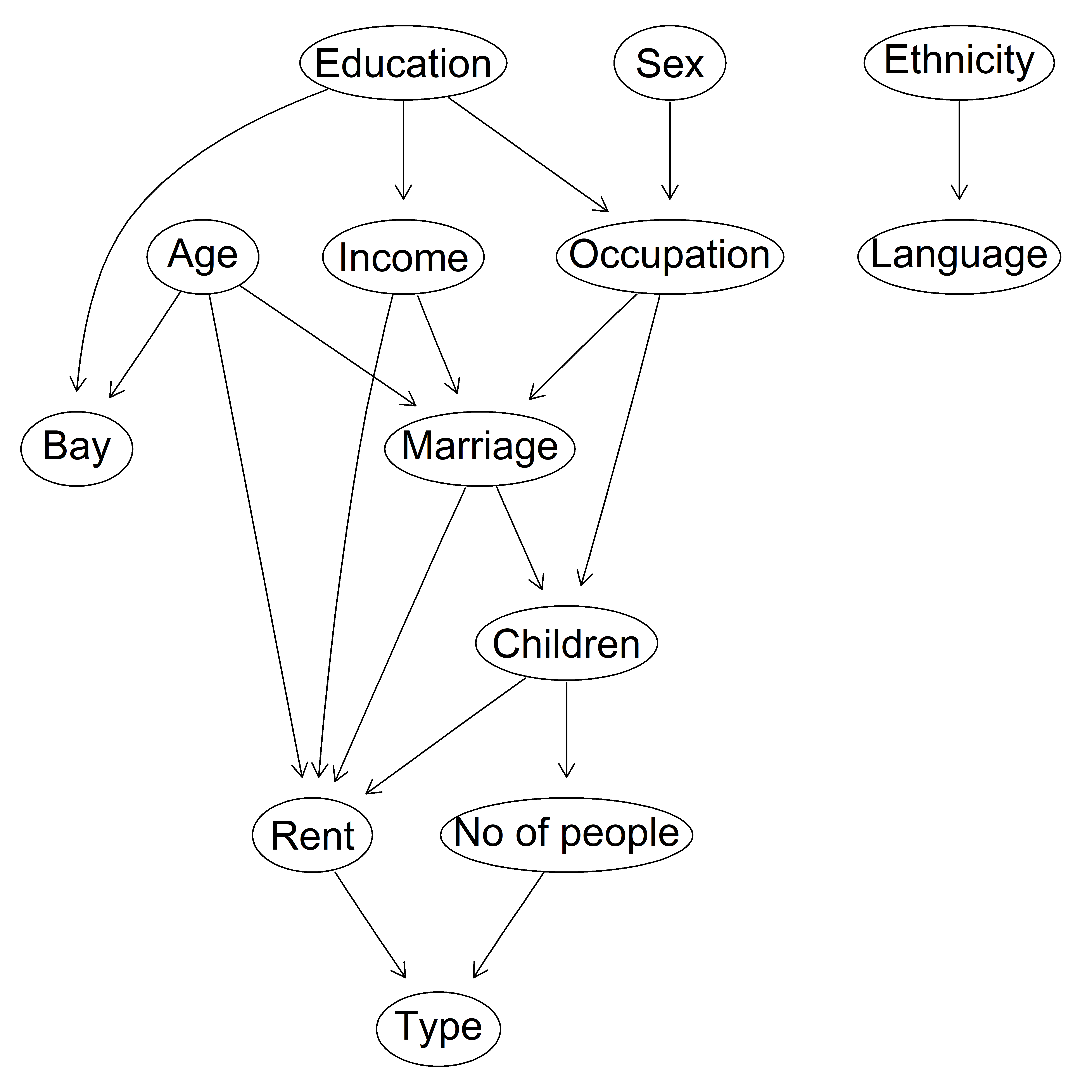}  &
\includegraphics[scale = 0.48, trim = 20 0 0 0]{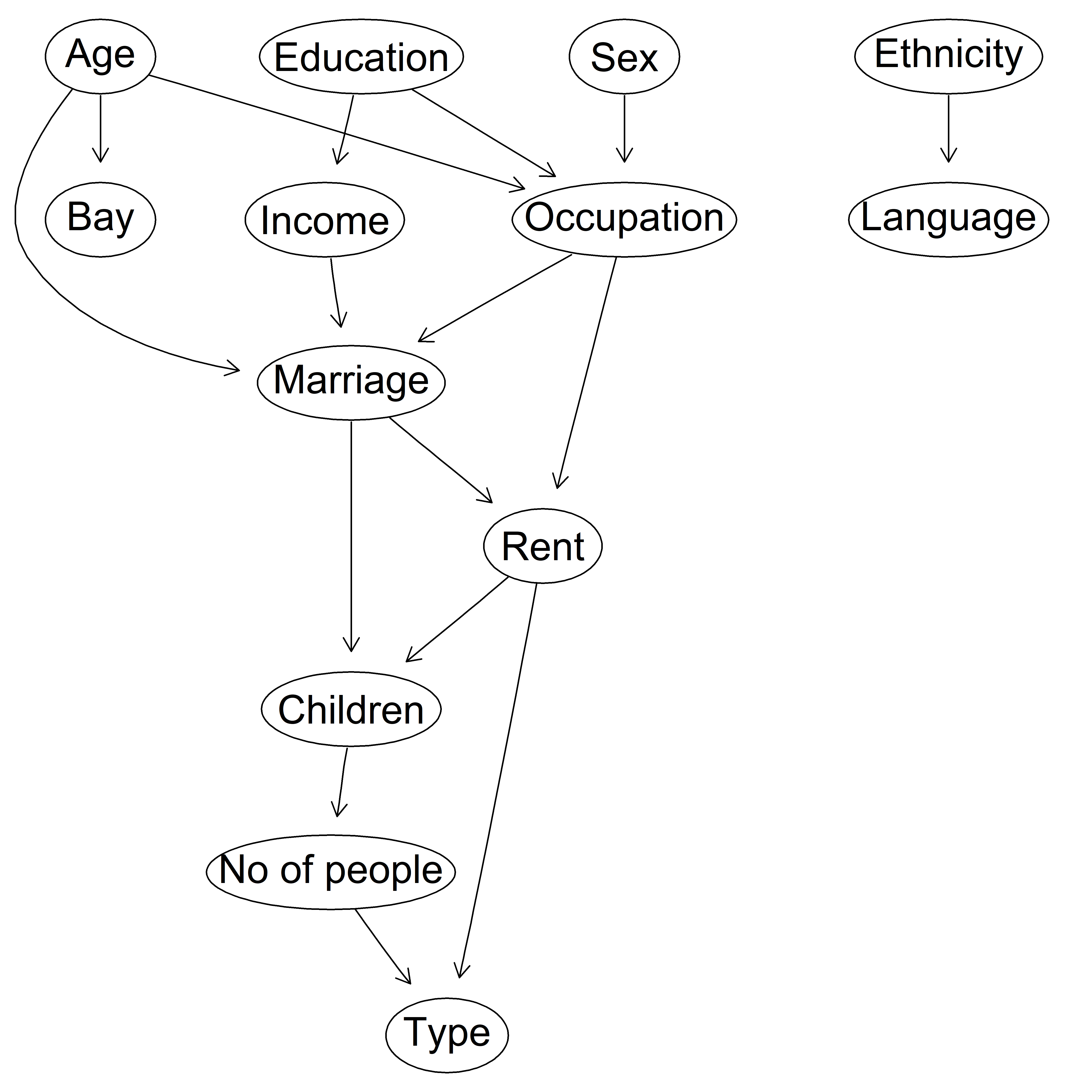}  \\
(a)  FEDHC.  &  (b) PCHC.  \\
\includegraphics[scale = 0.48, trim = 60 0 0 0]{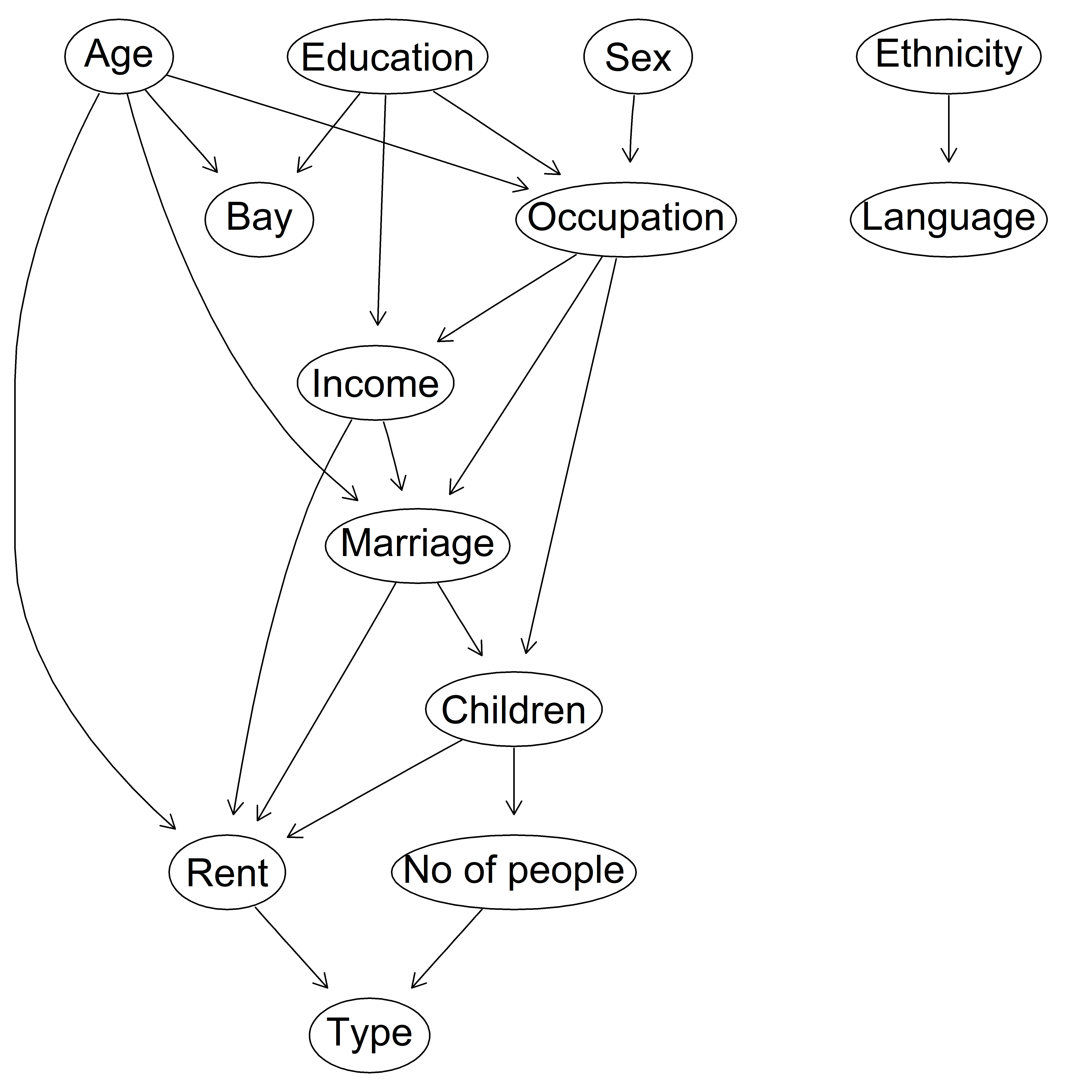}  &
\includegraphics[scale = 0.48, trim = 20 0 0 0]{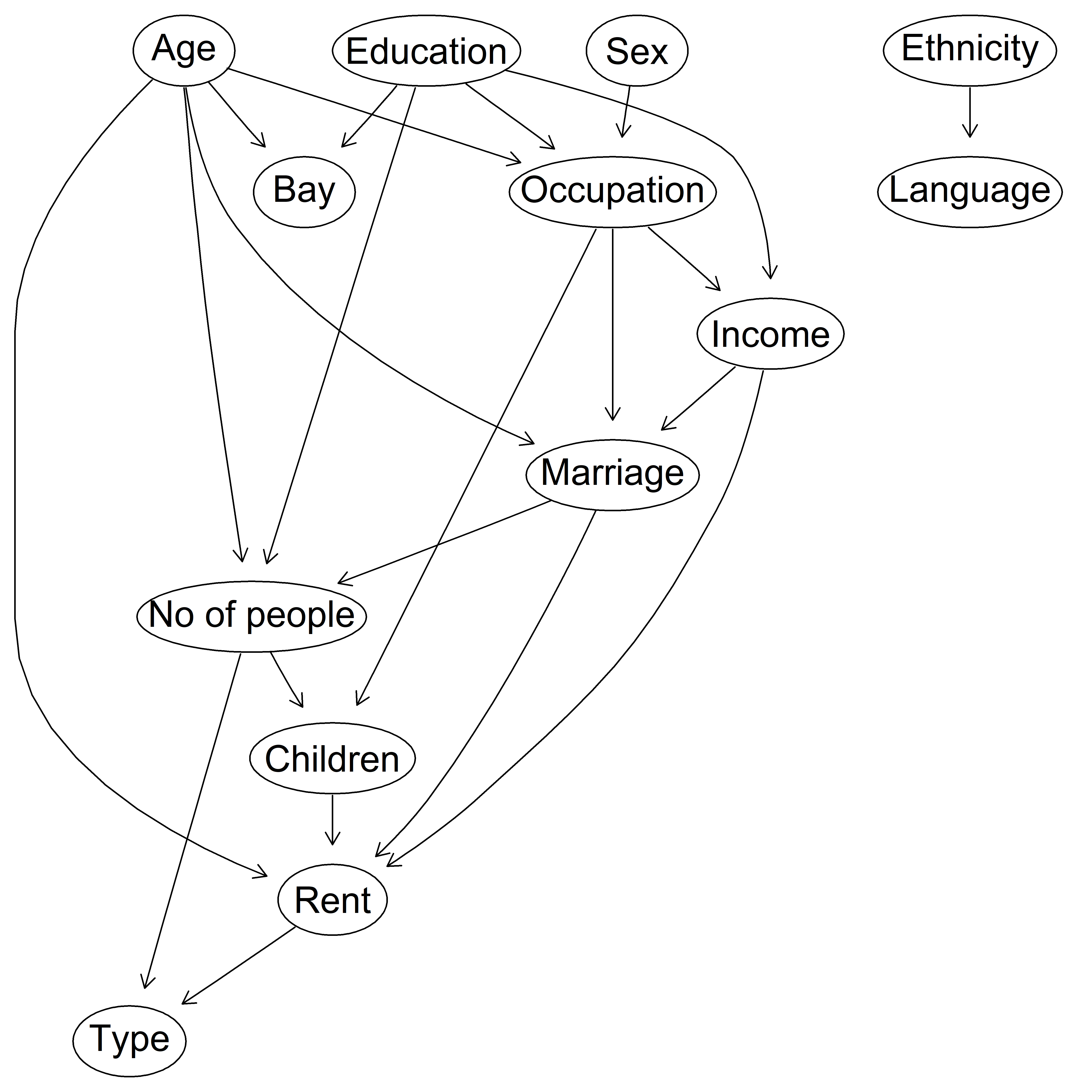}  \\
(c) MMHC-1.  &  (d) MMHC-2.  \\
\end{tabular}
\caption{\textbf{Income data}. Estimated BNs using (a) FEDHC, (b) PCHC, (c) MMHC-1 and (d) MMHC-2. \label{fig_income} }
\end{figure}

Figure \ref{fig_income} presents the fitted BNs of the MMHC-1, MMHC-2, PCHC and FEDHC algorithms. There are some distinct differences between the algorithms. For instance, PCHC is the only algorithm that has not identified \textbf{Education} as the parent of \textbf{Bay}. Also, the BN learned by MMHC-2 is the densest one (contains more arrows) whereas PCHC learned the BN with the fewest arrows.

This example further demonstrates the necessity of prior knowledge. BN learning algorithms fit a model to the data ignoring the underlying truthfulness, ignoring the relevant economic theory. Economic theory can be used as prior knowledge to help mitigate the errors and lead to more truthful BNs. The exclusion of the {blacklist} argument (forbidden directions) would yield some irrational directions, for instance that occupation of age might affect sex or marriage affects age, simply because these directions could increase the score. Finally, BNs are related to synthetic population generation where the data are usually categorical. This task requires the specification of the joint distribution of the data and BNs accomplish this \citep{sun2015}. Based on the Markov condition \ref{markov}, the joint distribution can be written down explicitly allowing for synthetic population generation in a sequential order. One commences by generating values for education and sex. Using these two variables, values for occupation. These values, along with income and age can be used to generate for the marital status and so on.

\section{Conclusions} \label{concl}
This paper proposed to combine the first phase of the FBED variable selection algorithm with the HC scoring phase leading to a new hybrid algorithm, termed FEDHC. Additionally, a new implementation of the MMHC algorithm was provided. Finally the paper presented robustified (against outliers) versions of FEDHC, PCHC and MMHC. The robustified version of FEDHC was shown to be even nearly 40 times faster than the raw version and yielded BNs of higher quality, when outliers were present. Simulation studies manifested that in terms of computational efficiency, FEDHC is comparable to PHCHC and along with MMHC-2 FEDHC was able to fit BNs to continuous data with sample sizes at the order of hundreds of thousands in a few seconds and at the order of millions in a few minutes. It must be highlighted though that the skeleton identification phase of FEDHC and MMHC-1 have been implemented in \textit{R} and not in \textit{C++}. Additionally, FEDHC was always executing significantly fewer CI tests than its competitors. Ultimately, in terms of accuracy, FEDHC outperformed is competitors with continuous data, and it was more accurate than  or on par with MMHC-1 and MMHC-2 with categorical data, but less accurate than PCHC. 

The rationale of MMHC and PCHC is to perform variable selection to each node and then apply a HC to the resulting network. On the same spirit, \cite{meinshausen2006} used LASSO for variable selection with the scopus of constructing an un-directed network. The HC phase could be incorporated in the graphical LASSO to learn the underlying BN. Broadly speaking, the combination of a network learning phase with a scoring phase can yield hybrid algorithms. Other modern hybrid methods for BN learning include \cite{kuipers2020} on hybrid structure learning and sampling. They combine constraint-based pruning with MCMC inference schemes (also to improve the overall search space) and find a combination that is relatively efficient with relatively good performance. The constraint-based part is interchangeable and could connect well with MMHC, PCHC, or FEDHC. 

FEDHC is not the first algorithm that has outperformed MMHC. Recent algorithms include PCHC \citep{tsagris2021}, the SP algorithm for Gaussian DAGs \citep{raskutti2018} and the NOTEARS \citep{zheng2018}. The algorithms of \cite{zhang2012} and \cite{chalupka2018} were also shown to outperform MMHC in the presence of latent confounders, not examined here. The advantage of the latter two is that employ non-parametric tests, such as kernel CI test, thus allowing for non-linear relationships. BNs that detect non-linear relationships among the variable, such as the algorithms proposed by \cite{zhang2012} and \cite{chalupka2018} is what this paper did not cover. Further, our comparative analysis was only with MMHC \citep{tsamardinos2006} and PCHC \citep{tsagris2021} due to their close relationship with FEDHC.

Future research includes a comparison of all algorithms in terms of more directions. For instance, a) the effect of Pearson and Spearman CI tests and the effect of $X^2$ and $G^2$ CI tests, b) the effect of the outliers, c) the effect of the scoring methods (Tabu search and HC) the effect of the average neighbours (network density), and e) the effect of the number of variables on the quality of the BN learned by either algorithm. These directions can be used to numerically evaluate the asymptotic properties of the BN learning algorithms with tens of millions of observations. Another interesting direction is the incorporation of fast non-linear CI tests, such as the distance correlation \citep{szekely2007,szekely2014,huo2016,shen2022}. The distance correlation could be utilized during the skeleton identification of the FEDHC mainly because it performs fewer CI tests than its competitors.

%\begin{table}
%\begin{tabular}{lll}
%\hline\noalign{\smallskip}
%\end{tabular}
%\end{table}

%\begin{acknowledgements}
%\end{acknowledgements}

\clearpage
\section*{Appendix}
\subsection*{A: Conditional independence tests}
The type of CI tests executed during the skeleton identification phase depends upon the nature of the data and they are used to test the following. Let $X$ and $Y$ be two random variables, and $\mathbf{Z}$ be a (possibly empty) set of random variables. Statistically speaking, $X$ and $Y$ are conditionally independent given $\mathbf{Z}$ $\left( X\perp\!\!\!\perp Y | {\bf Z} \right)$, if $P(X,Y|\mathbf{Z}) = P(X|\mathbf{Z}) \cdot P(Y|\mathbf{Z})$ and this holds for all values of $X$, $Y$ and $\mathbf{Z}$. Equivalently, conditional independence of $X$ and $Y$ given $\mathbf{Z}$ implies $P(X|Y,\mathbf{Z}) = P(X|\mathbf{Z})$ and $P(Y|X,\mathbf{Z}) = P(Y|\mathbf{Z})$.

\subsubsection*{Pearson correlation for continuous data}
A frequently employed CI test for two continuous variables $X$ and $Y$ conditional on a set of variables ${\bf Z}$ is the partial correlation test \citep{baba2004} that assumes linear relationships among the variables. The test statistic for the partial Pearson correlation is given by
\begin{eqnarray} \label{pearson}
T_p = \frac{1}{2}\left|\log{\frac{1+r_{X,Y|{\bf Z}}}{1-r_{X,Y|{\bf Z}}}} \right| \sqrt{n - |{\bf Z}| - 3},
\end{eqnarray}
where $n$ is the sample size, $|{\bf Z}|$ denotes the number of conditioning variables and $r_{X,Y|{\bf z}}$ is the partial Pearson correlation\footnote{The partial correlation is efficiently computed using the correlation matrix of $X$, $Y$ and ${\bf Z}$ \citep{baba2004}.} of $X$ and $Y$ conditioning on $\bf Z$\footnote{In the \textit{R} package \textit{Rfast}'s implementation of the PC algorithm compares $T_p$ (\ref{pearson}) against a $t$ distribution with $n-|{\bf Z}|-3$ degrees of freedom, whereas the MMHC algorithm in the \textit{R} package \textit{bnlearn} compares $T$ against the standard normal distribution. The differences are evident in small sample sizes, but become negligible when the sample sizes are at the order of a few tens.}. When ${\bf Z}$ is empty ($|{\bf Z}| = 0$), the partial correlation drops to the usual Pearson correlation coefficient. 

\subsubsection*{Spearman correlation for continuous data}
The non-parametric alternative that is assumed to be more robust to outliers is the Spearman correlation coefficient. Spearman correlation is equivalent to the Pearson correlation applied to the ranks of the variables. Its test statistic though is given by $T_s = T_p \times 1.029563$ \citep{fieller1957,fieller1961}. 

\subsubsection*{$G^2$ test of independence for categorical data}
The $G^2$ test of independence of two categorical variables $X$ and $Y$ conditional on a set of variables ${\bf Z}$ is defined as \citep{agresti2002}
\begin{eqnarray} \label{g2}
G^2=2\sum_l\sum_{i, j}O_{ij|l}\log{\frac{O_{ij|l}}{E_{ij|l}}},
\end{eqnarray}
where $O_{ij}$ are the observed frequencies of the $i$-th and $j$-th values of $X$ and $Y$ respectively for the $l$-th value of $\bf Z$. The $E_{ij}$ are their corresponding expected frequencies computed by $E_{ij}=\frac{O_{i+|l}O_{+j|l}}{O_{++|l}}$, where $O_{i+|l} = \sum_{j=1}^nO_{ij|l}$, $O_{+j|l}=\sum_{i=1}^nO_{ij|l}$ and $O_{++|l}=n_l$. Under the conditional independence assumption, the $G^2$ test statistic follows the $\chi^2$ distribution with $(|X| - 1) (|X| - 1) (|{\bf Z}| - 1)$ degrees of freedom, where $|{\bf Z}|$ refers to the cardinality of ${\bf Z}$, the total number of values of ${\bf Z}$

\subsubsection*{$X^2$ test of independence for categorical data}
Alternatively, one could use the Pearson $X^2$ test statistic 
$X^2= \sum_l\sum_{i, j}\frac{\left(O_{ij|l} - E_{ij|l}\right)^2}{E_{ij|l}^2}$ that has the same properties as the $G^2$ test statistic (\ref{g2}). The drawback of $X^2$ is that it cannot be computed when $E_{ij|l}=0$. On the contrary, $G^2$ is computed in such cases since $\lim_{x\rightarrow 0}x\log x = 0$. For either aforementioned test, when $|{\bf Z}|$ is the empty set, both tests examine the unconditional association between variables $X$ and $Y$\footnote{For a practical comparison between the two tests based on extensive simulation studies see \cite{alenazi2020}.}. 

\subsubsection*{Permutation based p-values}
The aforementioned test statistics produce asymptotic p-values. In case of small sample sizes computationally intensive methods like permutations might be preferable. With continuous variables for instance, when testing for unconditional independence the idea is to distort the pairs multiple times and each time calculate the relevant test statistic. For the conditional independence of $X$ and $Y$ conditional on $\bf Z$ the partial correlation is computed from the residuals of two linear regression models, $X \sim {\bf Z}$ and $Y \sim {\bf Z}$. In this instance, the pairs of the residual vectors are distorted multiple times. With categorical variables, this approach is more complicated and care must be taken so as to retain the row and column totals of the resulting contingency tables. For either case, the p-value is then computed as the proportion of times the permuted test statistics exceed the observed test statistic that computed using the original data. Permutation based techniques have shown to improve the quality of BNs \citep{tsamardinos2010} in small sample sized cases. On the contrary, the FEDHC algorithm aims at making inference on datasets with large sample sizes, for which asymptotic statistical tests are valid and reliable enough to produce correct decisions. 

\subsection*{B: Computational details of FEDHC}
With continuous data, the correlation matrix is computed once and utilised throughout the skeleton identification phase. FEDHC returns the correlation matrix and the matrix of the p-values of all pairwise associations that is useful in a second run of the algorithm with a different significance level. This is a significant advantage when BNs have to fit to large scale datasets and the correlation matrix can be given as an input to FEDHC to further reduce FEDHC's computational cost. 

The partial correlation coefficient is given by
\begin{eqnarray*}
r_{X,Y|{\bf Z}}=
\left\lbrace
\begin{array}{cc}
\frac{R_{X,Y} - R_{X,z} R_{Y,z}}{
\sqrt{ \left(1 - R_{X,Z}^2\right)^T \left(1 - R_{Y,z}^2\right) }} & \text{if} \ \ |{\bf Z}|=1 \\
-\frac{ {\bf A}_{1,2} }{ \sqrt{{\bf A}_{1,1}{\bf A}_{2,2}} } & \text{if} \ \ |{\bf Z}| > 1 
\end{array}
\right\rbrace,
\end{eqnarray*}
where $R_{X,Y}$ is the correlation between variables $X$ and $Y$, $R_{X,Z}$ and $R_{Y,Z}$ denote the correlations between $X$ \& $Z$ and $Y$ \& $Z$. ${\bf A}=R_{X,Y,{\bf Z}}^{-1}$, with ${\bf A}$ denoting the sub correlation matrix of variables $X, Y, {\bf Z}$ and $A_{i,j}$ symbolises the element in the $i$-row and $j$-th column of matrix $A$.

The CI tests executed during the initial phase compute the logarithm of the p-value, instead of the p-value itself, to avoid numerical overflows observed with a large test statistic that produces a p-value equal to $0$. Additionally, the computational cost of FEDHC's first phase can be further reduced via parallel programming. 

It is also possible to store the p-values of each CI test for future reference. When a different significance level must be used, this will further decrease the associated computational cost of the skeleton identification phase in a second run. However, as will be exposed in section \ref{scalability}, the cost of this phase is really small (a few seconds) even for millions of observations. The largest portion of this phase's computational cost is attributed to the calculation of the correlation matrix, which can be passed into subsequent runs of the algorithm. 

Finally, \cite{tsagris2021} disregarded the potential of applying the PC-orientation rules \citep{spirtes1991,spirtes2000} prior to the scoring phase as a means of improving the performance of FEDHC and MMHC and this is not pursued any further. 

\subsection*{C: The \textit{R} package \textit{pchc}}
The package \textit{pchc} was first launched in \textit{R} in July 2020 and initially contained the PCHC algorithm. It now includes the FEDHC and MMHC-2 algorithms, functions for testing (un)conditional independence with continuous and categorical data, data generation, BN visualisation and utility functions. It imports the \textit{R} packages \textit{bnlearn}, \textit{Rfast} and the built-in package \textit{stats}. \textit{pchc} is distributed as part of the CRAN R package repository and is compatible with MacOS-X, Windows, Solaris and Linux operating systems. Once the package is installed and loaded
\begin{verbatim}
> install.packages("pchc")
> library(pchc)
\end{verbatim}
\noindent it is ready to use without internet connection. The signature of the function \textbf{fedhc} along with a short explanation of its arguments is displayed below. \\
\begin{verbatim}
> fedhc(x, method = "pearson", alpha = 0.05, robust = FALSE, ini.stat = NULL,
+  R = NULL, restart = 10, score = "bic-g", blacklist = NULL, whitelist = NULL)
\end{verbatim}

\begin{itemize}
\item x: A numerical matrix with the variables. If you have a data.frame (i.e. categorical data) turn them into a matrix using. Note, that for the categorical case data, the numbers must start from 0. No missing data are allowed.
\item method: If you have continuous data, this must be "pearson" (default value) or "cat" if you have categorical data though. With categorical data one has to make sure that the minimum value of each variable is zero. The function \textit{g2test} from the package \textit{Rfast} and the relevant functions work that way.
\item alpha: The significance level for assessing the p-values. The default value is 0.05.
\item robust: If outliers are be removed prior to applying the FEDHC algorithm this must be set to TRUE.
\item ini.stat: If the initial test statistics (univariate associations) are available, they can passed to this argument.
\item R: If the correlation matrix is available, pass it here.
\item restart: An integer, the number of random restarts. The default value is 10.
\item score: A character string, the label of the network score to be used in the algorithm. If none is specified, the default score is the Bayesian Information Criterion for continuous data sets. The available score for continuous variables are: "bic-g" (default value), "loglik-g", "aic-g", "bic-g" or "bge". The available score categorical variables are: "bde", "loglik" or "bic".
\item blacklist: A data frame with two columns (optionally labeled "from" and "to"), containing a set of forbidden directions.
\item whitelist: A data frame with two columns (optionally labeled "from" and "to"), containing a set of must-add directions.
\end{itemize}

The output of the \textbf{fedhc} function is a list including:
\begin{itemize} 
\item ini: A list including the output of the \textit{fedhc.skel} function.
\item dag: A "bn" class output, a list including the outcome of the Hill-Climbing phase. See the package \textit{bnlearn} for more details.
\item scoring: The highest score value observed during the scoring phase.
\item runtime: The duration of the algorithm.
\end{itemize}

\end{document}